\documentclass[10pt,twocolumn,letterpaper]{article}

\usepackage[pagenumbers]{cvpr} %

\usepackage{graphicx}
\usepackage{amsmath}
\usepackage{amssymb}
\usepackage{booktabs}
\usepackage{amsmath}
\usepackage{physics}
\usepackage{multirow}
\usepackage{array}
\usepackage{algpseudocode}
\usepackage{algorithm}
\usepackage{tcolorbox}
\usepackage{colortbl}

\usepackage[dvipsnames]{xcolor}

\definecolor{cvprblue}{rgb}{0.21,0.49,0.74}
\usepackage[pagebackref,breaklinks,colorlinks,citecolor=cvprblue]{hyperref}

\usepackage[capitalize]{cleveref}
\crefname{section}{Sec.}{Secs.}
\Crefname{section}{Section}{Sections}
\Crefname{table}{Table}{Tables}
\crefname{table}{Tab.}{Tabs.}

\def\paperID{9374} %
\def\confName{ICCV}
\def\confYear{2025}

\begin{document}

\title{Identity Preserving 3D Head Stylization with Multiview Score Distillation}

\author{Bahri Batuhan Bilecen$^{1,2,3}$ \quad Ahmet Berke Gokmen$^{1,4}$ \quad Furkan Guzelant$^{1}$ \quad Aysegul Dundar$^{1}$\\
$^{1}$Bilkent University $^{2}$ETH Zurich $^{3}$Max Planck Institute\\$^{4}$INSAIT, Sofia University “St. Kliment Ohridski”\\
{\tt\small \{batuhan.bilecen@, berke.gokmen@ug., furkan.guzelant@, adundar@cs.\}bilkent.edu.tr}
}

\twocolumn[{%
\renewcommand\twocolumn[1][]{#1}%
\maketitle
\vspace{-1cm}
\begin{center}
    \centering
    \captionsetup{type=figure}
    \includegraphics[width=0.99\textwidth]{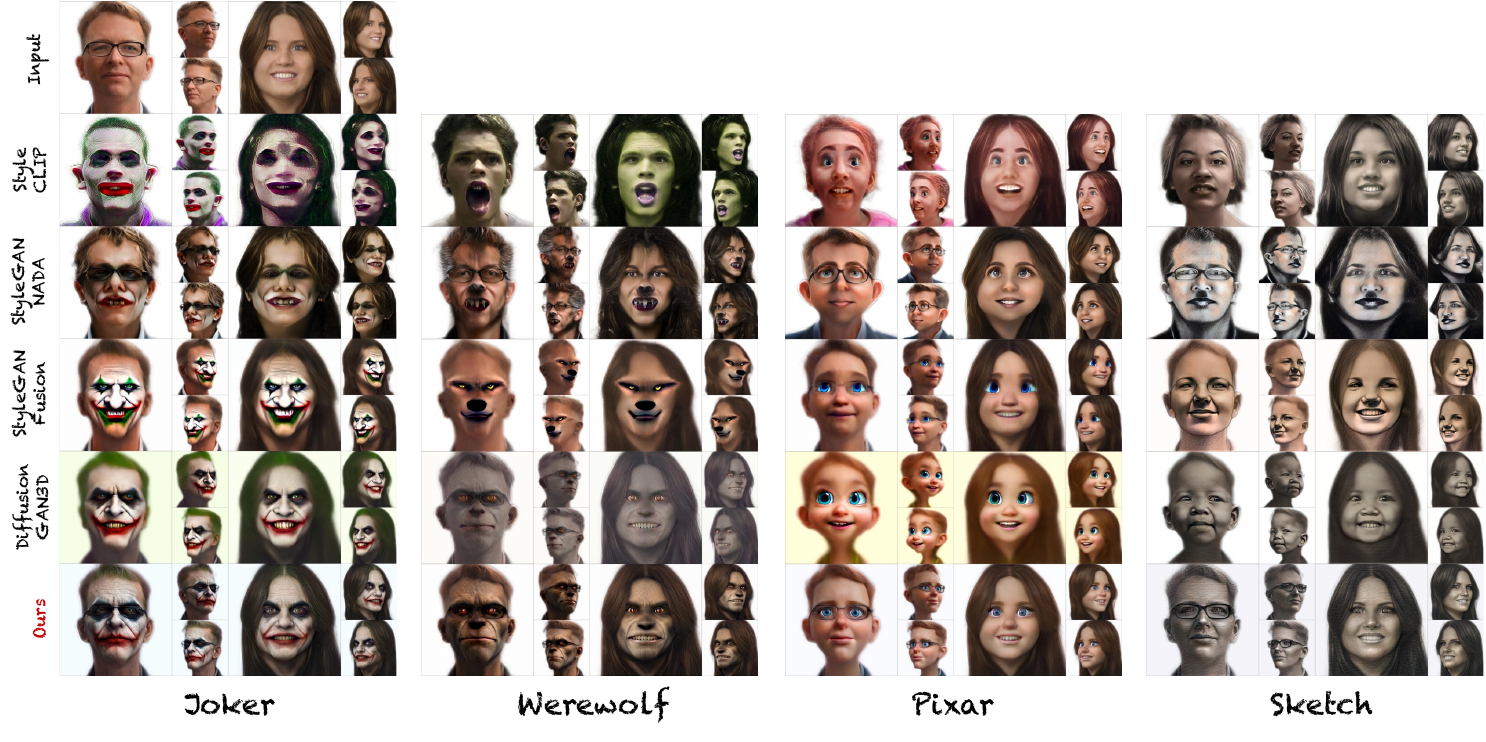}
    \captionof{figure}{Our method effectively balances stylization and identity preservation, whereas other approaches often struggle to achieve high-quality stylization and accurate identity retention. Notably, other methods generate identical faces for different inputs (e.g., Joker stylization), whereas ours preserves unique features for each input, ensuring distinct and personalized results for every individual face.}
    \label{fig:teaser}
\end{center}%
}]
\thispagestyle{plain}
\pagestyle{plain}

\begin{abstract}
\noindent3D head stylization transforms realistic facial features into artistic representations, enhancing user engagement across applications such as gaming and virtual reality. While 3D-aware generators have made significant advancements, many 3D stylization methods primarily provide near-frontal views and struggle to preserve the unique identities of original subjects, often resulting in outputs that lack diversity and individuality. Leveraging the PanoHead model which provides 360-degree consistent renders, we propose a novel framework that employs negative log-likelihood distillation (LD) to enhance identity preservation and improve stylization quality. By integrating multi-view grid score and mirror gradients within the 3D GAN architecture and introducing a score rank weighing technique, our approach achieves substantial qualitative and quantitative improvements. Our findings not only advance the state of 3D head stylization but also provide valuable insights into effective distillation processes between diffusion models and GANs, focusing on the critical issue of identity preservation. Please visit the \textbf{\underline{\href{https://three-bee.github.io/head_stylization}{project page}}}.
\end{abstract}
\vspace{-0.5cm}

\section{Introduction}

3D head stylization refers to transforming realistic facial and head features into artistic representations that can be rendered from multiple viewpoints. 3D head stylization has diverse applications in entertainment, augmented-reality, advertising, education, and creating relatable characters, making it a popular research topic.

Previously, face stylization was extensively explored in 2D image domains~\cite{yang2022pastiche, gal2022stylegan, Patashnik_2021_ICCV}, particularly through generative models like StyleGAN~\cite{karras2020analyzing}, which allowed for the manipulation of stylistic elements to create diverse, high-quality facial representations. With recent advancements in realistic 3D-aware face generators~\cite{chan2022efficient, an2023panohead}, research in stylization has now shifted to the 3D domain~\cite{abdal20233davatargan, kim2023datid, kim2023podia, zhang2023deformtoon3d, song2024agilegan3d, bai20243dpe, song2022diffusion, lei2023diffusiongan3d}, as these models enable the creation of lifelike representations that can be viewed from multiple angles, significantly enhancing user experiences.%

Most of these methods proposed for 3D stylization are built on the EG3D generator~\cite{chan2022efficient}, which primarily synthesizes near-frontal views. This limitation restricts the generation of 3D scenes from diverse viewpoints. 
In this work, we focus on PanoHead~\cite{an2023panohead}, which excels in synthesizing images from a 360-degree perspective. We establish applicable baselines using PanoHead, highlighting the challenges of stylizing 360-degree heads. Our findings demonstrate that most current methods fall short of effectively achieving cohesive stylization across all angles.
This limitation arises because many of these methods rely on stylized images that are primarily available in near-frontal views that fine-tuned StyleGAN models generate~\cite{abdal20233davatargan, song2024agilegan3d, zhang2023deformtoon3d}.

There are other works that leverage text-based image diffusion or CLIP models to either generate datasets or guide training~\cite{kim2023datid, kim2023podia, bai20243dpe, song2022diffusion, lei2023diffusiongan3d} providing more flexibility, especially for 3D generators.
Especially, Score Distillation Sampling (SDS)~\cite{poole2023dreamfusion} has demonstrated its effectiveness in %
stylization of 3D portraits~\cite{song2022diffusion, lei2023diffusiongan3d}, guiding the 3D representation to higher-density regions of the text-conditioned diffusion model. %
However, it has been noted that SDS can result in a loss of diversity in the generated outputs~\cite{lei2023diffusiongan3d}. For example, when the generator is fine-tuned for a Joker or Pixar style, the outputs from random samplings tend to exhibit significant similarity.
Even though DiffusionGAN3D~\cite{lei2023diffusiongan3d} introduces a relative distance loss to mitigate diversity loss, we find that its outputs, along with those from other competing methods, tend to be similar for different inputs. This results in challenges with \textit{identity preservation}, as illustrated in~\cref{fig:teaser}. The outputs tend to be similar across different inputs, resulting in a loss of the original identity.%

Based on this observation, we propose a framework that achieves 3D head stylization with identity preservation in this work. Specifically,
\begin{itemize}
    \item We propose using distillation with \textbf{negative log-likelihood distillation} (LD)~\cite{huang2024placiddreameradvancingharmonytextto3d} for domain adaptation of 3D-aware generators, yielding sharper and more ID-preserving results compared to SDS~\cite{poole2023dreamfusion}. Improving ID preservation has not been done before on GANs via changing distillation characteristics; instead, the focus was diverted to additional losses to solve the underlying issues of SDS~\cite{song2022diffusion, lei2023diffusiongan3d}.
    \item We propose \textbf{rank weighing for score tensors} on latent VAE channels to regularize the LD gradients, which we show achieve better input color and ID preservation in the distillation process. This introduces a new approach to control stylization via distillation.
    \item In compliance with the 3D GAN architecture~\cite{an2023panohead}, we extend LD with \textbf{multi-view grid and mirror score gradients} for improved stylization quality. Noting the importance of super-resolution (SR) networks in style-based GANs in domain adaptation, \textbf{we avoid distilling grid scores to SR layers} to further improve stylization.
\end{itemize}

Our method has shown significant qualitative and quantitative improvements over the relevant head stylization methods, presents good practices for distillation in one-step generators, and important insights into distillation from diffusion to GAN backbones.

\section{Related Work}

\noindent\textbf{3D-aware generators.}
Generative Adversarial Networks (GANs), when integrated with differentiable renderers, have made significant advancements in generating 3D-aware images consistent across multiple views. While earlier methods focus on mesh representations~\cite{pavllo2020convolutional, henderson2020leveraging, dundar2023fine, dundar2023progressive}, the latest innovations leverage implicit representations~\cite{chan2021pi, gu2021stylenerf, or2022stylesdf, niemeyer2021giraffe}.
Among implicit representations, triplane representations have become particularly popular due to their computational efficiency and the high quality of the generated outputs, %
like EG3D~\cite{chan2022efficient} and PanoHead~\cite{an2023panohead} which are reminiscent of the StyleGAN2 structure~\cite{karras2020analyzing}. These frameworks consist of mapping and synthesis networks that create triplanes, which are then projected into 2D images through volumetric rendering processes similar to those used in NeRF~\cite{mildenhall2021nerf}.
While EG3D is trained on the FFHQ dataset~\cite{karras2019style} with limited angle diversity, PanoHead allows for a 360-degree perspective in face generation, attributed to its dataset selection and model enhancements. In our work, we investigate PanoHead to achieve 3D portrait stylization.
We are interested in 3D portrait stylization of real, in-the-wild input images, which requires embedding images into PanoHead's latent space. 
This task is referred to as image inversion and extensively studied for StyleGAN generators \cite{abdal2019image2stylegan, roich2022pivotal,  alaluf2022hyperstyle, pehlivan2023styleres, yildirim2023diverse, yildirim2024warping} and recently for 3D-aware GAN models \cite{xie2023high, yuan2023make,bhattarai2024triplanenet, li2024generalizable, bilecen2024dual}.

\noindent\textbf{Adopting image generators to new domains.}
A popular method for stylizing 2D and 3D portraits involves fine-tuning generators initially trained to produce realistic faces. This process, known as domain adaptation of generators, %
can be accomplished using either a small number of samples for each style \cite{liu2020towards, ojha2021few} or by employing text-guided models to direct the training process \cite{kim2023datid, kim2023podia, bai20243dpe, song2022diffusion, lei2023diffusiongan3d}.
Using text-based models to guide stylization offers the advantage of generating styles based on various prompts, which can be challenging to sample directly. These models, trained on extensive datasets, effectively acquire knowledge in a disentangled manner, enabling them to produce meaningful results for intriguing prompts that may not have corresponding data available \cite{saharia2022photorealistic, podell2023sdxl, openai2023dalle3}.
Moreover, this guidance can be provided from multiple poses for 3D-aware generators, allowing for flexibility beyond the poses in the collected datasets.
While previously, CLIP loss has been a popular choice for the guidance \cite{mohammad2022clip, hong2022avatarclip, sanghi2022clip}, more recently, the Score Distillation Sampling (SDS) has demonstrated impressive performance \cite{poole2023dreamfusion, song2022diffusion, lei2023diffusiongan3d}.
However, previous research indicates that when generators are fine-tuned solely using SDS, they may collapse into producing a fixed image pattern, irrespective of the input noise \cite{song2022diffusion, lei2023diffusiongan3d}. To promote diversity, both StyleGANFusion \cite{song2022diffusion} and DiffusionGAN3D \cite{lei2023diffusiongan3d} introduce regularizers aimed at enhancing variety. However, we observe that this diversity remains constrained, leading to issues with identity preservation, where different inputs yield similar outputs. In this work, we propose a framework that enables 3D portrait stylization guided by text prompts while maintaining identity preservation.

\section{Method}

\begin{figure*}[t]
    \centering
    \includegraphics[width=1.0\linewidth]{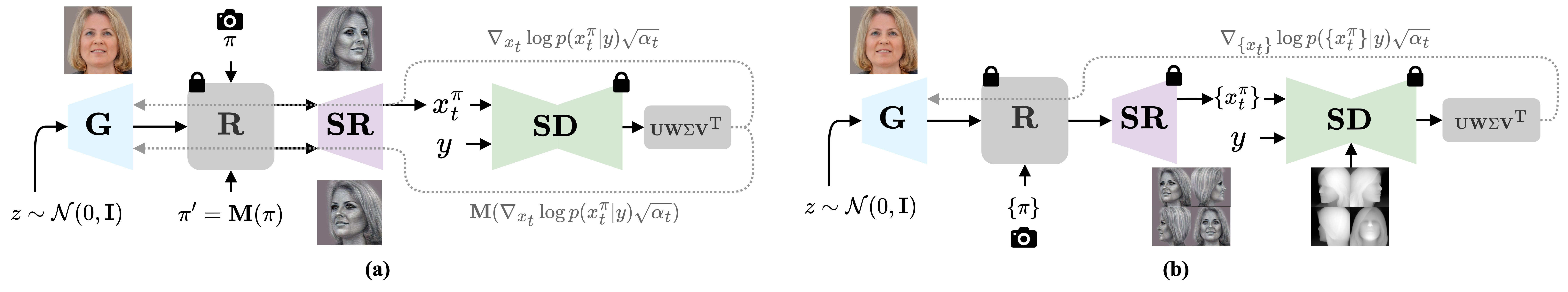}
    \caption{Our proposed training approach with mirror gradients \textbf{(a)} and grid distillation \textbf{(b)}. Dashed and non-dashed lines show backpropagation and forward paths. \textbf{G}, \textbf{R}, \textbf{SR}, and \textbf{SD} denote generator, neural renderer, super-resolver, and denoising UNet, respectively. Output of \textbf{SD} is the score, $\nabla_{x_t} \log p(x_t^\pi|y)$. We employ a depth-conditioned ControlNet in \textbf{(b)}. For the sake of simplicity, we omit CFG here.}
    \label{fig:method}
\end{figure*}

Our method involves efficient fine-tuning the pre-trained parameters of PanoHead for generating images from different domains.
This section will go through the development of the method.
Specifically, we will explain the adoption of LD to 3D-aware image generators, mention the differences between SDS, extend LD with mirror pose dependencies and grid denoising, and use rank weighing on score tensors.

\subsection{Preliminaries}
\noindent\textbf{Denoising diffusion probabilistic models (DDPM).} In the forward process of DDPM~\cite{NEURIPS2020_4c5bcfec}, Gaussian noise $\epsilon$ is gradually added to the initial data point $x_0$ over time steps $t$, reaching a pure Gaussian $\mathcal{N}$ as $t\rightarrow T$ (\cref{eqn:ddpm_forward}). Since we use a latent diffusion model~\cite{Rombach_2022_CVPR}, we denote image latent vectors with $x$ for simplicity. In the reverse process, the denoising UNet $\hat{\epsilon}$ is trained to predict the noise $\epsilon$ added at each time step $t$ with the MSE objective (\cref{eqn:ddpm_denoising}).
\begin{equation}
    \sqrt{\Bar{\alpha_t}}x_0 + \sqrt{1-\Bar{\alpha_t}}\epsilon = x_t, \quad \epsilon \sim \mathcal{N}(0,\mathbf{I})
    \label{eqn:ddpm_forward}
\end{equation}
\begin{equation}
\mathbb{E}_{x_0,\epsilon,t}[\ \Vert\epsilon-\hat{\epsilon}(x_t,t)\Vert^2 ]\
\label{eqn:ddpm_denoising}
\end{equation}

\noindent\textbf{Score-based stochastic differential equations (SDE).} In the score-based SDE framework~\cite{song2021scorebased}, the forward process~\cref{eqn:score_forward} adds Gaussian noise to $x$ over time, where $g(t)=\sqrt{1-\Bar{\alpha_t}}$ is the noise scale and $w$ is Wiener process. The reverse process, \cref{eqn:score_denoising}, is modeled as a reverse-time SDE and denoises $x$ by moving in the direction of higher probability density, where $\nabla_{x_t} \log p(x|y)$ is the \textit{score function} and $y$ is the condition signal (text prompt). We keep the drift terms $f=0$ for simpler notation.
\begin{equation}
    \text{d}x = g(t) \text{d}w
    \label{eqn:score_forward}
\end{equation}
\begin{equation}
    \text{d}x = -g(t)^2 \nabla_{x_t} \log p(x|y)\text{d}t + g(t)\text{d}\Bar{w}
    \label{eqn:score_denoising}
\end{equation}

The relation between the score function and the noise prediction can be described in~\cref{eqn:score_noise_conn}:

\begin{equation}
    \nabla_{x_t} \log p(x|y) \approx {-{g(t)}^{-1}\Hat{\epsilon}(x_t;t,y)}
    \label{eqn:score_noise_conn}
\end{equation}

\noindent\textbf{PanoHead.} PanoHead~\cite{an2023panohead} generates hybrid 3D representations called \textit{triplanes} with a StyleGAN-like architecture which takes 512-dimensional style vectors and outputs $256\times256$  resolution triplanes. These triplanes are rendered from a specified camera pose using a neural volumetric renderer to synthesize a 2D head image with size $64\times64$. Finally, a convolution-based super-resolution (SR) network performs upscaling from $64\times64$ images to $512\times512$ final images to refine details. Since the SR network upscales significantly, it plays a critical role in maintaining geometric consistency while prioritizing a rich texture and color.

\noindent\textbf{Score distillation sampling (SDS).} SDS~\cite{poole2023dreamfusion} leverages pre-trained 2D diffusion models to guide the generation of content in other domains, such as 3D, shown in~\cref{eqn:sds_eqn}:

\vspace{-0.5cm}
\begin{equation}
\nabla_\theta L_{\text{SDS}}(\phi, x = g(\theta)) \triangleq \mathbb{E}_{t, \epsilon}\big[\omega(t)(\hat{\epsilon}(x_t; y, t) - \epsilon)\frac{\partial x}{\partial \theta}\big]
\label{eqn:sds_eqn}
\end{equation}

\noindent where \(\omega(t)\) weights the timestep contribution, $\theta$ is the 3D representation, and $g$ is the differentiable renderer to project 3D to 2D. Intuitively, SDS guides $\theta$ toward higher-density regions in the diffusion model's learned distribution.

\subsection{ID-Preserving Stylization with Distillation}

\noindent\textbf{Likelihood distillation (LD) objective.} We partially adapt the distillation procedure in~\cite{huang2024placiddreameradvancingharmonytextto3d} and tailor it to GANs, explained in this section. A detailed derivation is provided in Supplementary for completeness.

Assume that distribution ($q$) of the 3D representation ($\theta$) conditioned on text prompt ($y$) is proportional to the prompt-conditioned distribution ($p$) of independent 2D renders ($x_0^i$) on different poses ($i$):
\begin{equation}
    q(\theta|y) \propto p(x_0^0,x_0^1,...,x_0^N|y) = \prod_i^N p(x_0^i|y)
    \label{eqn:initial_dists}
\end{equation}
We optimize negative log-likelihood of ~\cref{eqn:initial_dists} to find $\theta$:
\begin{equation}
    -\log q(\theta|y) = -\log \prod_i^N p(x_0^i|y) = - \sum_i^N \log p(x_0^i|y)
    \label{eqn:nll}
\end{equation}
Define the loss $\mathcal{L}_\text{LD}$ and find gradient $\nabla_\theta$ to update $\theta$ via gradient descent:
\begin{equation}
\nabla_\theta \mathcal{L}_\text{LD} = -\mathbb{E}_\pi\{\nabla_\theta \log p(x_0^\pi|y)\}
\label{eqn:loss_sd}
\end{equation}
Combining~\cref{eqn:loss_sd,eqn:ddpm_forward} results in:
\begin{equation}
\nabla_\theta \log p(x_0^\pi|y) = \nabla_{x_t} \log p(x_t^\pi|y) \frac{\partial x_t^\pi}{\partial x_0^\pi} \frac{\partial x_0^\pi}{\partial \theta}
\label{eqn:score}
\end{equation}
where $\nabla_{x_t} \log p(x_t^\pi|y)$ is the score function estimation. Plugging~\cref{eqn:score} into~\cref{eqn:loss_sd} yields the update direction:
\begin{equation}
    \nabla_\theta \mathcal{L}_\text{LD} = -\mathbb{E}_{\pi,x_t}\{ \nabla_{x_t} \log p(x_t^\pi|y) \frac{\partial x_t^\pi}{\partial x_0^\pi} \frac{\partial x_0^\pi}{\partial \theta} \}
    \label{eqn:nabla_theta_ls}
\end{equation}
where $\frac{\partial x_t^\pi}{\partial x_0^\pi}$ is $\sqrt{\Bar{\alpha_t}}$ from~\cref{eqn:ddpm_forward}. Notice that to update $\theta$, we do not need to back-propagate through the denoiser and can acknowledge the output as a part of the gradient. For this task, we assign $\theta$ as the PanoHead layers.

\noindent\textbf{SDS vs LD on GANs.} Note that LD is different than SDS~\cite{poole2023dreamfusion} as it optimizes negative log-likelihood instead of reverse KL-divergence. Compliant with previous observations~\cite{wang2023prolificdreamer}, we also observe that SDS yields blurry results, as LD is sharp in domain adaptation tasks. We attribute this to the fact that SDS is inherently mode-seeking due to the subtraction of ground truth noise $\epsilon$ from the estimated noise $\hat{\epsilon}$ and high classifier free guidance (CFG)~\cite{ho2022classifierfreediffusionguidance} weight to avoid divergence~\cite{poole2023dreamfusion}. In contrast, LD is diversity-seeking (by not utilizing $\epsilon$), does not require very high CFG weights, and is more suitable for our task considering that we utilize a GAN prior. \textit{This issue had not been investigated before on GANs, and the focus was diverted to additional losses to solve the underlying issues of SDS, e.g.~\cite{song2022diffusion,lei2023diffusiongan3d}}.

\begin{figure}[ht!]
\centering
\renewcommand{\arraystretch}{0.0}
\setlength{\tabcolsep}{0.1pt}
\footnotesize
\begin{tabular}{ccccccccc}
& \multicolumn{2}{c}{Input} & \multicolumn{2}{c}{LD} &\multicolumn{2}{c}{PlacidDreamer} & \multicolumn{2}{c}{Ours}  \\
\rotatebox{90}{\quad \scriptsize Wolf}&
\includegraphics[width=0.115\linewidth]{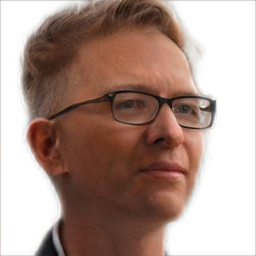}& 
\includegraphics[width=0.115\linewidth]{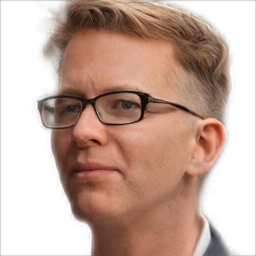}& 
\includegraphics[width=0.115\linewidth]{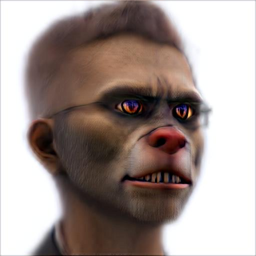}& 
\includegraphics[width=0.115\linewidth]{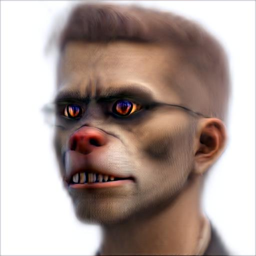}& 
\includegraphics[width=0.115\linewidth]{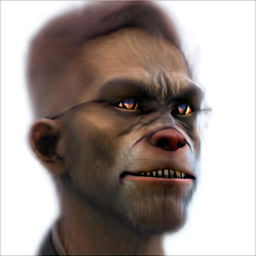}& 
\includegraphics[width=0.115\linewidth]{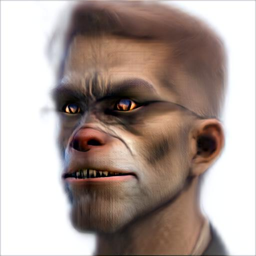}& 
\includegraphics[width=0.115\linewidth]{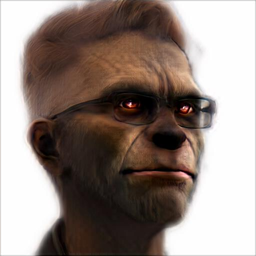}& 
\includegraphics[width=0.115\linewidth]{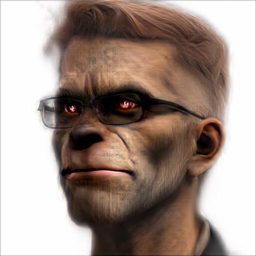}
\\
\rotatebox{90}{\quad \scriptsize Joker}&
\includegraphics[width=0.115\linewidth]{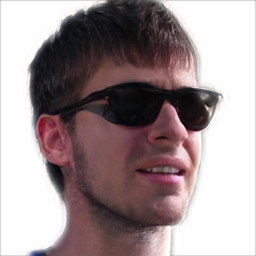}& 
\includegraphics[width=0.115\linewidth]{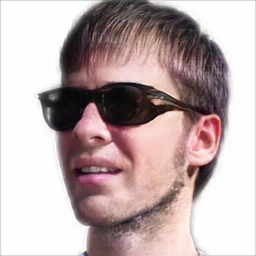}& 
\includegraphics[width=0.115\linewidth]{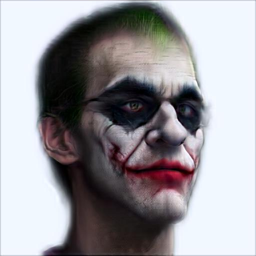}& 
\includegraphics[width=0.115\linewidth]{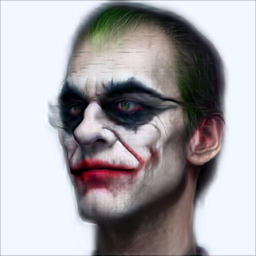}& 
\includegraphics[width=0.115\linewidth]{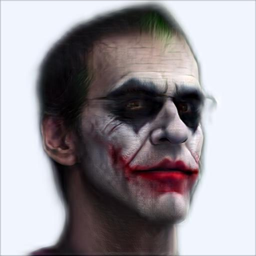}& 
\includegraphics[width=0.115\linewidth]{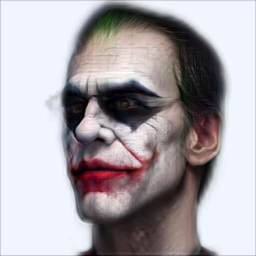}& 
\includegraphics[width=0.115\linewidth]{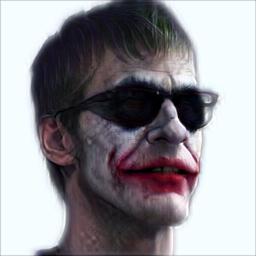}& 
\includegraphics[width=0.115\linewidth]{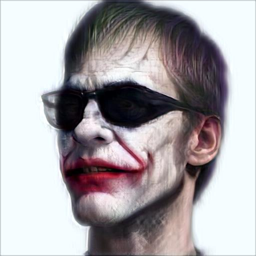}
\\
\rotatebox{90}{\quad \scriptsize Pixar}&
\includegraphics[width=0.115\linewidth]{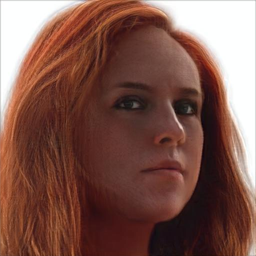}& 
\includegraphics[width=0.115\linewidth]{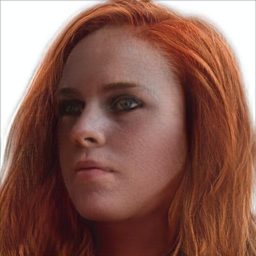}& 
\includegraphics[width=0.115\linewidth]{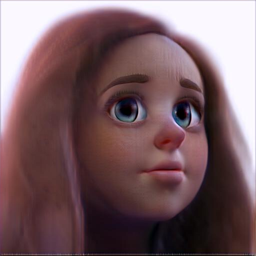}& 
\includegraphics[width=0.115\linewidth]{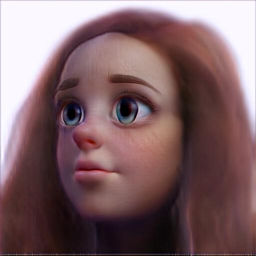}& 
\includegraphics[width=0.115\linewidth]{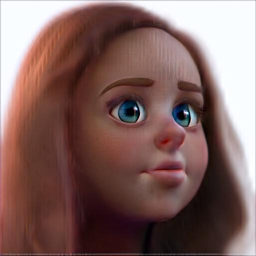}& 
\includegraphics[width=0.115\linewidth]{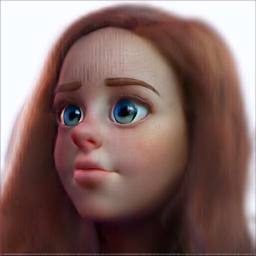}& 
\includegraphics[width=0.115\linewidth]{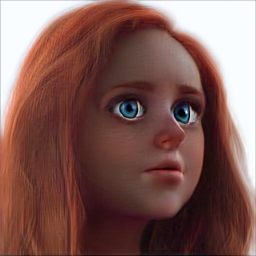}& 
\includegraphics[width=0.115\linewidth]{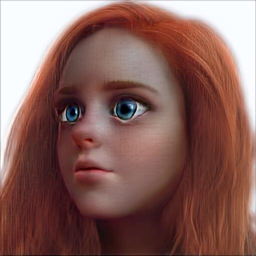}
\end{tabular}   
\vspace{-0.25cm}
\caption{Difference between LD, PlacidDreamer's additions on LD, and our additions on LD.}
\label{fig:comp-ld}
\vspace{-0.25cm}
\end{figure}

We also note that~\cite{huang2024placiddreameradvancingharmonytextto3d} proposes multiple-gradient descent on top of~\cref{eqn:nabla_theta_ls}, which we simply omit since the pretrained GAN backbone is strong enough to avoid divergence. However, for the sake of completeness, we also compare it with LD in~\cref{fig:comp-ld}, along with our full pipeline. In our comparisons, LD and PlacidDreamer alter expressions (e.g., opening a closed mouth in~\cref{fig:comp-ld}, row 1, or removing visible teeth, row 2), misrepresent accessories like glasses, and distort features such as hair in stylized domains. Our full pipe preserves these details more accurately.

\begin{figure}[t!]
\centering
\footnotesize
\setlength{\tabcolsep}{0.1pt}
\scalebox{0.95}{
\begin{tabular}{cccc}
\multicolumn{2}{c}{Input} & \multicolumn{2}{c}{{LD + \cite{Zhang_2023_ICCV} (full-rank)}} 
\\
\includegraphics[width=0.24\linewidth]{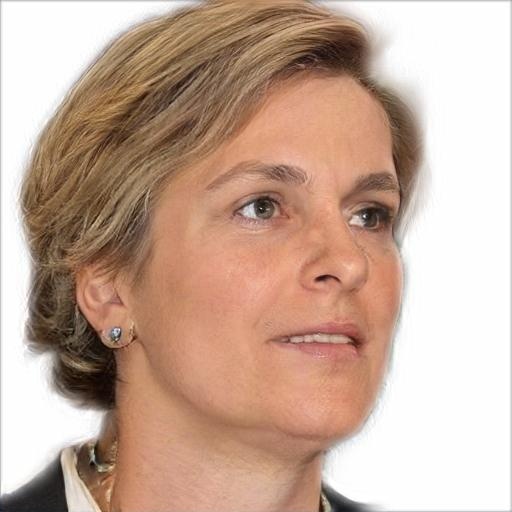}&
\includegraphics[width=0.24\linewidth]{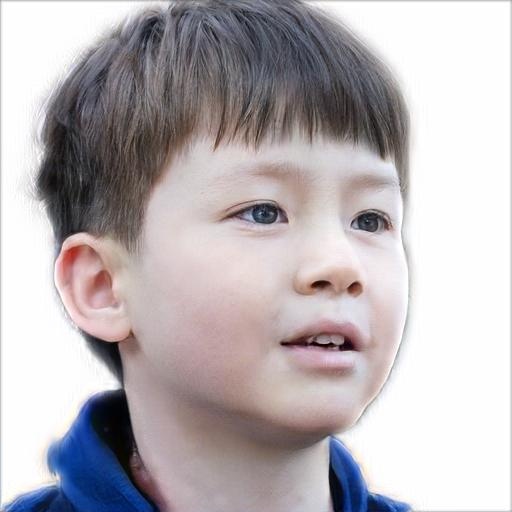}&
\includegraphics[width=0.24\linewidth]{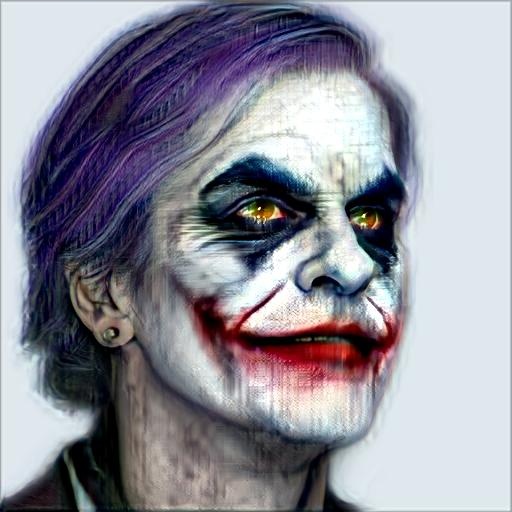}&
\includegraphics[width=0.24\linewidth]{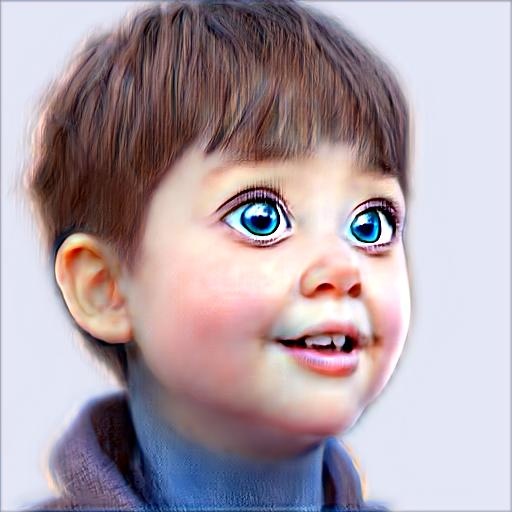}
\\
\multicolumn{2}{c}{{LD + \cite{Zhang_2023_ICCV} (weighted-rank)}} & \multicolumn{2}{c}{\textbf{Ours}}
\\
\includegraphics[width=0.24\linewidth]{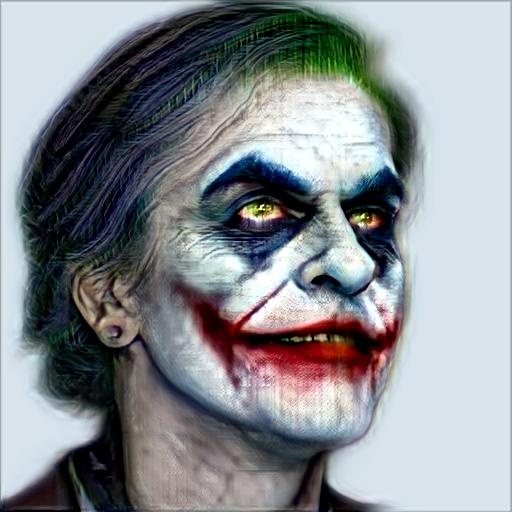}&
\includegraphics[width=0.24\linewidth]{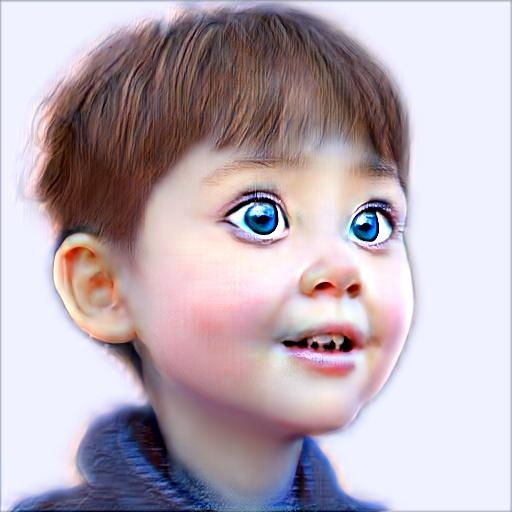}&
\includegraphics[width=0.24\linewidth]{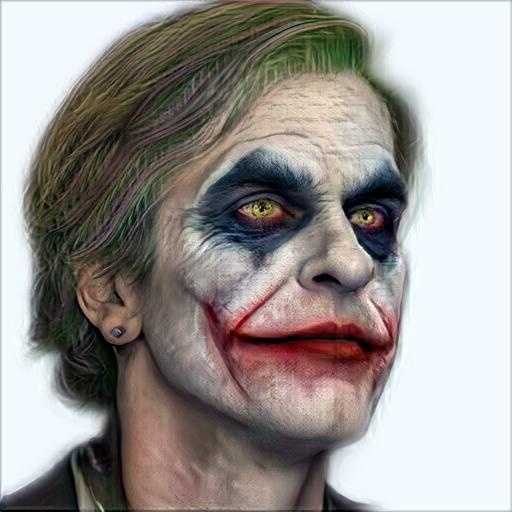}&
\includegraphics[width=0.24\linewidth]{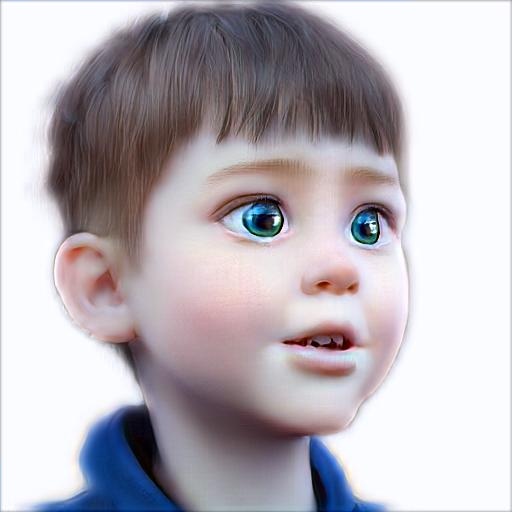}
\\
\begin{minipage}{0.24\linewidth}
    \centering
    \includegraphics[width=0.48\linewidth]{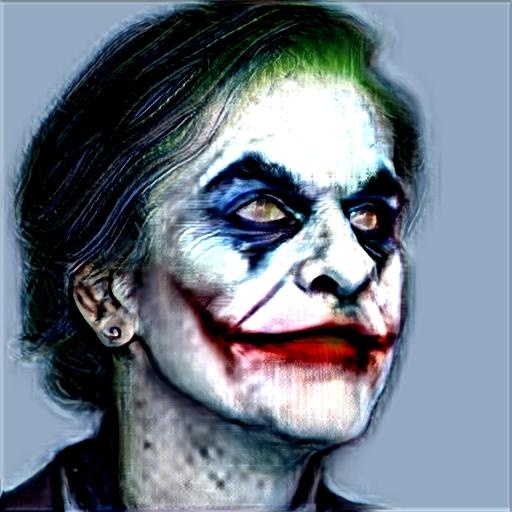}
    \includegraphics[width=0.48\linewidth]{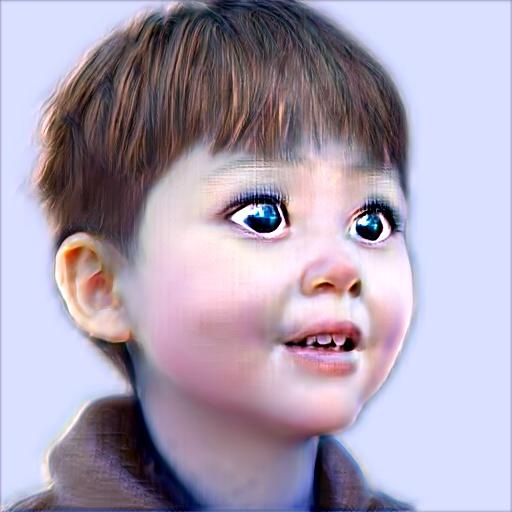}
\end{minipage}\hspace{0pt} &
\begin{minipage}{0.24\linewidth}
    \centering
    \includegraphics[width=0.48\linewidth]{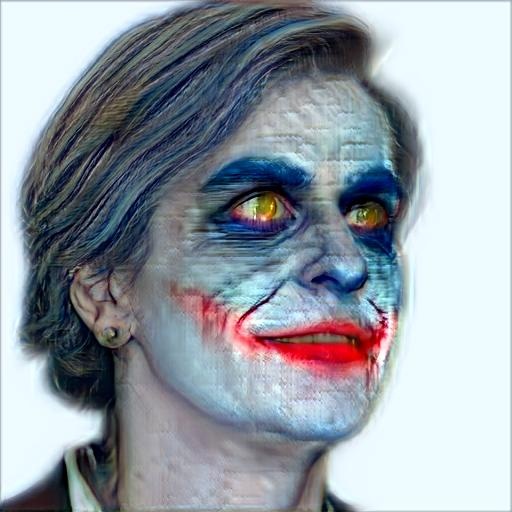}
    \includegraphics[width=0.48\linewidth]{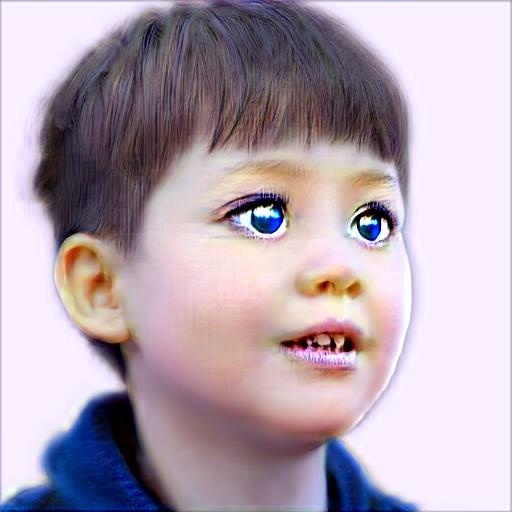}
\end{minipage}\hspace{0pt} &
\begin{minipage}{0.24\linewidth}
    \centering
    \includegraphics[width=0.48\linewidth]{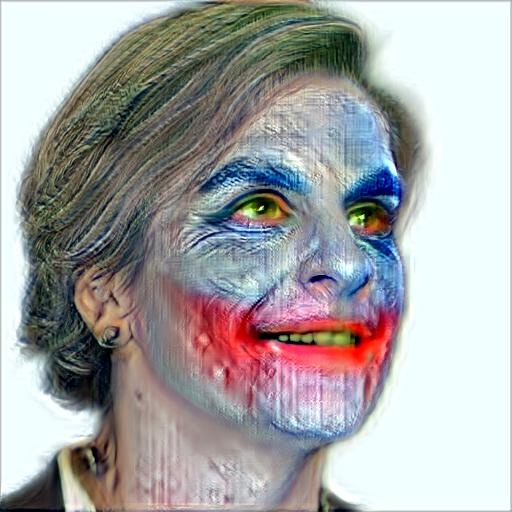}
    \includegraphics[width=0.48\linewidth]{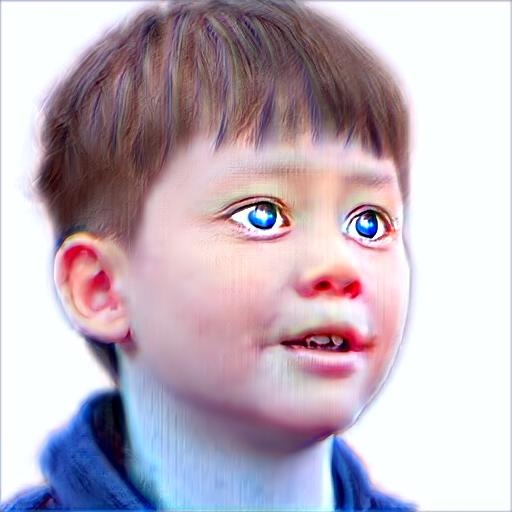}
\end{minipage}\hspace{0pt} &
\begin{minipage}{0.24\linewidth}
    \centering
    \includegraphics[width=0.48\linewidth]{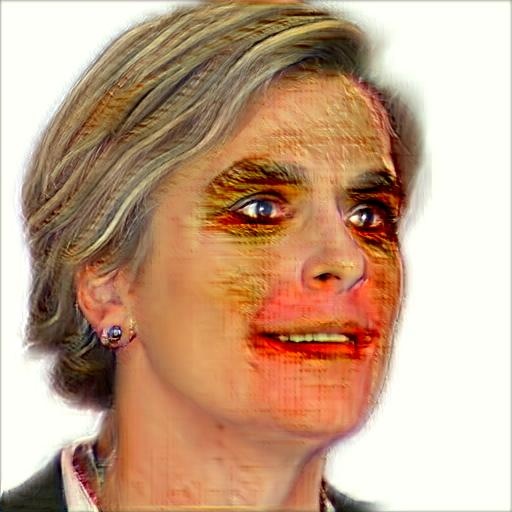}
    \includegraphics[width=0.48\linewidth]{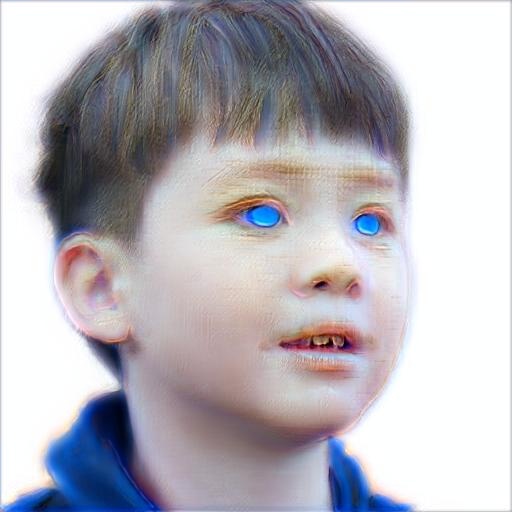}
\end{minipage}
\\
k=0 & k=1 & k=2 &  k=3
\end{tabular} 
}
\caption{Ablation study on rank weighing. After SVD, four rank-1 matrices are obtained, and $\text{k}^\text{th}$ are chosen for reconstruction. Weighted rank improves the results compared to full rank. We also present final results (ours) for comparisons with LD + weighted rank and our multi-view and mirror gradient.}
\vspace{-0.5cm}
\label{fig:rank}
\end{figure}

\noindent\textbf{Rank weighted score tensors.} While LD reduces the smoothness issues, we may come across some artifacted results, especially around the hair, ears, and neck on some prompts (\cref{fig:rank}, full-rank). 
Notice that the artifacts are more focused on incorrect color distribution rather than the style itself. Since diffusion latents must also contain color distribution information~\cite{sdxl_color}, we decide to investigate the SVD of the score tensors along the VAE latent channel (4) dimension. {We notice that the first rank contains most of the stylization and surpassing the contributions of the lower three ranks can mitigate undesired tints.} Hence, we re-weigh the diagonal singular value matrix \textbf{S} with linearly decaying coefficients \textbf{W} from the largest singular value to the smallest. Specifically,

\begin{equation}
\begin{aligned}
    \mathbf{U}\mathbf{\Sigma}\mathbf{V^T} = \text{SVD}(\nabla_\theta \log p(x_0^\pi|y))
\\
\nabla_\theta \log \Tilde{p}(x_0^\pi|y) = \mathbf{U}\mathbf{W}\mathbf{\Sigma}\mathbf{V^T}
\end{aligned}
\end{equation}

$\Tilde{p}$ is the rank-weighted score distribution, $\mathbf{\Sigma}_\text{4x4}=\text{diag}(\sigma_1, \sigma_2, ..., \sigma_4)$, $\mathbf{U}_\text{4x4}=[u_1, u_2, ..., u_4]$ , $\mathbf{V}_\text{4096x4}=[v_1, v_2, ..., v_4]$, and $\mathbf{W}=\text{diag}(1,0.75,0.5,0.25)$.

\begin{figure}
\centering
\setlength{\tabcolsep}{0.1pt}
\footnotesize
\begin{tabular}{cccccccccc}
& Input & \multicolumn{2}{c}{SD1.5} & \multicolumn{2}{c}{RV3.0} & \multicolumn{2}{c}{RV5.1} & \multicolumn{2}{c}{SDXL} \\
\rotatebox{90}{~~~\scriptsize Pixar}&
\includegraphics[width=0.105\linewidth]{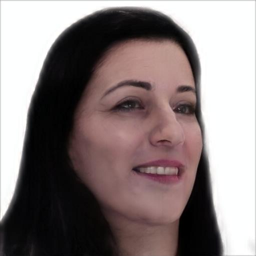}& 
\includegraphics[width=0.105\linewidth]{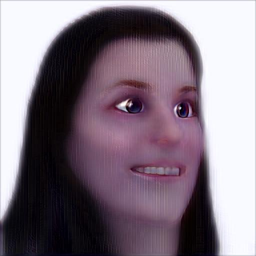}& 
\includegraphics[width=0.105\linewidth]{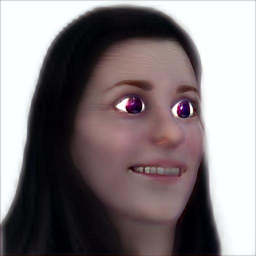}& 
\includegraphics[width=0.105\linewidth]{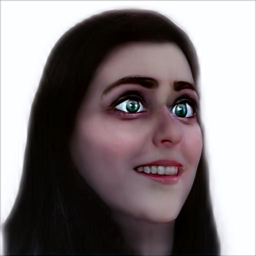}& 
\includegraphics[width=0.105\linewidth]{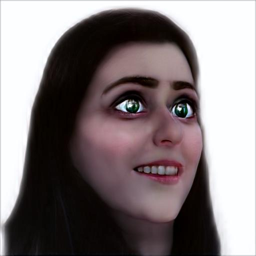}& 
\includegraphics[width=0.105\linewidth]{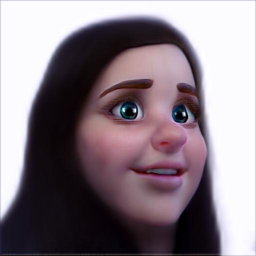}& 
\includegraphics[width=0.105\linewidth]{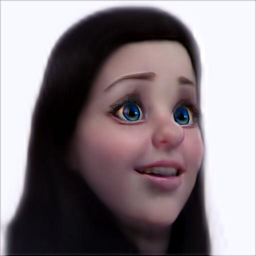}& 
\includegraphics[width=0.105\linewidth]{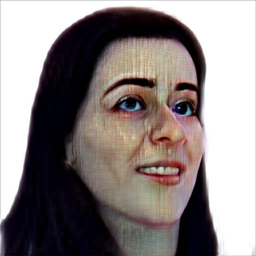}& 
\includegraphics[width=0.105\linewidth]{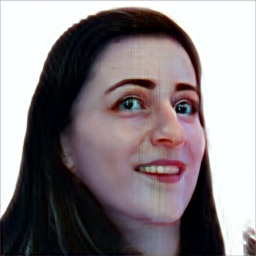} 
\\
\rotatebox{90}{~~\scriptsize Statue}&
\includegraphics[width=0.105\linewidth]{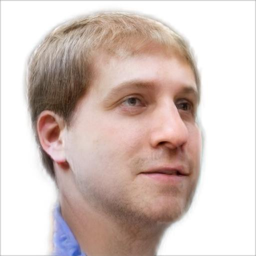}& 
\includegraphics[width=0.105\linewidth]{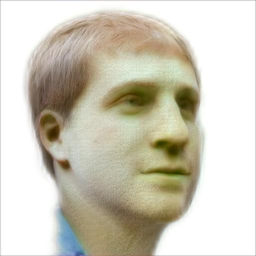}& 
\includegraphics[width=0.105\linewidth]{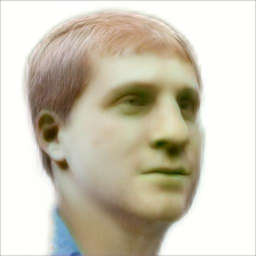}& 
\includegraphics[width=0.105\linewidth]{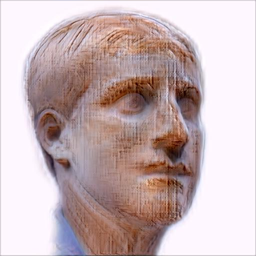}& 
\includegraphics[width=0.105\linewidth]{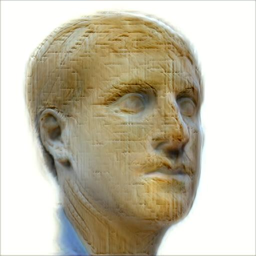}& 
\includegraphics[width=0.105\linewidth]{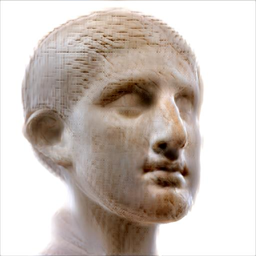}& 
\includegraphics[width=0.105\linewidth]{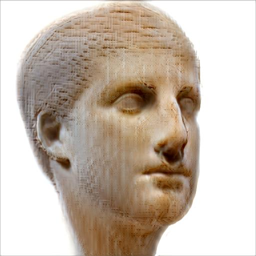}& 
\includegraphics[width=0.105\linewidth]{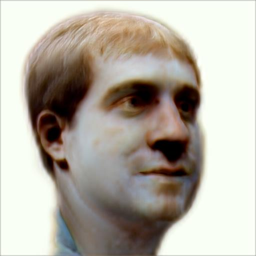}& 
\includegraphics[width=0.105\linewidth]{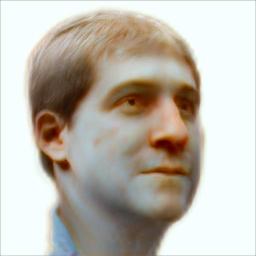}

\end{tabular}
\vspace{-0.25cm}
\caption{Distillation from different latent spaces, with default \textbf{W} and (left) weighted $\mathbf{W}=\text{diag}(1,0.75,0.5,0.25)$ (right).}
\label{fig:comp-dist-sds}
\vspace{-0.5cm}
\end{figure}

As seen in~\cref{fig:rank}, rank-weighted scores eliminate the problematic tints and set a good baseline for the following improvements.  
We set the weight scores based on our empirical analysis, and use the same weights for all prompts. We also show that our weighing generalizes to other latent spaces in~\cref{fig:comp-dist-sds}, revealing that emphasizing the largest rank-1 matrix via weighted \textbf{W} yields more consistent stylization. More results are in Supplementary.

\noindent\textbf{Extending LD via cross-dependencies with mirror poses.} The expression in~\cref{eqn:nabla_theta_ls} does not consider the correlation between different views. To add cross-dependencies, we account for another poses where $\pi' \neq \pi$ in~\cref{eqn:cross_2}:

\begin{equation}
\begin{aligned}
    \nabla_\theta \mathcal{L}_\text{LD} = -\mathbb{E}_{\pi,x_t}\{ \nabla_{x_t} \log p(x_t^\pi|y) \frac{\partial x_t^\pi}{\partial x_0^\pi} \frac{\partial x_0^\pi}{\partial \theta} \\ +\sum_{\pi \neq \pi'} \nabla_{x_t} \log p(x_t^\pi|y) \frac{\partial x_t^\pi}{\partial x_0^{\pi'}} \frac{\partial x_0^{\pi'}}{\partial \theta} \}
    \label{eqn:cross_2}
\end{aligned}
\end{equation}
By utilizing the symmetry prior for human heads, we assume that if $\pi$ and $\pi'$ are yaw-symmetric camera matrices, then $x_t^\pi = \mathbf{M}(x_t^{\pi'})$ where $\mathbf{M}$ is the vertical mirror (flip) operator. Then, by~\cref{eqn:ddpm_forward}, $\frac{\partial x_t^\pi}{\partial x_0^{\pi'}}$ simply becomes $\mathbf{M}\sqrt{\Bar{\alpha_t}}$.
Further simplifying the expression~\cref{eqn:cross_2} yields:
\begin{equation}
\begin{aligned}
    \nabla_\theta \mathcal{L}_\text{LD} = -\mathbb{E}_{\pi,x_t}\{ \nabla_{x_t} \log p(x_t^\pi|y) \sqrt{\Bar{\alpha_t}}(\frac{\partial x_0^\pi}{\partial \theta}+\mathbf{M} \frac{\partial x_0^{\pi'}}{\partial \theta}) \}
\end{aligned}
\end{equation}
As shown in~\cref{fig:method} (a), this intuitively means that we utilize the same score estimation for mirror poses but must also mirror the gradients while back-propagating to the generator. Since 3D-consistent face domain editing with 2D-diffusion models is challenging, keeping the same score for different renders of the same ID under logical constraints avoids deviating from the convergence path. This is because $\nabla_{x_t} \log p(x_t^\pi|y)$ and $\nabla_{x_t} \log p(x_t^{\pi'}|y)$ indicating the same direction for the same $y$ is not guaranteed. 

Notice the mirror approach can be extended to any render pose tuple ($\pi, \pi'$) as long as we can find a tractable gradient chain for $\frac{\partial x_t^\pi}{\partial x_0^{\pi'}}$. Our experiments show that mirror gradients further improve stylization while accentuating 3D-aware features like glasses (\cref{fig:ablation_mirror_grid}).

\noindent\textbf{Multi-view distillation via grid denoising.} Mirror gradients only correlate yaw-symmetric poses. Now, let us consider the joint probability distribution $p$ and not assume explicit independence among poses, yielding~\cref{eqn:grid}:
\begin{equation}
\begin{aligned}
q(\theta|y) \propto p(x_0^0,x_0^1,...,x_0^N|y)=p(\{x_0^{\pi}\}|y) \\
\nabla_\theta \mathcal{L}_{\text{LD}_g} = -\mathbb{E}_{\pi,\{x_t\}}\{ \nabla_{\{x_t\}} \log p(\{x_t^{\pi}\}|y) \frac{\partial \{x_t^{\pi}\}}{\partial \{x_0^{\pi}\}} \frac{\partial \{x_0^{\pi}\}}{\partial \theta} \}
\end{aligned}
\label{eqn:grid}
\end{equation}
where $\{x_0^{\pi}\}$ is defined to have all poses $\{0,1,...,N\}$. However, this is not computationally feasible. {Instead, as shown in~\cref{fig:method} (b), we approximate $\{x_0^{\pi}\}$ with a 2$\times$2 grid, where each element is a $x_0^\pi$ with different render pose $\pi$.} This way, denoising UNet can correlate between different renders of $\theta$, improving stylization consistency across views. We further employ a depth-conditioned ControlNet~\cite{Zhang_2023_ICCV} to ensure the estimated score does not collapse to a single fused image. We also leverage CFG~\cite{ho2022classifierfreediffusionguidance} on each backpropagation path in~\cref{fig:method}, but omit from figure for simplicity.

Our approach does not fine-tune any diffusion model to accommodate multiple inputs~\cite{Liu_2023_ICCV,yang2023dreamcomposer,liu2023syncdreamer} for multi-view consistency. Instead, it can use any pre-trained diffusion model and re-purposes grid structures for distillation task, first proposed for video generation~\cite{lee2024grid} and editing~\cite{kara2024rave} with spatiotemporal consistency.

\noindent\textbf{Effect of super-resolution network in style-based 3D GANs.}  The PanoHead model generates images at a resolution of $512 \times 512$ \cite{an2023panohead}, and the diffusion model similarly processes $512 \times 512$ images when using LD~\cite{Rombach_2022_CVPR}. For grid construction, we arrange four images in a $2 \times 2$ layout but reduce the individual image size to $256 \times 256$ to avoid memory issues during forward passes. This way, the final grid image has the size of  $512 \times 512$. Although this maintains a good correlation between poses and consistent stylization, it creates a resolution mismatch when gradients are directly propagated from the PanoHead output. 
To resolve this, we experiment with feeding the gradients before the SR network, where the renderer outputs images at a lower resolution of $64 \times 64$. This adjustment is only necessary for multi-view distillation, where we backpropagate $\nabla_\theta \mathcal{L}_{\text{LD}_g}$ before the SR layers. For mirror poses, we do not need to skip the super-resolution layers since they match the correct resolution, as visualized in~\cref{fig:method}. This strategy improves the stylization quality and reduces unwanted artifacts, such as blur, over-saturation, and color tint, seen in~\cref{fig:ablation_mirror_grid}.

\begin{figure*}[t!]
    \centering
    \includegraphics[width=1.0\linewidth]{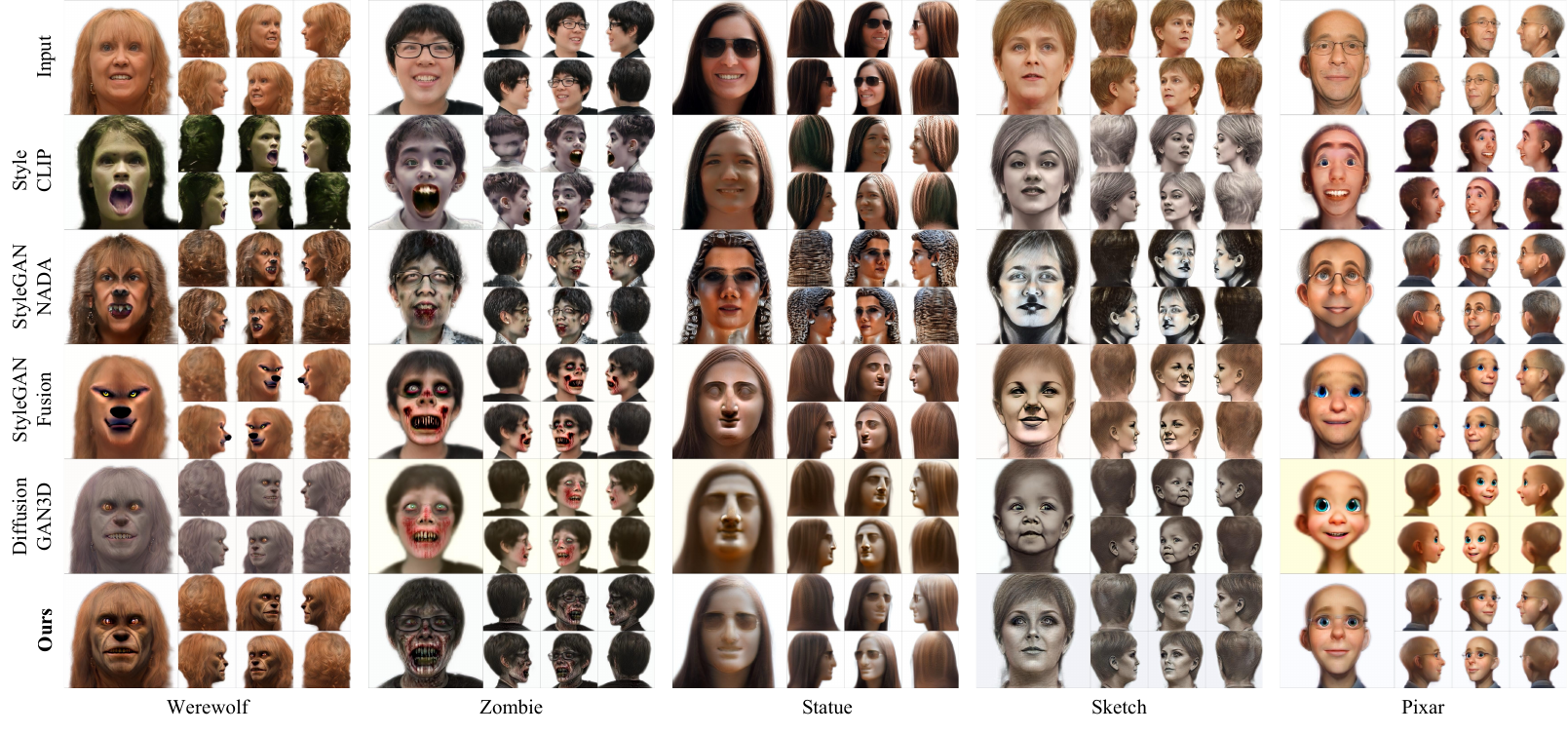}
    \caption{Qualitative stylization results of 3D domain-adaptation methods, provided in 360-degree views.}
    \vspace{-0.25cm}
    \label{fig:results_visual}
    \vspace{-0.25cm}
\end{figure*}

\section{Experiments}

\noindent \textbf{Baselines.}
We include various domain adaptation methods for our comparisons and adopt all of them for the  PanoHead generator. StyleCLIP~\cite{Patashnik_2021_ICCV} trains a $\mathcal{W}^+$ mapper network with the CLIP loss. StyleGAN-NADA~\cite{gal2022stylegan} employs the CLIP objective to train adaptively-selected generator layers rather than modifying $\mathcal{W}^+$. StyleGANFusion~\cite{song2022diffusion} utilizes SDS and a directional regularizer via the frozen generator for more stable distillation. DiffusionGAN3D~\cite{lei2023diffusiongan3d} further improves the regularization of~\cite{song2022diffusion} by employing a relative distance loss.
After tuning the generators, for inference, we invert images by optimizing $\mathcal{W}^+$ to reconstruct the input image with the original PanoHead generator. The optimized $\mathcal{W}^+$ is then fed to these different domain-adapted generators, and rendering from various views is outputted.
We also include 2D-3D editing methods such as InstantID~\cite{wang2024instantid}, InstructPix2Pix~\cite{brooks2022instructpix2pix} and InstructNerf2Nerf~\cite{instructnerf2023} for a broader comparison.
Details are given in Supplementary.

\noindent \textbf{Training setup.} We train the generator with synthetic $z_{1\times512}\sim\mathcal{N}(0,\mathbf{I})$ data for 10k iterations with batch size 1, where the truncation parameter of the generator's mapping network is $\psi=0.8$. The optimizer is Adam with a learning rate $1e^{-4}$. The CFG weight and  ControlNet guidance weight are set to 7.5 and 1.0, respectively. Depth ground truths are extracted from~\cite{depth_anything_v2}. More details of the training recipe are provided in Supplementary.

\noindent \textbf{Metrics.}
For stylization performance, we measure Fréchet Inception Distance (FID)~\cite{heusel2017gans} and CLIP embedding similarity; for ID preservation, we measure ArcFace-based~\cite{deng2019arcface} ID similarity and multi-view render depth $\mathcal{L}_2$ difference $\Delta\mathcal{D}$. To construct our test set, we perform $\mathcal{W}^+$ inversion to randomly chosen $\sim$100 FFHQ images.
These $\mathcal{W}^+$'s are input to each generator to stylize a real identity. 

Specifically, in FID and CLIP, we construct two different edited image distributions, wherein one, the images are edited with a  Stable Diffusion pipeline in the 2D image domain (ground-truth distribution), and in the other, the images are edited via the domain-adapted 3D-aware generators.
We provide the example images from the ground-truth distribution in Supplementary.
We calculate FID between those two image distributions and CLIP scores between paired images. In ID and $\Delta\mathcal{D}$, we use ground truth unedited images and calculate the scores between the ground-truth original images and stylized images.

\noindent \textbf{Results.}
 We provide quantitative and qualitative results in~\cref{table:results,fig:teaser,fig:results_visual,fig:rebuttal_ip2p_in2n}, respectively. 
The quantitative results demonstrate that our method outperforms others in terms of stylization quality, as measured by FID and CLIP scores, as well as in identity preservation, as evaluated by ID similarity and $\Delta\mathcal{D}$ across nearly all prompts. These numerical findings are consistent with the visual assessment.
As shown in~\cref{fig:teaser,fig:results_visual}, StyleCLIP and StyleGAN-NADA produce stylizations of lower quality with noticeable artifacts. While StyleGANFusion and DiffusionGAN3D are capable of producing stylized images, they significantly compromise identity preservation. For instance, in the Joker example in~\cref{fig:teaser}, both methods generate identical facial features for two different identities. Similarly, for the sketch prompt, the identities are substantially altered.
Competing methods such as InstantID and IP2P also remove accessories such as eyeglasses and alter identity heavily on different views (\cref{fig:rebuttal_ip2p_in2n}). 
In contrast, ours effectively preserves the identity while delivering high-quality stylizations.

\definecolor{color1}{rgb}{0.478,0.722,0.42} %
\definecolor{color2}{rgb}{0.528,0.752,0.47}
\definecolor{color3}{rgb}{0.75, 0.87, 0.75}
\definecolor{color4}{rgb}{0.90, 0.95, 0.90}
\definecolor{color5}{rgb}{0.93, 0.97, 0.93}
\definecolor{color6}{rgb}{0.97, 0.99, 0.97}
\definecolor{color7}{rgb}{1.0, 1.0, 1.0}

\begin{table*}[t!]
\footnotesize
\centering
\setlength{\tabcolsep}{2pt}
\renewcommand{\arraystretch}{1.0}
\begin{tabular}{@{}rrcccccccccccccccccccc@{}}
 & & \multicolumn{4}{c}{Pixar}
   & \multicolumn{4}{c}{Joker}
   & \multicolumn{4}{c}{Werewolf}
   & \multicolumn{4}{c}{Sketch}
   & \multicolumn{4}{c}{Statue} \\
\cmidrule(l){3-6} \cmidrule(l){7-10} \cmidrule(l){11-14} \cmidrule(l){15-18} \cmidrule(l){19-22}
 & & {FID} & {CLIP} & {ID} & {$\Delta\mathcal{D}$}
   & {FID} & {CLIP} & {ID} & {$\Delta\mathcal{D}$}
   & {FID} & {CLIP} & {ID} & {$\Delta\mathcal{D}$}
   & {FID} & {CLIP} & {ID} & {$\Delta\mathcal{D}$}
   & {FID} & {CLIP} & {ID} & {$\Delta\mathcal{D}$} \\
\midrule
\multirow{2}{*}{\rotatebox[origin=c]{90}{\scriptsize 2D}}
& InstructPix2Pix%
  & \cellcolor{color4}144.4  %
  & \cellcolor{color2}0.82   %
  & \cellcolor{color7}0.40   %
  & \cellcolor{color5}0.024  %
  & \cellcolor{color5}116.8  %
  & \cellcolor{color1}0.90   %
  & \cellcolor{color5}0.44   %
  & \cellcolor{color5}0.023  %
  & \cellcolor{color3}178.1  %
  & \cellcolor{color2}0.82   %
  & \cellcolor{color7}0.28   %
  & \cellcolor{color5}0.013  %
  & \cellcolor{color1}89.5   %
  & \cellcolor{color6}0.65   %
  & \cellcolor{color7}0.24   %
  & \cellcolor{color1}0.002  %
  & \cellcolor{color1}60.9   %
  & \cellcolor{color1}0.91   %
  & \cellcolor{color6}0.39   %
  & \cellcolor{color4}0.011  %
\\

& InstantID%
  & \cellcolor{color5}160.8  %
  & \cellcolor{color7}0.68   %
  & \cellcolor{color4}0.59   %
  & \cellcolor{color7}0.050  %
  & \cellcolor{color7}170.3  %
  & \cellcolor{color7}0.69   %
  & \cellcolor{color3}0.47   %
  & \cellcolor{color7}0.066  %
  & \cellcolor{color4}183.3  %
  & \cellcolor{color7}0.58   %
  & \cellcolor{color3}0.53   %
  & \cellcolor{color7}0.065  %
  & \cellcolor{color7}162.7  %
  & \cellcolor{color7}0.65   %
  & \cellcolor{color2}0.59   %
  & \cellcolor{color7}0.047  %
  & \cellcolor{color5}160.4  %
  & \cellcolor{color6}0.71   %
  & \cellcolor{color2}0.56   %
  & \cellcolor{color6}0.038  %
\\
\midrule

\multirow{5}{*}{\rotatebox[origin=c]{90}{\scriptsize 3D domain apt.}}
& StyleCLIP%
  & \cellcolor{color3}118.2  %
  & \cellcolor{color5}0.77   %
  & \cellcolor{color5}0.52   %
  & \cellcolor{color4}0.022  %
  & \cellcolor{color2}97.6   %
  & \cellcolor{color6}0.75   %
  & \cellcolor{color7}0.38   %
  & \cellcolor{color6}0.031  %
  & \cellcolor{color7}248.8  %
  & \cellcolor{color6}0.63   %
  & \cellcolor{color5}0.42   %
  & \cellcolor{color6}0.039  %
  & \cellcolor{color5}103.7  %
  & \cellcolor{color2}0.75   %
  & \cellcolor{color3}0.54   %
  & \cellcolor{color5}0.026  %
  & \cellcolor{color7}181.6  %
  & \cellcolor{color7}0.63   %
  & \cellcolor{color1}0.61   %
  & \cellcolor{color5}0.012  %
\\

& StyleGAN-NADA%
  & \cellcolor{color2}81.1   %
  & \cellcolor{color4}0.81   %
  & \cellcolor{color2}0.61   %
  & \cellcolor{color2}0.013  %
  & \cellcolor{color4}116.3  %
  & \cellcolor{color4}0.81   %
  & \cellcolor{color2}0.50   %
  & \cellcolor{color4}0.011  %
  & \cellcolor{color6}212.9  %
  & \cellcolor{color4}0.75   %
  & \cellcolor{color4}0.45   %
  & \cellcolor{color3}0.012  %
  & \cellcolor{color4}99.7   %
  & \cellcolor{color4}0.71   %
  & \cellcolor{color5}0.51   %
  & \cellcolor{color6}0.034  %
  & \cellcolor{color4}154.5  %
  & \cellcolor{color5}0.76   %
  & \cellcolor{color7}0.36   %
  & \cellcolor{color7}0.039  %
\\

& StyleGANFusion%
  & \cellcolor{color6}168.0  %
  & \cellcolor{color6}0.76   %
  & \cellcolor{color3}0.60   %
  & \cellcolor{color1}0.003  %
  & \cellcolor{color6}119.3  %
  & \cellcolor{color5}0.80   %
  & \cellcolor{color6}0.41   %
  & \cellcolor{color2}0.007  %
  & \cellcolor{color5}203.9  %
  & \cellcolor{color5}0.70   %
  & \cellcolor{color6}0.41   %
  & \cellcolor{color4}0.012  %
  & \cellcolor{color3}99.0   %
  & \cellcolor{color3}0.74   %
  & \cellcolor{color4}0.53   %
  & \cellcolor{color3}0.012  %
  & \cellcolor{color2}85.5   %
  & \cellcolor{color2}0.83   %
  & \cellcolor{color4}0.47   %
  & \cellcolor{color2}0.007  %
\\

& DiffusionGAN3D%
  & \cellcolor{color7}189.3  %
  & \cellcolor{color3}0.82   %
  & \cellcolor{color6}0.46   %
  & \cellcolor{color6}0.024  %
  & \cellcolor{color3}110.7  %
  & \cellcolor{color3}0.86   %
  & \cellcolor{color4}0.47   %
  & \cellcolor{color3}0.009  %
  & \cellcolor{color2}132.3  %
  & \cellcolor{color3}0.81   %
  & \cellcolor{color1}0.68   %
  & \cellcolor{color1}0.002  %
  & \cellcolor{color6}159.4  %
  & \cellcolor{color5}0.71   %
  & \cellcolor{color6}0.48   %
  & \cellcolor{color4}0.016  %
  & \cellcolor{color6}166.2  %
  & \cellcolor{color3}0.82   %
  & \cellcolor{color5}0.43   %
  & \cellcolor{color3}0.009  %
\\

& \textbf{Ours}
  & \cellcolor{color1}77.6   %
  & \cellcolor{color1}0.86   %
  & \cellcolor{color1}0.69   %
  & \cellcolor{color3}0.014  %
  & \cellcolor{color1}67.7   %
  & \cellcolor{color2}0.89   %
  & \cellcolor{color1}0.56   %
  & \cellcolor{color1}0.003  %
  & \cellcolor{color1}99.7   %
  & \cellcolor{color1}0.85   %
  & \cellcolor{color2}0.56   %
  & \cellcolor{color2}0.002  %
  & \cellcolor{color2}91.6   %
  & \cellcolor{color1}0.77   %
  & \cellcolor{color1}0.75   %
  & \cellcolor{color2}0.005  %
  & \cellcolor{color3}144.5  %
  & \cellcolor{color3}0.82   %
  & \cellcolor{color3}0.55   %
  & \cellcolor{color1}0.006  %
\\
\bottomrule
\end{tabular}
\caption{Quantitative scores with competitive domain adaptation and 2D-3D editing methods.}
\label{table:results}
\vspace{-0.25cm}
\end{table*}

\begin{figure}[t]
\centering
\begin{minipage}{0.8\linewidth}
\includegraphics[width=\linewidth]{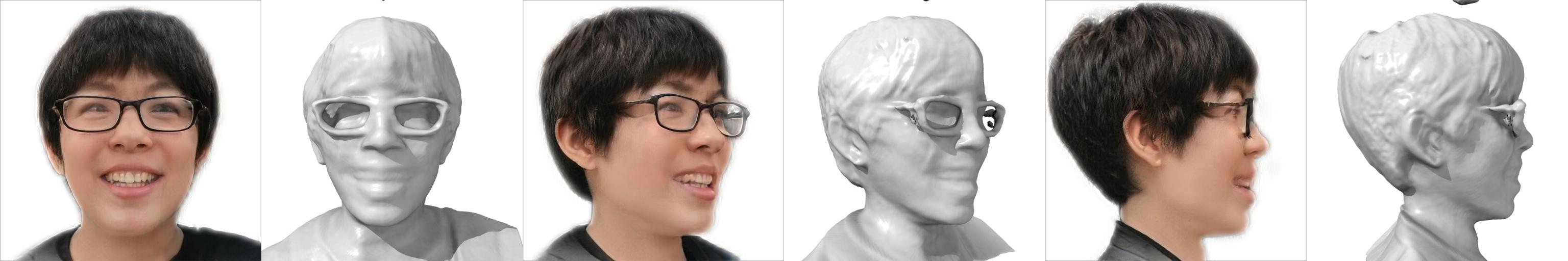}
\end{minipage}%
\begin{minipage}{0.05\linewidth}
\centering
\rotatebox[origin=c]{270}{\scriptsize Input}
\end{minipage}

\begin{minipage}{0.8\linewidth}
\includegraphics[width=\linewidth]{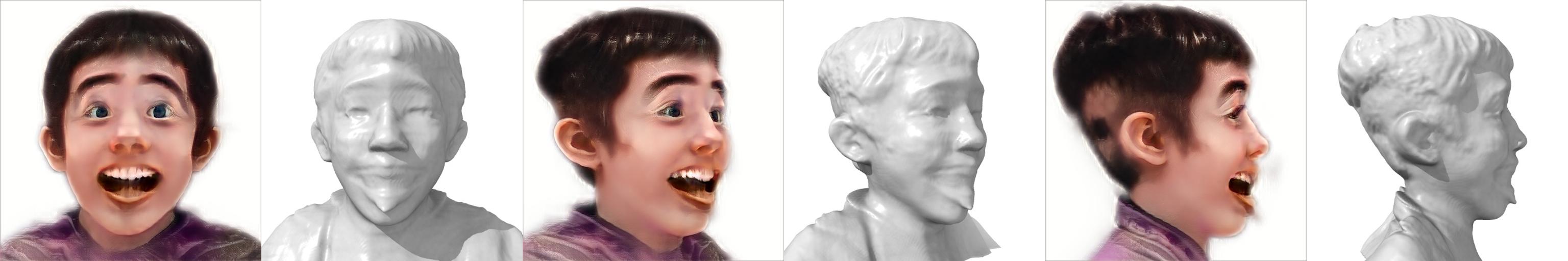}
\end{minipage}%
\begin{minipage}{0.05\linewidth}
\centering
\rotatebox[origin=c]{270}{\scriptsize StyleCLIP} 
\end{minipage}

\begin{minipage}{0.8\linewidth}
\includegraphics[width=\linewidth]{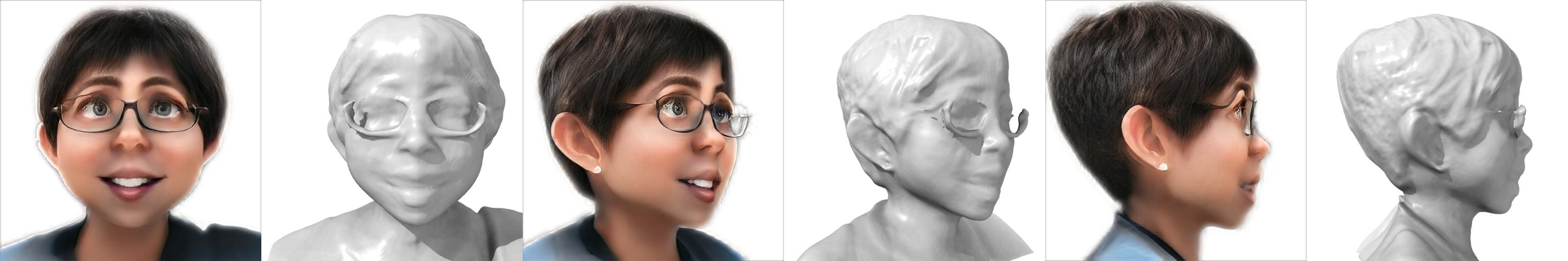}
\end{minipage}%
\begin{minipage}{0.05\linewidth}
\centering
\rotatebox[origin=c]{270}{\scriptsize \shortstack{StyleGAN\\NADA}} 
\end{minipage}

\begin{minipage}{0.8\linewidth}
\includegraphics[width=\linewidth]{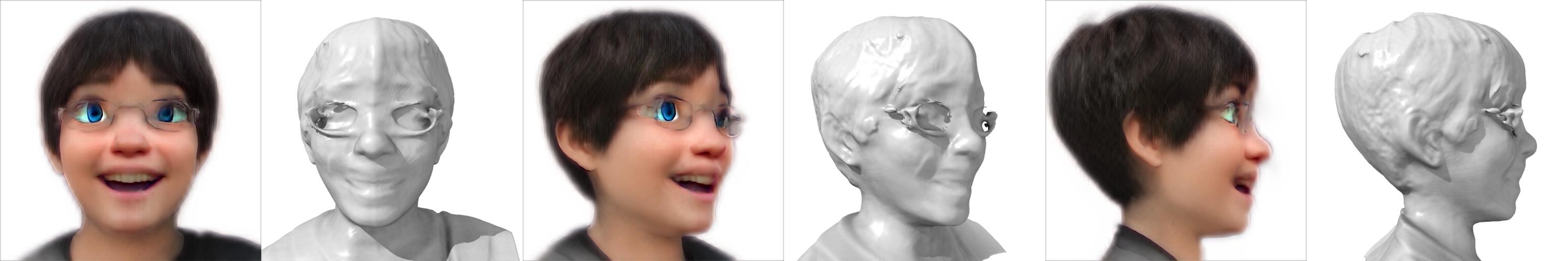}
\end{minipage}%
\begin{minipage}{0.05\linewidth}
\centering
\rotatebox[origin=c]{270}{\scriptsize \shortstack{StyleGAN\\Fusion}} 
\end{minipage}

\begin{minipage}{0.8\linewidth}
\includegraphics[width=\linewidth]{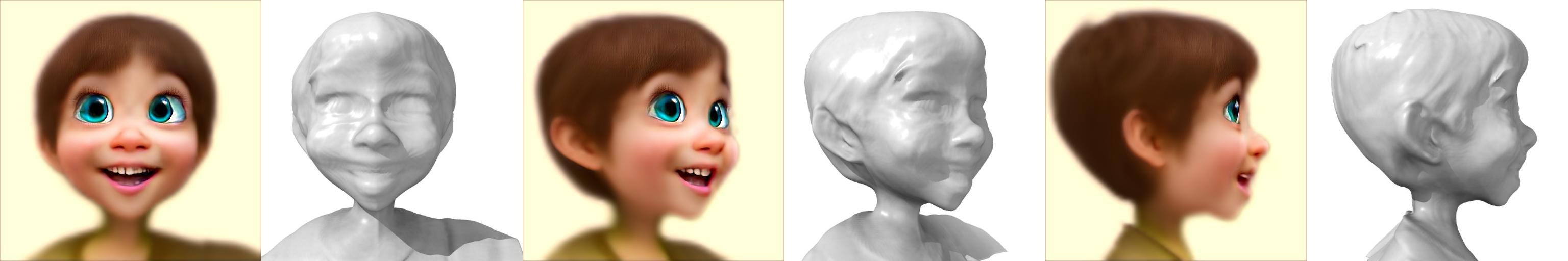}
\end{minipage}%
\begin{minipage}{0.05\linewidth}
\centering
\rotatebox[origin=c]{270}{\scriptsize \shortstack{Diffusion\\GAN3D}} 
\end{minipage}

\begin{minipage}{0.8\linewidth}
\includegraphics[width=\linewidth]{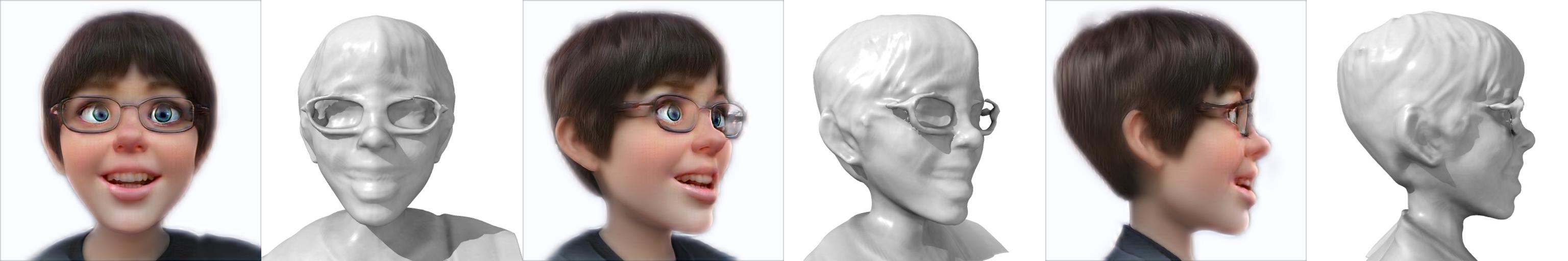}
\end{minipage}%
\begin{minipage}{0.05\linewidth}
\centering
\rotatebox[origin=c]{270}{\scriptsize Ours} 
\end{minipage}

\caption{Mesh and rendering visualizations of ours and competing methods for the Pixar prompt.%
}  
\vspace{-0.5cm}
\label{fig:meshes}
\vspace{-0.25cm}
\end{figure}

In~\cref{fig:meshes}, we present the mesh predictions for the Pixar-style stylizations produced by our method and the competing approaches. The first row shows reconstructions of the input image from different views, in RGB and mesh. The results demonstrate that our method achieves significantly better identity preservation both in terms of facial features and distinctive accessories, such as eyeglasses, unlike other methods which completely remove or distort them. %

\begin{figure}[ht!]
\setlength\tabcolsep{1pt}
\scriptsize
\centering
\begin{tabular}{cccccc}
Input & Pixar & Joker & Werewolf & Sketch
\\
\includegraphics[width=0.09\textwidth]{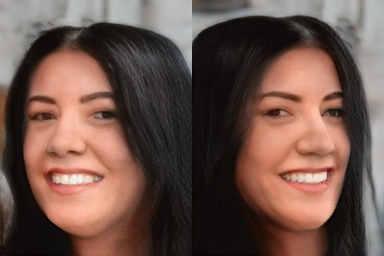}&
\includegraphics[width=0.09\textwidth]{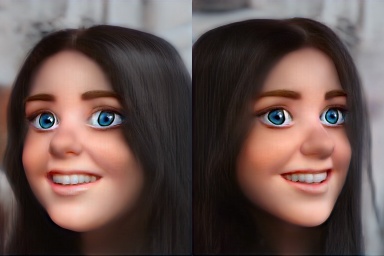}&
\includegraphics[width=0.09\textwidth]{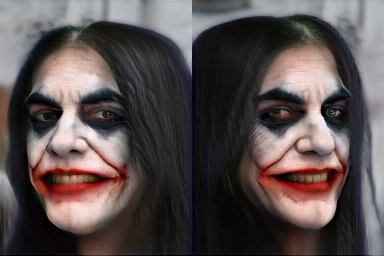}&
\includegraphics[width=0.09\textwidth]{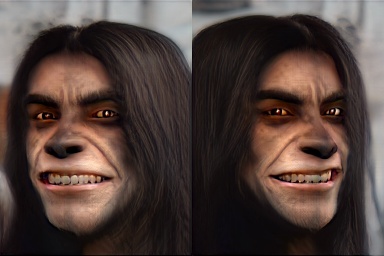}&
\includegraphics[width=0.09\textwidth]{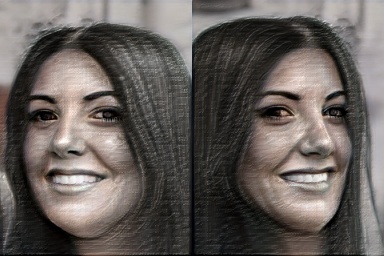}&
\rotatebox{90}{~~~~\textbf{Ours}}
\\
&
\includegraphics[width=0.09\textwidth]{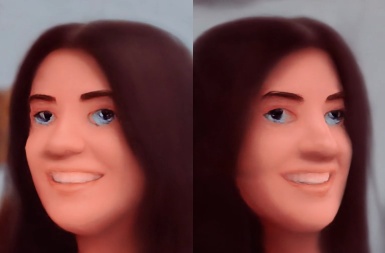}&
\includegraphics[width=0.09\textwidth]{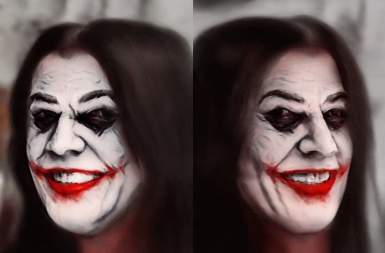}&
\includegraphics[width=0.09\textwidth]{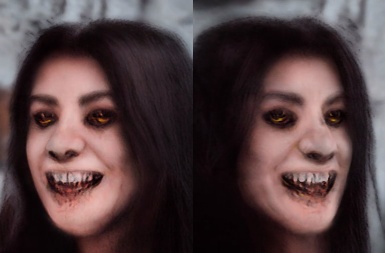}&
\includegraphics[width=0.09\textwidth]{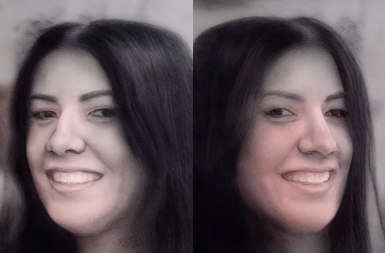}&
\rotatebox{90}{~~~~IN2N}
\\
&
\includegraphics[width=0.09\textwidth]{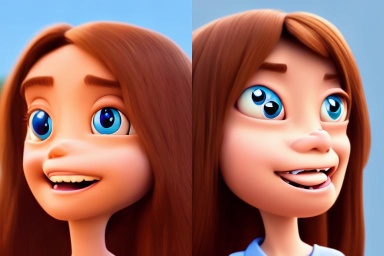}&
\includegraphics[width=0.09\textwidth]{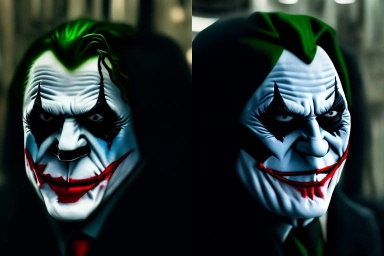}&
\includegraphics[width=0.09\textwidth]{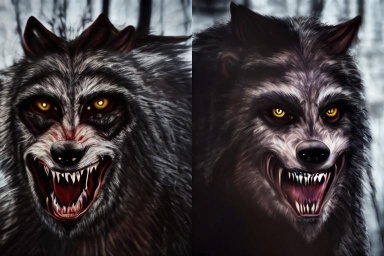}&
\includegraphics[width=0.09\textwidth]{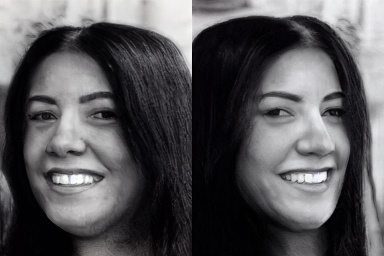}&
\rotatebox{90}{~~~~IP2P}
\\
&
\includegraphics[width=0.09\textwidth]{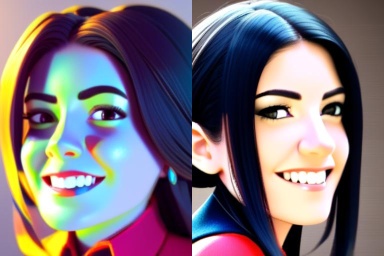}&
\includegraphics[width=0.09\textwidth]{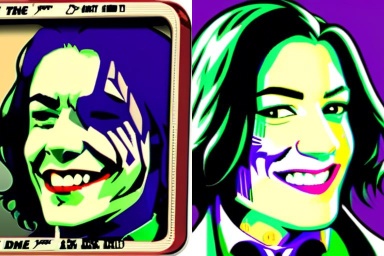}&
\includegraphics[width=0.09\textwidth]{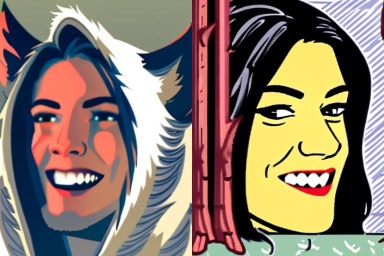}&
\includegraphics[width=0.09\textwidth]{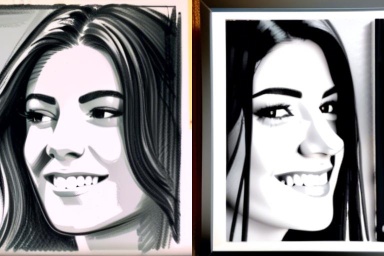}&
\rotatebox{90}{InstantID}
\\
\includegraphics[width=0.09\textwidth]{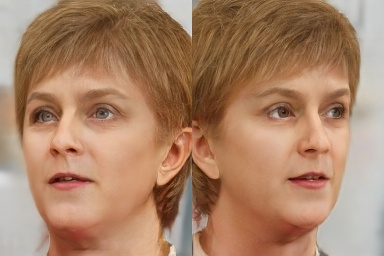}&
\includegraphics[width=0.09\textwidth]{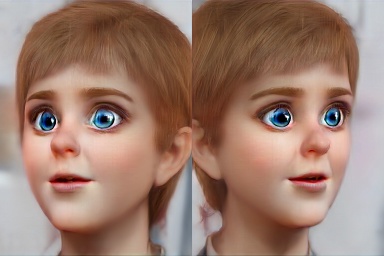}&
\includegraphics[width=0.09\textwidth]{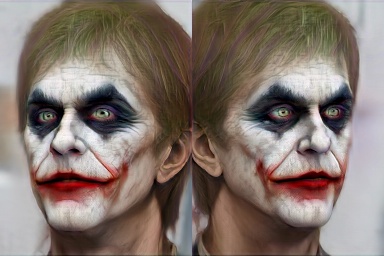}&
\includegraphics[width=0.09\textwidth]{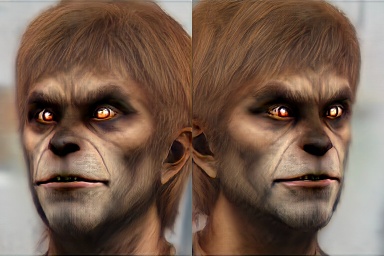}&
\includegraphics[width=0.09\textwidth]{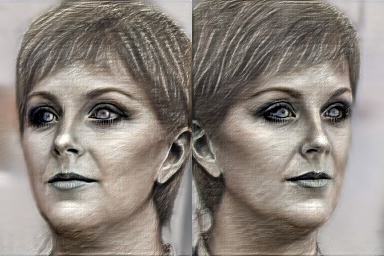}&
\rotatebox{90}{~~~~\textbf{Ours}}
\\
&
\includegraphics[width=0.09\textwidth]{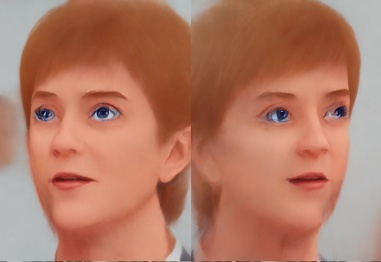}&
\includegraphics[width=0.09\textwidth]{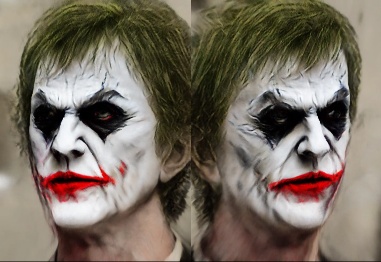}&
\includegraphics[width=0.09\textwidth]{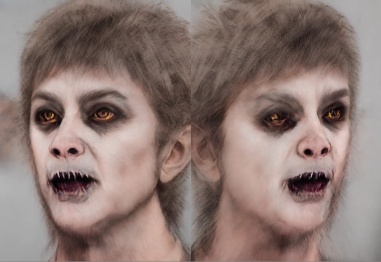}&
\includegraphics[width=0.09\textwidth]{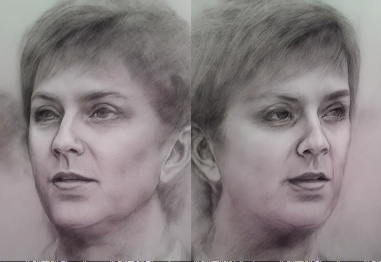}&
\rotatebox{90}{~~~~IN2N}
\\
&
\includegraphics[width=0.09\textwidth]{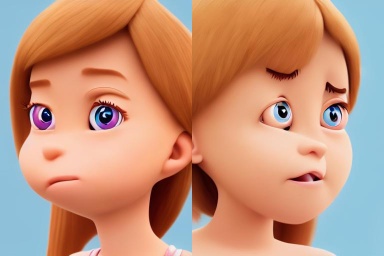}&
\includegraphics[width=0.09\textwidth]{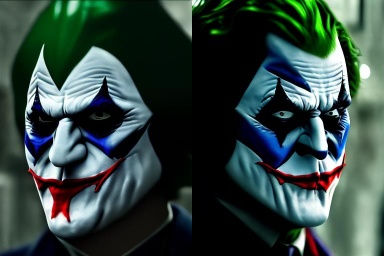}&
\includegraphics[width=0.09\textwidth]{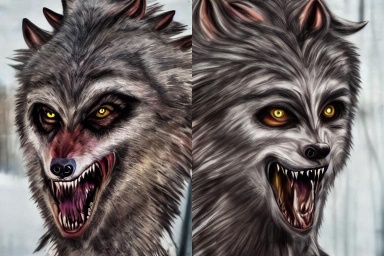}&
\includegraphics[width=0.09\textwidth]{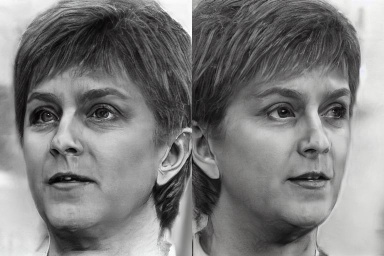}&
\rotatebox{90}{~~~~IP2P}
\\
&
\includegraphics[width=0.09\textwidth]{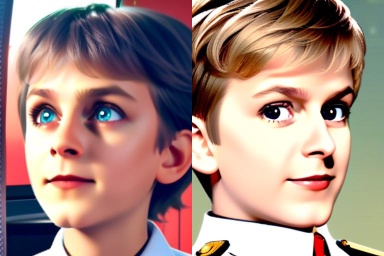}&
\includegraphics[width=0.09\textwidth]{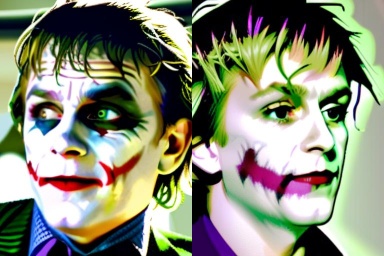}&
\includegraphics[width=0.09\textwidth]{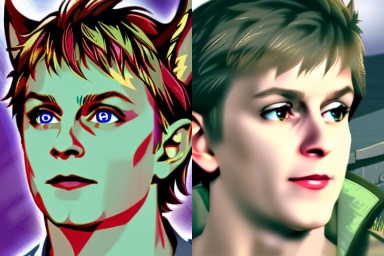}&
\includegraphics[width=0.09\textwidth]{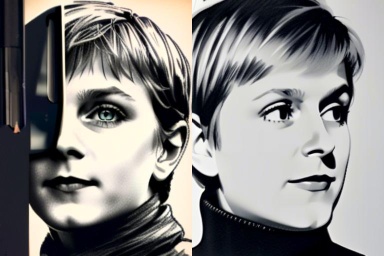}&
\rotatebox{90}{InstantID}
\\
\end{tabular}
\vspace{-0.3cm}
\caption{Additional comparisons with other 2D-3D editing methods. 2D methods are run with the same seed for different views.}
\label{fig:rebuttal_ip2p_in2n}
\end{figure}

\noindent \textbf{Ablation Study.} We begin our ablation study with the DiffusionGAN3D~\cite{lei2023diffusiongan3d} baseline and build upon it by incorporating various improvements.~\cite{lei2023diffusiongan3d} uses SDS for generator optimization, along with a relative distance loss. This relative distance loss aims to preserve the distance between two images generated by the original generator. The fine-tuned generator is then optimized to maintain this same distance between the corresponding generated samples.

\newcommand{\interpfigtdk}[1]{\includegraphics[trim=0 0 0cm 0, clip, width=0.9\linewidth]{#1}}

\begin{figure}[t!]
\centering
\scriptsize
\setlength{\tabcolsep}{0.7pt}
\scalebox{1.0}{
\begin{tabular}{cccccc}
\raisebox{0.2in}{\rotatebox[origin=t]{90}{Werewolf}}&
\includegraphics[width=0.15\linewidth]{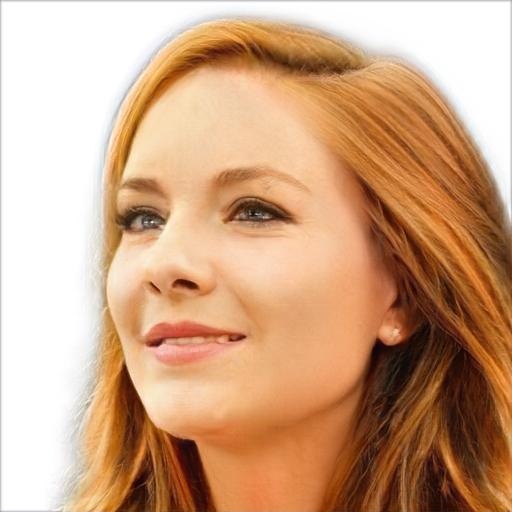}&
\includegraphics[width=0.15\linewidth]{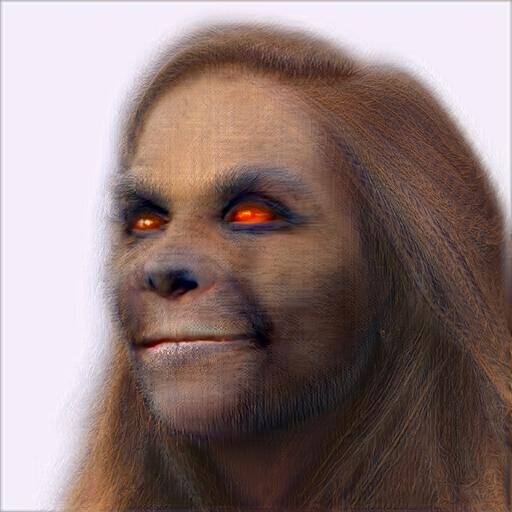}&
\includegraphics[width=0.15\linewidth]{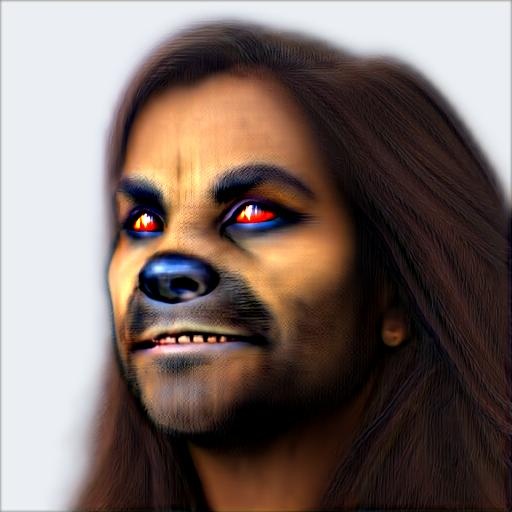}&
\includegraphics[width=0.15\linewidth]{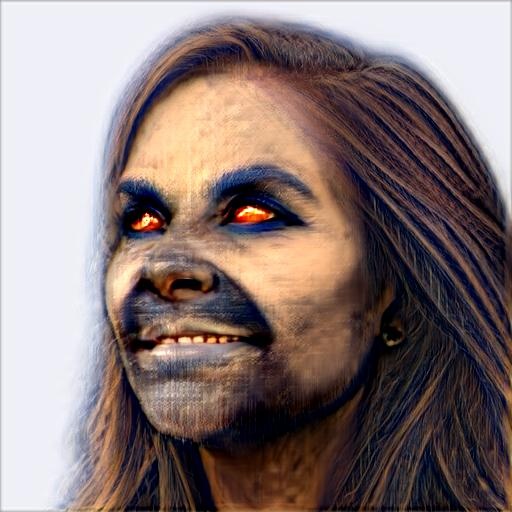}&
\includegraphics[width=0.15\linewidth]{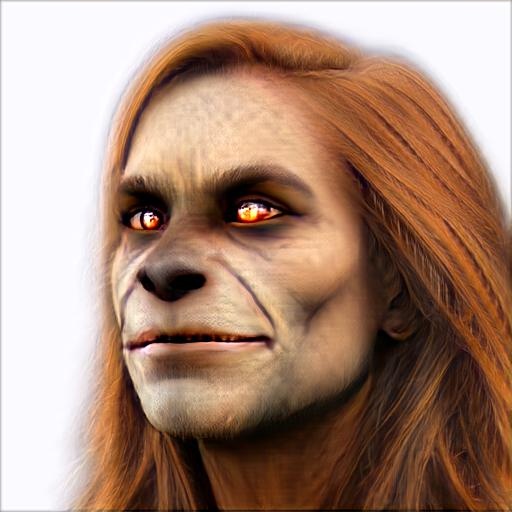}
\\

\raisebox{0.2in}{\rotatebox[origin=t]{90}{Pixar}}&
\includegraphics[width=0.15\linewidth]{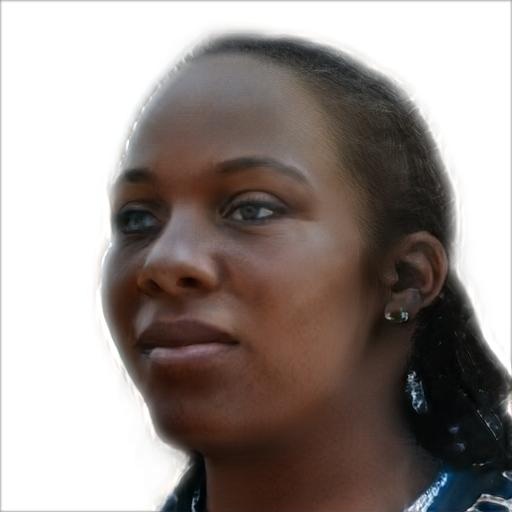}&
\includegraphics[width=0.15\linewidth]{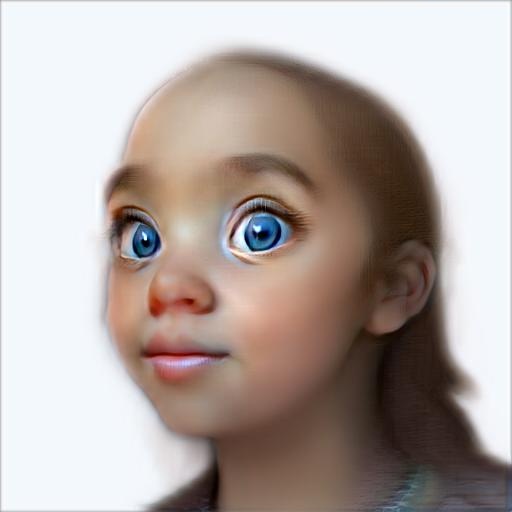}&
\includegraphics[width=0.15\linewidth]{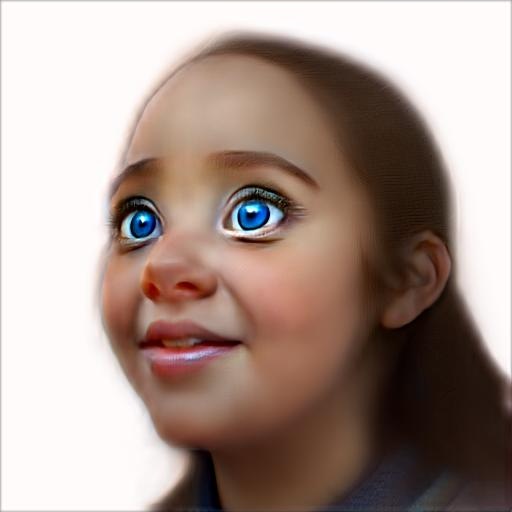}&
\includegraphics[width=0.15\linewidth]{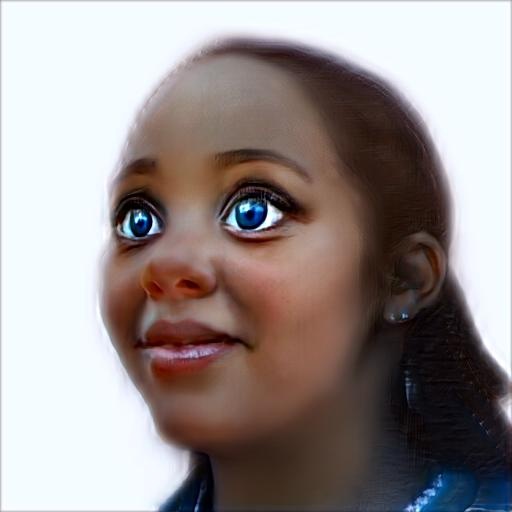}&
\includegraphics[width=0.15\linewidth]{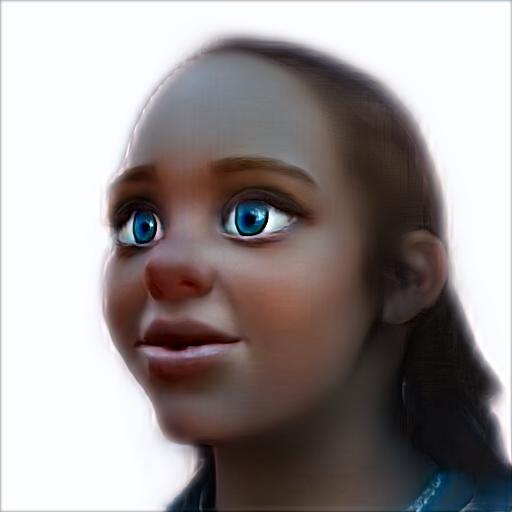}
\\

\raisebox{0.2in}{\rotatebox[origin=t]{90}{Joker}}&
\includegraphics[width=0.15\linewidth]{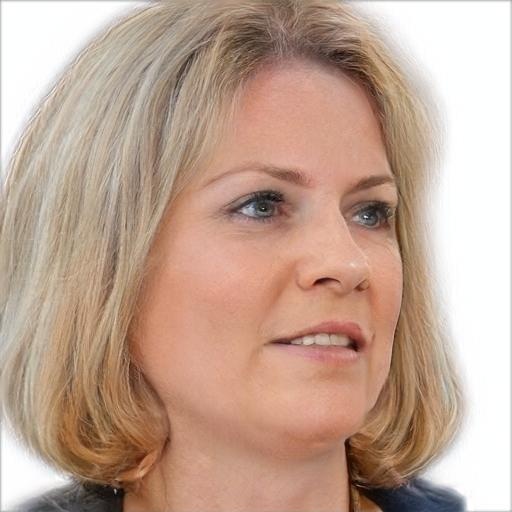}&
\includegraphics[width=0.15\linewidth]{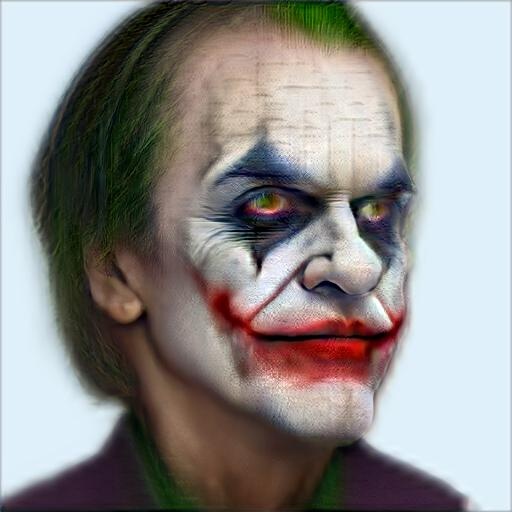}&
\includegraphics[width=0.15\linewidth]{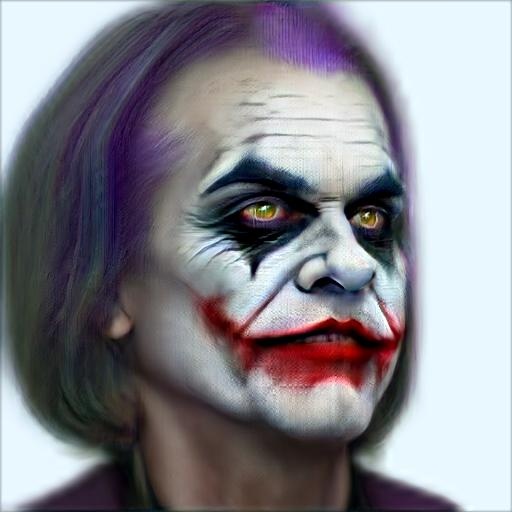}&
\includegraphics[width=0.15\linewidth]{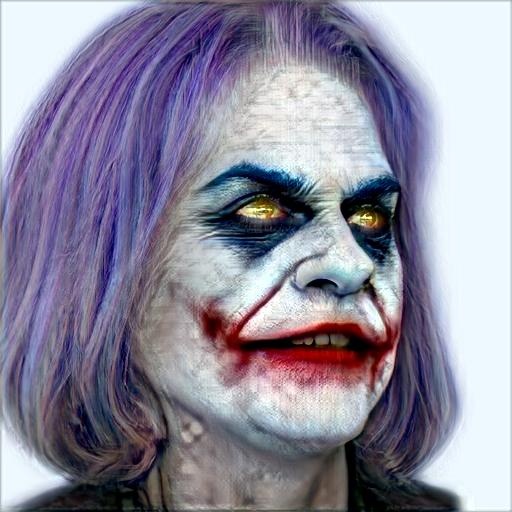}&
\includegraphics[width=0.15\linewidth]{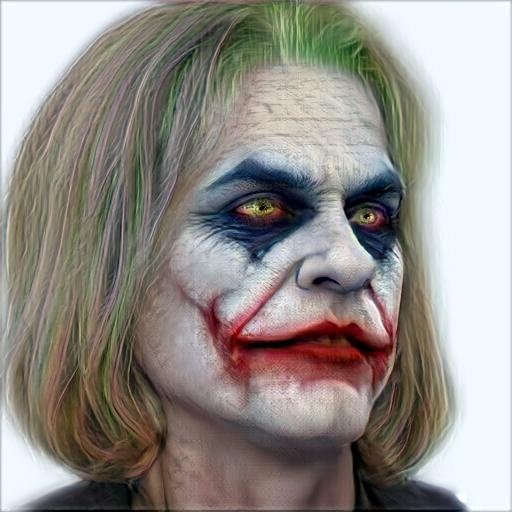}
\\

\raisebox{0.2in}{\rotatebox[origin=t]{90}{Sketch}}&
\includegraphics[width=0.15\linewidth]{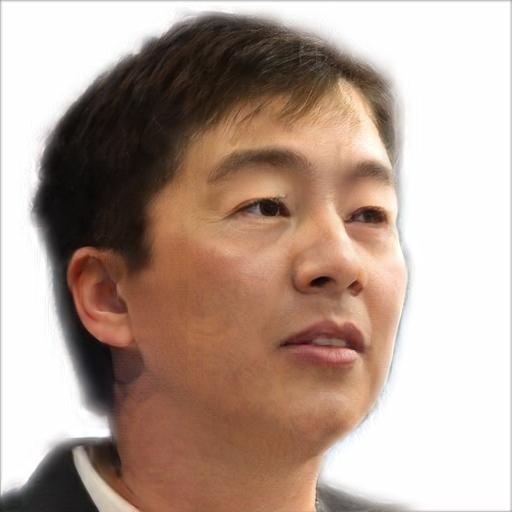}&
\includegraphics[width=0.15\linewidth]{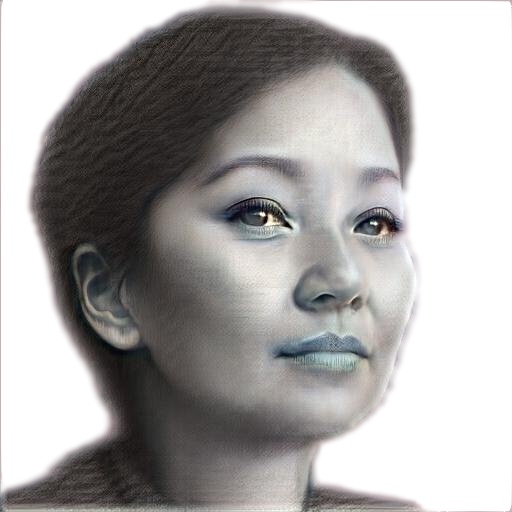}&
\includegraphics[width=0.15\linewidth]{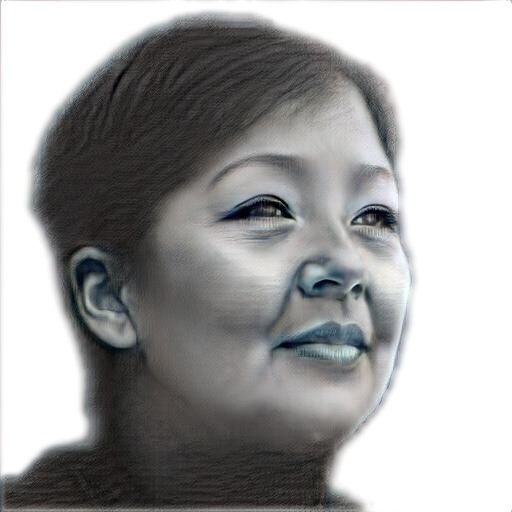}&
\includegraphics[width=0.15\linewidth]{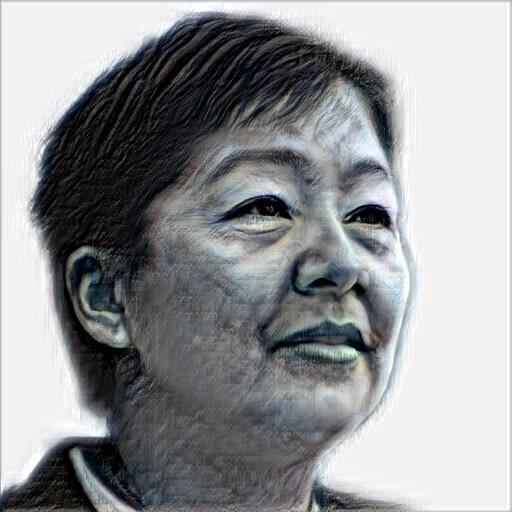}&
\includegraphics[width=0.15\linewidth]{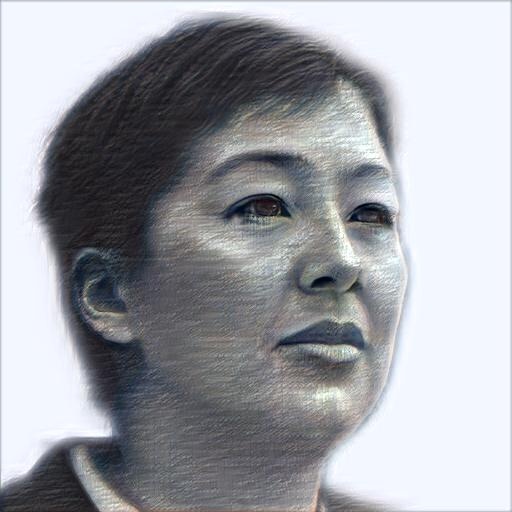}
\\

& Input & \cite{lei2023diffusiongan3d} & \cite{lei2023diffusiongan3d} + \cite{Zhang_2023_ICCV}& LD + \cite{Zhang_2023_ICCV} & \textbf{Ours}

\end{tabular}
}
\caption{Qualitative Ablation Study I. LD and other regularizers like grid and mirror scores help keep the original attributes.}
\label{fig:ablation_ld}
\vspace{-0.3cm}
\end{figure}

As shown in~\cref{fig:ablation_ld}, while their method successfully achieves stylization, the results fail to preserve the characteristics of the input images. Next, we attach depth-based ControlNet~\cite{Zhang_2023_ICCV} to the text-based diffusion model, which helps retain more characteristics from the input image, though its effectiveness remains limited.
In the next setting, we apply negative log-likelihood distillation (LD) alongside the depth-based ControlNet. This results in less saturation, and we observe improved identity preservation. 
We also show our final results demonstrating the improved color and geometry alignment with the input images that include the improvements from grid and mirror gradients in the last column.
These visuals are consistent with the quantitative improvements presented in~\cref{tab:ablation}.

We continue our ablation study in~\cref{fig:ablation_mirror_grid}.
We start from the previous study with the results of LD, including rank-weighted score tensors and depth-based ControlNet guidance.
The ablation study for rank-weighting is presented in~\cref{fig:rank} to provide a clearer explanation of the method, so we omit that discussion here.
Next, we add the grid-based denoising, which provides multi-view consistent distillation. 
We observe that the multi-view grid gradients improve the results even when we feed them after the super-resolution layers, but if we feed the gradients to the generator before the super-resolution network, the results improve further both quantitatively and qualitatively, as shown in~\cref{tab:ablation} and~\cref{fig:ablation_mirror_grid}, respectively. Finally, we incorporate cross-dependencies with mirror poses, enhancing intricate details, as evidenced by the zoomed-in views.

\noindent\textbf{Runtime.} Our method requires $\sim$3h of generator tuning per prompt on a single A100 GPU. After tuning, inference on any head image takes only $\sim$0.5s. In comparison,~\cite{instructnerf2023} requires $\sim$1h to fit NeRF to a single identity and +3h for stylizing the scene.~\cite{wang2024instantid,brooks2022instructpix2pix} take 50 iterations through denoiser, taking $\sim$10-15s. \cite{lei2023diffusiongan3d,song2022diffusion} also tune their generators, and their training \& inference times are similar to ours.

\begin{table}[t!]
\footnotesize
\centering
\setlength{\tabcolsep}{3pt}
\begin{tabular}{rcccccc}
\toprule
& FID & {$\Delta$} & CLIP & {$\Delta$} & ID & {$\Delta$}\\
\midrule
\cite{lei2023diffusiongan3d} & 147.19 & - & 0.74 & - & 0.35 & -\\
\cite{lei2023diffusiongan3d} + \cite{Zhang_2023_ICCV} & 140.20 & \cellcolor{color2}{6.99} & 0.75 & \cellcolor{color2}{0.01} & 0.34 & \cellcolor{Melon}{0.01}\\
LD + \cite{Zhang_2023_ICCV} & 116.88 & \cellcolor{color2}{23.32} & 0.81 & \cellcolor{color2}{0.06} & 0.46 & \cellcolor{color2}{0.12}\\
LD + \cite{Zhang_2023_ICCV} + Weighted rank & 116.09 & \cellcolor{Melon}{0.79} & 0.82 & \cellcolor{color2}{0.01} & 0.48 & \cellcolor{color2}{0.02} \\
+ Grid after SR & 99.53 & \cellcolor{color2}{16.56} & 0.85 & \cellcolor{color2}{0.03} & 0.49 & \cellcolor{color2}{0.01}\\
+ Grid before SR & 81.11 & \cellcolor{color2}{18.42} & 0.85 & \cellcolor{Gray}{0.00} & 0.51 & \cellcolor{color2}{0.02}\\
+ Mirror gradients & 84.17 & \cellcolor{Melon}{3.06} & 0.86 & \cellcolor{color2}{0.01} & 0.52 & \cellcolor{color2}{0.01}\\
\bottomrule
\end{tabular}
\caption{Quantitative ablation study. {$\Delta$} represents the difference between the results of the current row and the row above. Our approach improves stylization with ID retention.}
\label{tab:ablation}
\end{table}

\begin{figure}[t!]
\centering
\footnotesize
\setlength{\tabcolsep}{1pt}
\scalebox{1.0}{
\begin{tabular}{ccccccc}
& Input & \shortstack{LD + \cite{Zhang_2023_ICCV} \\ $\mathbf{\Sigma}$-weigh} & \shortstack{Grid \\ after SR} & \shortstack{Grid \\ before SR} & \shortstack{Grid + \\mirror}\\

\raisebox{0.2in}{\rotatebox[origin=t]{90}{Werewolf}}&
\includegraphics[width=0.15\linewidth]{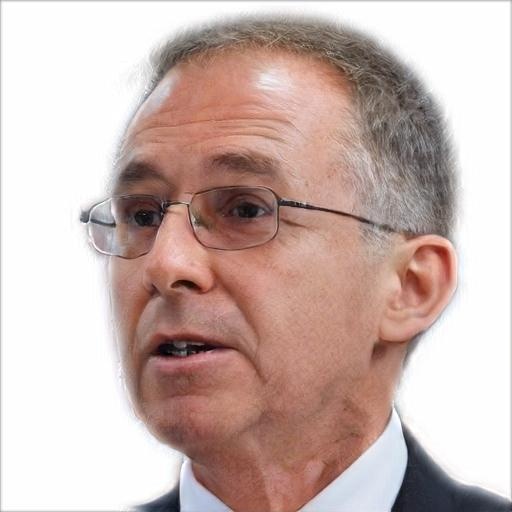}&
\includegraphics[width=0.15\linewidth]{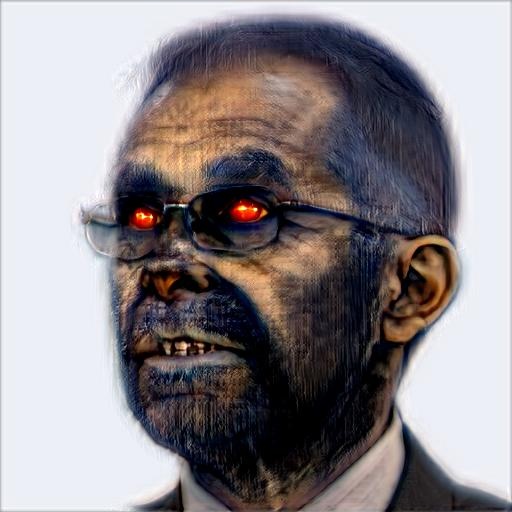}&
\includegraphics[width=0.15\linewidth]{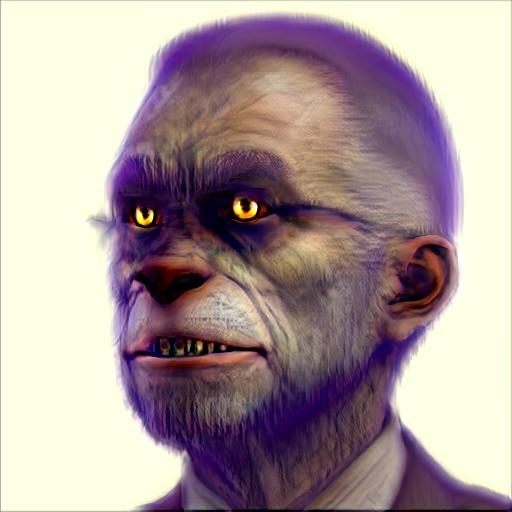}&
\includegraphics[width=0.15\linewidth]{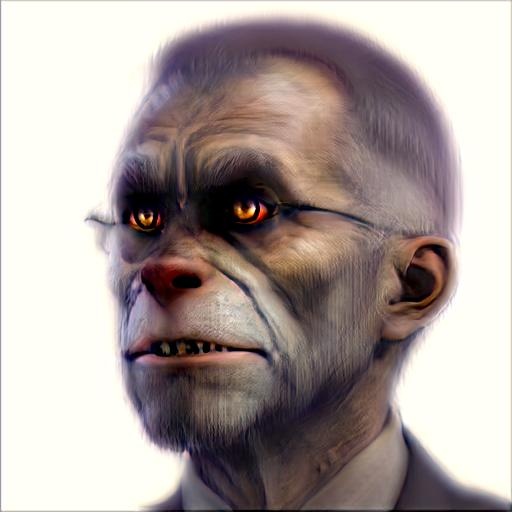}&
\includegraphics[width=0.15\linewidth]{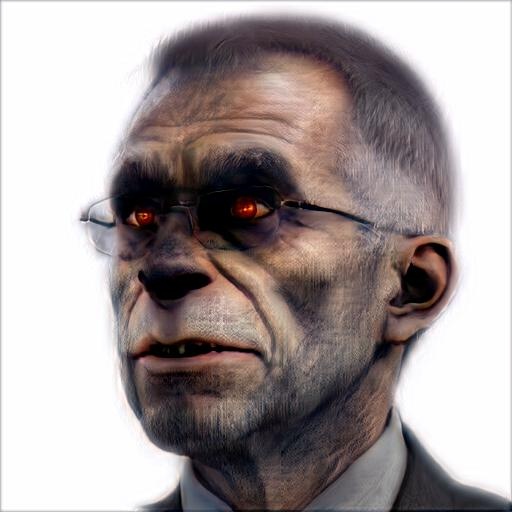}
\\
&
\includegraphics[width=0.15\linewidth,trim=65 150 205 150, clip]{figures/ablation_grid_mirror/275_11/abl_25_1.jpg}&
\includegraphics[width=0.15\linewidth,trim=65 150 205 150, clip]{figures/ablation_grid_mirror/275_11/abl_25_2.jpg}&
\includegraphics[width=0.15\linewidth,trim=65 150 205 150, clip]{figures/ablation_grid_mirror/275_11/abl_25_3.jpg}&
\includegraphics[width=0.15\linewidth,trim=65 150 205 150, clip]{figures/ablation_grid_mirror/275_11/abl_25_4.jpg}&
\includegraphics[width=0.15\linewidth,trim=65 150 205 150, clip]{figures/ablation_grid_mirror/275_11/abl_25_7.jpg}
\\
\raisebox{0.2in}{\rotatebox[origin=t]{90}{Pixar}}&
\includegraphics[width=0.15\linewidth]{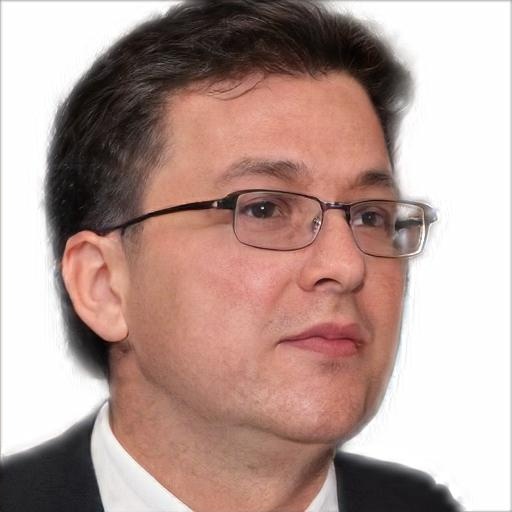}&
\includegraphics[width=0.15\linewidth]{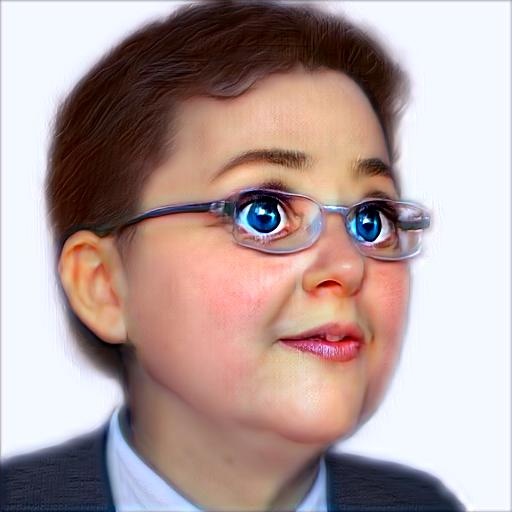}&
\includegraphics[width=0.15\linewidth]{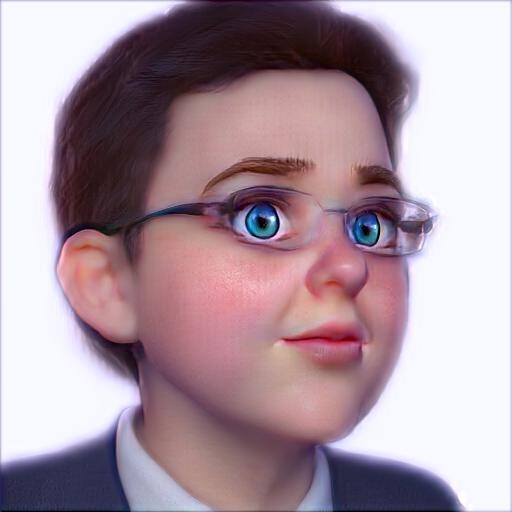}&
\includegraphics[width=0.15\linewidth]{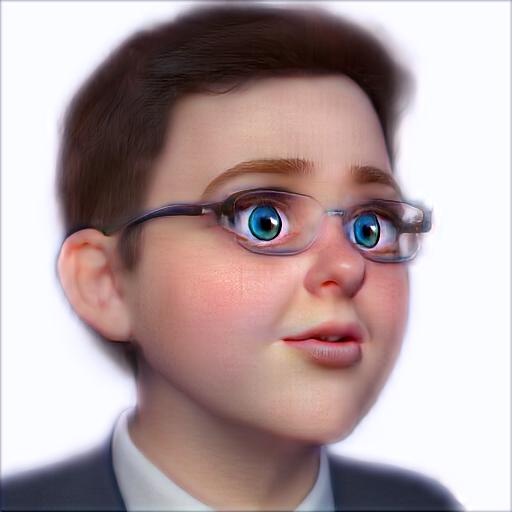}&
\includegraphics[width=0.15\linewidth]{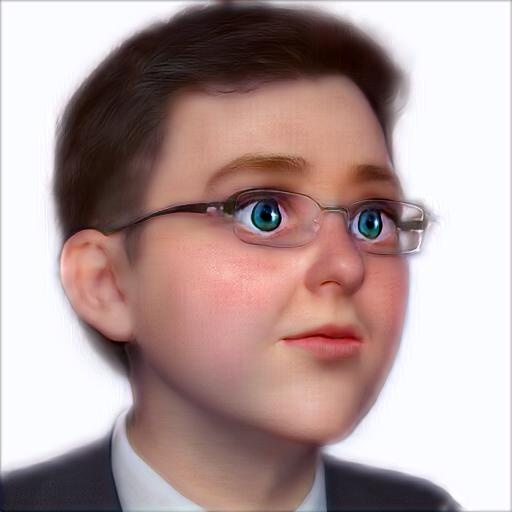}
\\
&
\includegraphics[width=0.15\linewidth,trim=190 150 75 150, clip]{figures/ablation_grid_mirror/166_6/abl_25_1.jpg}&
\includegraphics[width=0.15\linewidth,trim=190 150 75 150, clip]{figures/ablation_grid_mirror/166_6/abl_25_2.jpg}&
\includegraphics[width=0.15\linewidth,trim=190 150 75 150, clip]{figures/ablation_grid_mirror/166_6/abl_25_3.jpg}&
\includegraphics[width=0.15\linewidth,trim=190 150 75 150, clip]{figures/ablation_grid_mirror/166_6/abl_25_4.jpg}&
\includegraphics[width=0.15\linewidth,trim=190 150 75 150, clip]{figures/ablation_grid_mirror/166_6/abl_25_9.jpg}
\end{tabular}
}
\caption{Qualitative Ablation Study II. Grid distillation improves stylization while preserving identity, and mirror gradients focus on symmetric intricate features like glasses.}
\label{fig:ablation_mirror_grid}
\vspace{-0.5cm}
\end{figure}

\begin{figure}
\centering
\scriptsize
\setlength{\tabcolsep}{0.7pt}
\scalebox{0.96}{
\begin{tabular}{cl}
& \;\;\;\quad Input \\
\raisebox{0.2in}{\rotatebox[origin=t]{90}{Elf}}&\includegraphics[width=0.8\linewidth]{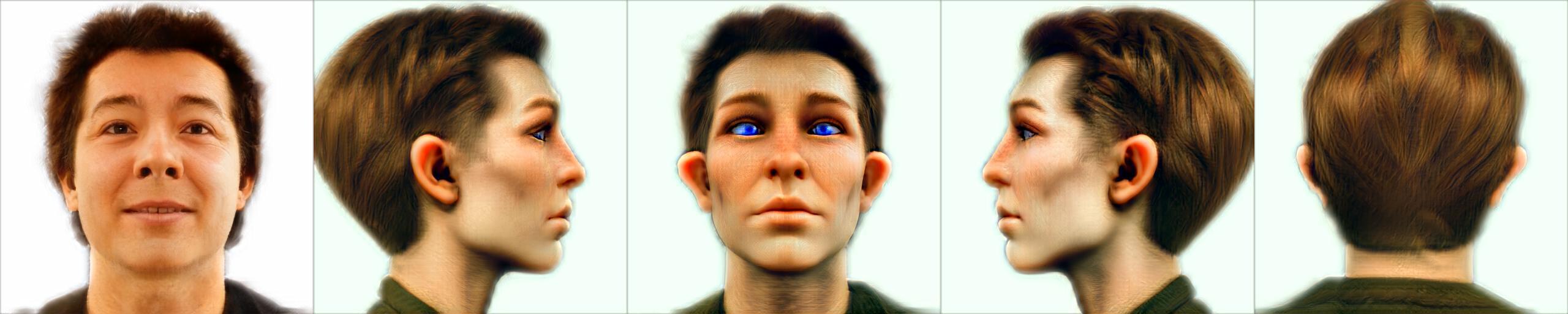}
\\
\raisebox{0.22in}{\rotatebox[origin=t]{90}{Moustache}}&\includegraphics[width=0.8\linewidth]{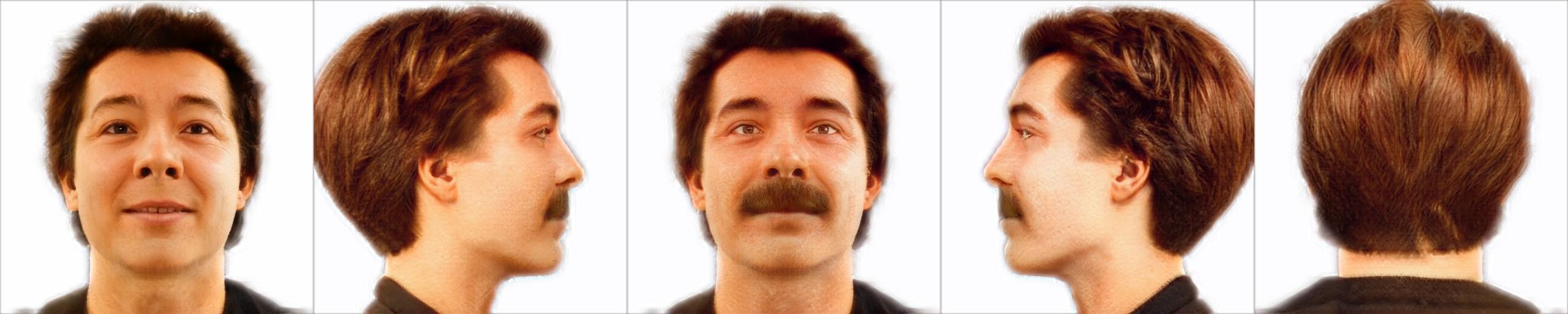}
\\
\raisebox{0.1in}{\rotatebox[origin=t]{90}{\shortstack{Rainbow  hair\\and makeup}}}&\includegraphics[width=0.8\linewidth]{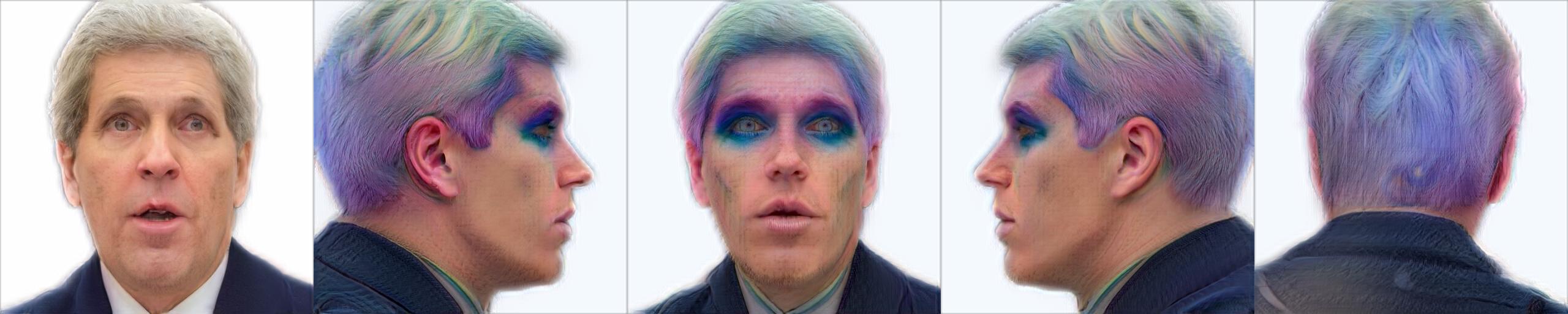}
\\
\raisebox{0.2in}{\rotatebox[origin=t]{90}{Pink hair}}&\includegraphics[width=0.8\linewidth]{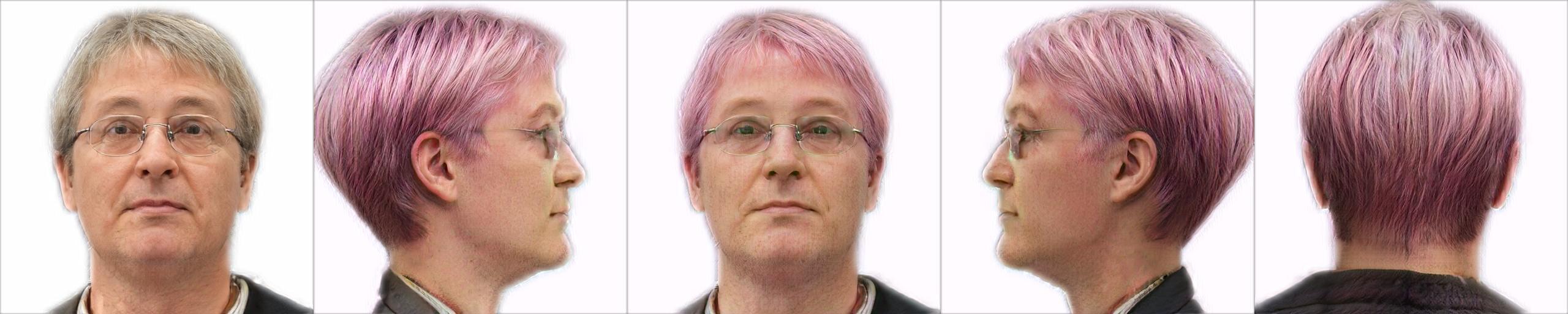}

\end{tabular}
}
\caption{Additional results with various prompts applied to the input image given in the first column.}
\label{fig:other_prompts}
\vspace{-0.3cm}
\end{figure}

\noindent \textbf{Results with other prompts.} We present additional results with various prompts in~\cref{fig:other_prompts}. These prompts include local stylization, such as adding a mustache or editing hair to pink. {For these experiments, we do not employ attribute-based masking for score gradients (for example, we do not mask the face when editing the hair).} Despite this, our method successfully achieves such transformations.%

\section{Conclusion}
We introduced a novel approach to 3D head stylization that improves identity preservation and stylization quality. By utilizing the PanoHead model for 360-degree image synthesis and incorporating negative log-likelihood distillation (LD), mirror and grid score gradients, and score rank weighing, our method overcomes the limitations of previous approaches that struggled with identity retention. Experimental results show significant qualitative and quantitative improvements, advancing the state of 3D head stylization. Our work also provides valuable insights into effective distillation between diffusion models and GANs, emphasizing the importance of ID preservation in stylization.

\begin{figure}[ht!]
\scriptsize
\centering
\setlength\tabcolsep{0.3pt}
\begin{tabular}{cc}
Input \& Joker & Input \& Sketch
\\
\includegraphics[width=0.2\textwidth]{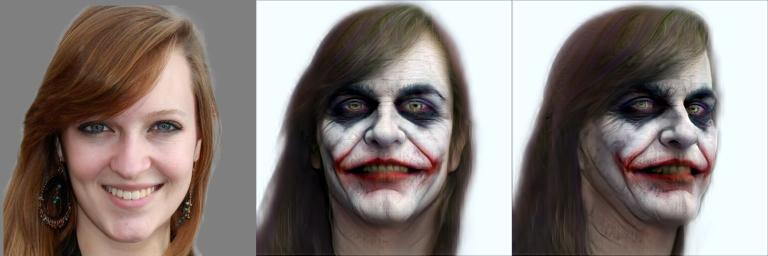}&
\includegraphics[width=0.2\textwidth]{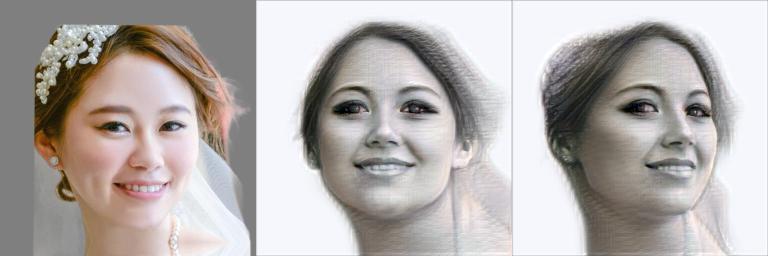}
\end{tabular}
\vspace{-0.3cm}
\caption{Limitations in preserving earrings and hair accessories.}
\label{fig:rebuttal_failure}
\vspace{-0.3cm}
\end{figure}

\noindent \textbf{Limitations and future work.}
A limitation, similar to other methods, is that stylization may not fully preserve earrings and hair accessories (\cref{fig:rebuttal_failure}). Another one is the output resolution, which is bounded by the utilized generative model. However, they can be mitigated by utilizing a heavier model. For the future work, we will extend our approach of domain adaptation to class-agnostic, large reconstruction models. We believe our findings for such pre-trained models will unlock new research directions. %

\noindent \textbf{Acknowledgements.} We acknowledge EuroHPC Joint Undertaking for awarding the project ID EHPC-AI-2024A02-031 access to Leonardo at CINECA, Italy.

\clearpage
\setcounter{page}{1}
\setcounter{figure}{0}
\setcounter{table}{0}
\setcounter{section}{0}
\setcounter{equation}{0}
\maketitlesupplementary
\definecolor{mygray}{RGB}{30, 30, 30}
\definecolor{mygreen}{RGB}{179, 200, 156}
\definecolor{myyellow}{RGB}{255, 228, 161}

\appendix
\section{LD objective}
This section will go through the detailed derivation.

Recall the DDPM forward process:
\begin{equation}
    \sqrt{\Bar{\alpha_t}}x_0 + \sqrt{1-\Bar{\alpha_t}}n = x_t, \quad n \sim \mathcal{N}(0,\mathbf{I})
    \label{eqn:der0}
\end{equation}

Assume that distribution ($q$) of the 3D representation ($\theta$) conditioned on generation prompt ($y$) is proportional to the prompt-conditioned distribution ($p$) of independent 2D renders ($x_0^i$) on different poses ($i$). In our setup, $\theta$ is the style-based 3D GAN layers.

\begin{equation}
q(\theta|y) \propto p(x_0^0,x_0^1,...,x_0^N|y) = \prod_i^N p(x_0^i|y)
\label{eqn:der1}
\end{equation}

We optimize negative log-likelihood of~\cref{eqn:der1} to find $\theta$:
\begin{equation}
-\log q(\theta|y) = -\log \prod_i^N p(x_0^i|y) = - \sum_i^N \log p(x_0^i|y)
\label{eqn:der2}
\end{equation}
Define the loss $\mathcal{L}_\text{LD}$ as the average of infinitely many $N$ render poses and find gradient $\nabla_\theta$ to update $\theta$ via gradient descent, where $\pi$ is any given pose:

\begin{equation}
    \begin{aligned}
&\mathcal{L}_\text{LD} = -\frac{1}{N}\lim_{N\rightarrow\infty}\log q(\theta|y)  = -\mathbb{E}_\pi\{\log p(x_0^\pi|y)\} \\
&    \nabla_\theta \mathcal{L}_\text{LD} = -\mathbb{E}_\pi\{\nabla_\theta \log p(x_0^\pi|y)\}
\end{aligned}
\label{eqn:der3}
\end{equation}

Using~\cref{eqn:der3,eqn:der0} and change of variables in probability:
\begin{equation}
p(x_0^\pi|y) = p(x_t^\pi|y) \abs{\frac{\partial x_t}{\partial x_0}}^{-1} = \frac{p(x_t^\pi|y)}{\sqrt{\Bar{\alpha_t}}}
\label{eqn:der4}
\end{equation}

Take the log of both sides, the partial derivative with respect to $x_0^\pi$, and decompose the right-hand side with chain rule using the relation in~\cref{eqn:der0}:
\begin{equation}
\begin{aligned}
\log p(x_0^\pi|y) = \log p(x_t^\pi|y) - \log \sqrt{\Bar{\alpha_t}}
\\
\frac{\partial \log p(x_0^\pi|y)}{\partial x_0^\pi} = \frac{\partial p(x_t^\pi|y)}{\partial x_t^\pi} \frac{\partial x_t^\pi}{\partial x_0^\pi}
\\
\nabla_{x_0} \log p(x_0^\pi|y) = \nabla_{x_t} \log p(x_t^\pi|y) \frac{\partial x_t^\pi}{\partial x_0^\pi}
\label{eqn:der5}
\end{aligned}
\end{equation}

Extend the partial gradient chain in~\cref{eqn:der5} to $\theta$ from $x_0^\pi$:
\begin{equation}
\nabla_\theta \log p(x_0^\pi|y) = \nabla_{x_t} \log p(x_t^\pi|y) \frac{\partial x_t^\pi}{\partial x_0^\pi} \frac{\partial x_0^\pi}{\partial \theta}
\label{eqn:der6}
\end{equation}

where $\nabla_{x_t} \log p(x_t^\pi|y)$ is the score function estimation. Plugging~\cref{eqn:der6} into~\cref{eqn:der3} yields the update direction:

\begin{equation}
\nabla_\theta \mathcal{L}_\text{LD} = -\mathbb{E}_{\pi,x_t}\{ \nabla_{x_t} \log p(x_t^\pi|y) \frac{\partial x_t^\pi}{\partial x_0^\pi} \frac{\partial x_0^\pi}{\partial \theta} \}
\label{eqn:der7}
\end{equation}

where $\frac{\partial x_t^\pi}{\partial x_0^\pi}$ is $\sqrt{\Bar{\alpha_t}}$ from~\cref{eqn:der0}. Notice that to update $\theta$, we do not need to back-propagate through denoising UNet and can acknowledge the UNet output as a part of the gradient.~\cref{alg:alg1} describes the domain adaptation procedure with PyTorch nomenclature:

\begin{algorithm}
\footnotesize
\caption{LD with mirror and grid grads}
\begin{algorithmic}[1]
\Require Generator $\mathbf{G}_\theta$, neural renderer $\mathbf{R}$, super-resolver $\mathbf{SR}$, depth extractor \textbf{D}, depth and text-conditioned denoising UNet $\mathbf{SD}$, generator mapping truncation parameter $\psi$, extrinsic triplane render matrix $\pi$, mirror pose $\pi'$, vertical flip operator \textbf{M}, rank weighing matrix \textbf{W}

\begin{tcolorbox}[colback=myyellow!40!white, colframe=white, arc=3pt, outer arc=3pt, boxrule=0pt, parbox=false]
\For{$i$ in $\{0,1,...,N\}$}
    \State $w^+ \leftarrow \texttt{sample\_latent($\psi$=0.8)}$
    \State $\pi,\pi' \leftarrow \texttt{sample\_pose()}$ \Comment{$\mathbb{E}_{\pi}$}
    \State $x_0^\pi,x_0^{\pi'} \leftarrow \mathbf{SR}(\mathbf{R}(\mathbf{G}_\theta(z),\pi,\pi'))$
    \State $n,t \leftarrow \texttt{noise\_scheduler(0.70,0.96)}$
    \State $x_t^\pi \leftarrow \sqrt{\Bar{\alpha_t}}x_0^\pi + \sqrt{1-\Bar{\alpha_t}}n $ \Comment{$\mathbb{E}_{x_t}$}
    \State \texttt{with no\_grad():}
    \State \hskip1.5em $\nabla_{x_t} \log p(x_t^\pi|y) / \sqrt{1-\Bar{\alpha_t}} \leftarrow \mathbf{SD}(x_t^\pi, y, t, \mathbf{D}(x_0^\pi))$
    \State $\texttt{grad} \leftarrow \nabla_{x_t} \log p(x_t^\pi|y) \sqrt{\Bar{\alpha_t}}$ 
    \State $\mathbf{U \Sigma V^\text{T}} \leftarrow \mathbf{SVD}\texttt{(grad)}$ 
    \State $\texttt{grad} \leftarrow \mathbf{U W \Sigma V^\text{T}}$ \Comment{rank weighing}
    \State $x_0^\pi$\texttt{.backward(grad)} \Comment{$\nabla_\theta \mathcal{L}_\text{LD}$}
    \State $x_0^{\pi'}$\texttt{.backward(}\textbf{M}\texttt{(grad))} \Comment{mirror gradients}
    \State \texttt{optimizer.step()}
\EndFor
\end{tcolorbox}

\begin{tcolorbox}[colback=mygreen!40!white, colframe=white, arc=3pt, outer arc=3pt, boxrule=0pt, parbox=false]
\For{$i$ in $\{0,1,...,N\}$}
    \State $w^+ \leftarrow \texttt{sample\_latent($\psi$=0.8)}$
    \State $\{\pi\} = \pi^0,\pi^1,\pi^2,\pi^3 \leftarrow \texttt{sample\_pose()}$ \Comment{$\mathbb{E}_{\{\pi\}}$}
    \State $\{x_0^\pi\}_\text{LR} \leftarrow \texttt{make\_grid(}\mathbf{R}(\mathbf{G}_\theta(w^+),\{\pi\})\texttt{)}$
    \State $\{x_0^\pi\} \leftarrow \texttt{make\_grid(}\mathbf{SR}(\mathbf{R}(\mathbf{G}_\theta(w^+),\{\pi\}))\texttt{)}$
    \State $n,t \leftarrow \texttt{noise\_scheduler(0.30,0.80)}$
    \State $\{x_t^\pi\} \leftarrow \sqrt{\Bar{\alpha_t}}\{x_0^\pi\} + \sqrt{1-\Bar{\alpha_t}}n $ \Comment{$\mathbb{E}_{\{x_t\}}$}
    \State \texttt{with no\_grad():}
    \State \hskip1.5em $\nabla_{\{x_t\}} \log p(\{x_t^\pi\}|y) / \sqrt{1-\Bar{\alpha_t}} \leftarrow \mathbf{SD}(\{x_t^\pi\}, y, t, \mathbf{D}(\{x_0^\pi\}))$
    \State $\texttt{grad} \leftarrow \nabla_{\{x_t\}} \log p(\{x_t^\pi\}|y) \sqrt{\Bar{\alpha_t}}$
    \State $\mathbf{U \Sigma V^\text{T}} \leftarrow \mathbf{SVD}\texttt{(grad)}$ 
    \State $\texttt{grad} \leftarrow \mathbf{U W \Sigma V^\text{T}}$ \Comment{rank weighing}
    \State $\{x_0^\pi\}_\text{LR}$\texttt{.backward(grad)} \Comment{grid gradients $\nabla_\theta \mathcal{L}_{\text{LD}_g}$}
    \State \texttt{optimizer.step()}
\EndFor
\end{tcolorbox}

\State \textbf{return}  $\mathbf{G}_\theta$
\end{algorithmic}
\label{alg:alg1}
\end{algorithm}

\texttt{sample\_latent} utilizes the mapping network of the generator and maps $z$ to $w^+$, later to be fed to the generator. \texttt{make\_grid} creates a 2$\times$2 grid with 4 inputs. \textbf{M} is realized with \texttt{torch.flip(x,dims=[-1])}. \texttt{with no\_grad()} disables PyTorch's gradient calculation. Note that each time $x_0$ is generated, we implicitly pass it through VAE to embed it into \textbf{SD}'s latent space.

\section{Implementation details}

\textbf{Baselines.} We train the latent mapper in StyleCLIP~\cite{Patashnik_2021_ICCV} with PanoHead's~\cite{an2023panohead} generator. For StyleGAN-NADA~\cite{gal2022stylegan} and StyleGANFusion~\cite{song2022diffusion}, we use~\cite{song2022diffusion}'s official repository and modify the generator backbone to PanoHead. For~\cite{song2022diffusion}, we utilize their EG3D config parameters for PanoHead, and implement the adaptive layer selection for~\cite{gal2022stylegan}. For DiffusionGAN3D~\cite{lei2023diffusiongan3d}, we implement the method based on the official paper since there is no published codebase. For our baseline, we utilize our implementation of~\cite{lei2023diffusiongan3d} with their distance loss for domain adaptation and build upon it with our proposed improvements. We stay faithful to each baseline's original hyperparameters (denoiser checkpoint selection, noise scheduler, learning rate, optimizer, etc.) unless the training diverges.

\textbf{Our training parameters.} We train the generator with synthetic $z_{1\times512}\sim\mathcal{N}(0,\mathbf{I})$ data for 10k iterations with batch size 1, where the truncation parameter of the generator's mapping network is $\psi=0.8$. We use Adam optimizer with a $1e^{-4}$ learning rate. We optimize the \texttt{G.backbone.synthesis} and \texttt{G.backbone.superresolution} sub-networks of the generator \texttt{G} and freeze all convolutional layer biases, using the same configuration as~\cite{song2022diffusion}. The classifier-free-guidance (CFG)~\cite{ho2022classifierfreediffusionguidance} weight and depth-conditioned ControlNet~\cite{Zhang_2023_ICCV}\footnote{\url{https://huggingface.co/lllyasviel/sd-controlnet-depth}} guidance weight are set to 7.5 and 1.0, respectively. Depth ground truths are extracted from~\cite{depth_anything_v2} since the neural renderer's depth estimations are low-resolution and require additional clipping ($64\times64$).

As the conditional denoiser for our method and the ablation study for showing the improvements upon~\cite{lei2023diffusiongan3d}, we employ RV v5.1\footnote{\url{https://huggingface.co/SG161222/Realistic_Vision_V5.1_noVAE}}. For qualitative and quantitative comparison among other methods, we employ the methods' suggested diffusion checkpoints in their papers and repositories\footnote{\url{https://huggingface.co/stable-diffusion-v1-5/stable-diffusion-v1-5}}\footnote{\url{https://huggingface.co/stabilityai/stable-diffusion-2}}.

For mirror and grid denoising, noise start timestep $t$ is uniformly selected among \texttt{(0.70,0.96)} and \texttt{(0.30,0.80)}, respectively, where $t$ is from $0\rightarrow1$. We use \texttt{DDIMScheduler} for the noise scheduler. The number of inference steps in the diffusion pipeline is always 1 since we perform score distillation.

\begin{figure}
\centering
\includegraphics[width=0.7\linewidth]{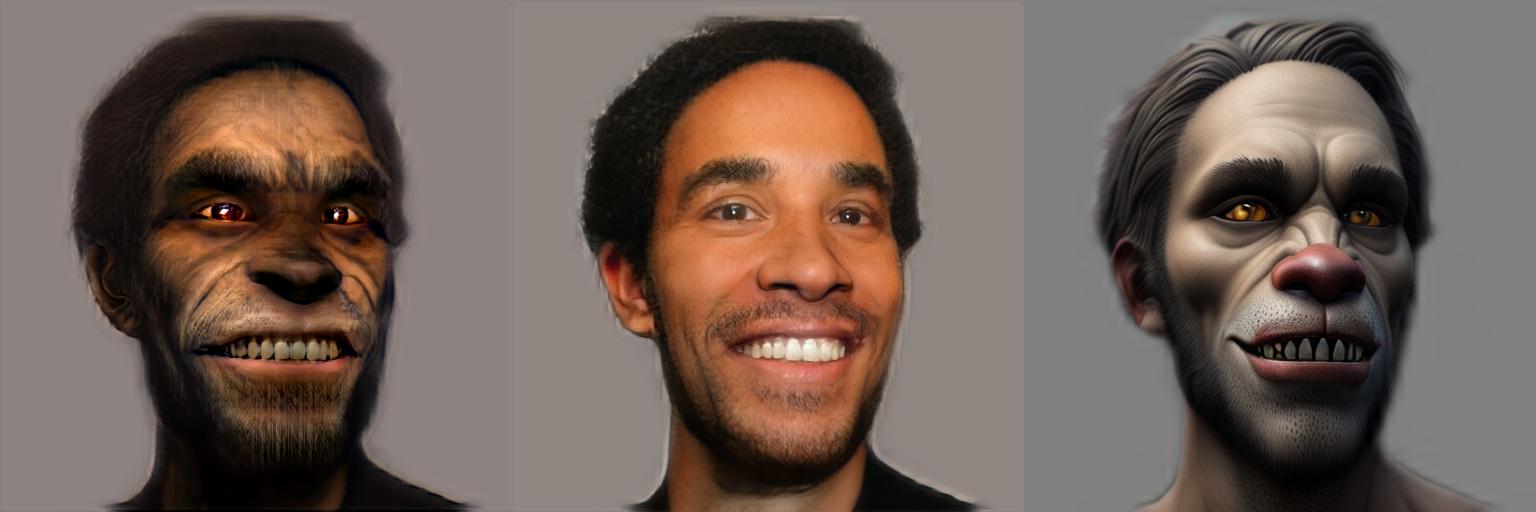}
\includegraphics[width=0.7\linewidth]{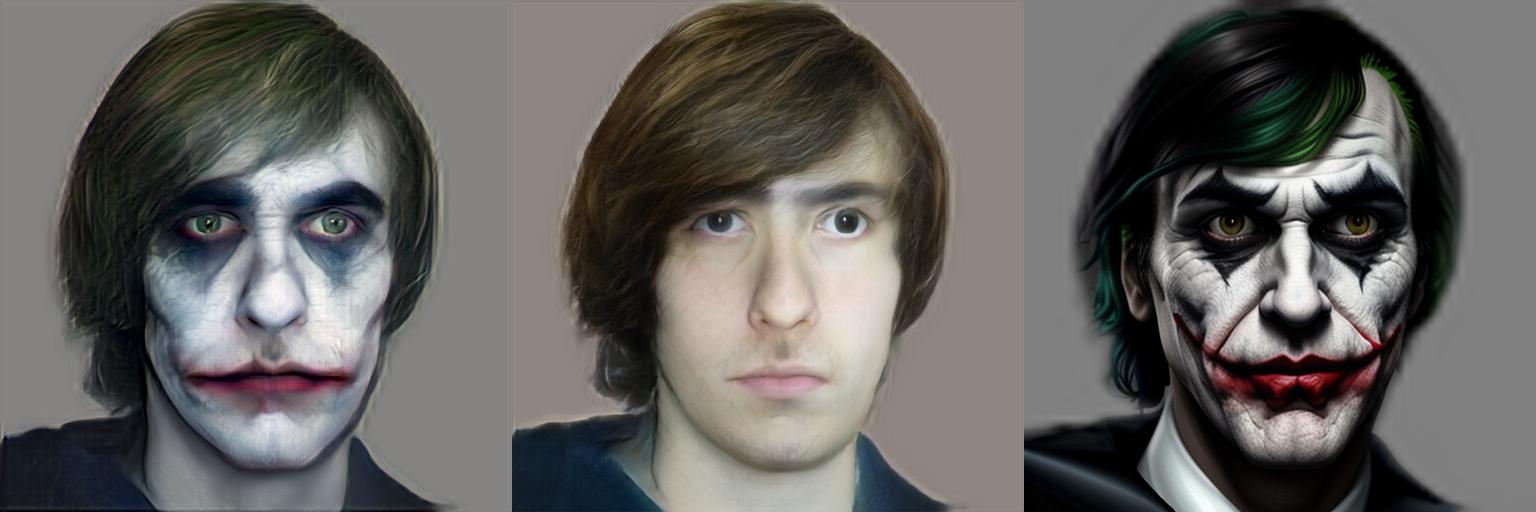}
\caption{From left to right: Ours (distribution \#2), ground truth unedited image (distribution \#3), edited image with full-step diffusion pipeline (distribution \#1).}
\label{fig:supp_quant}
\end{figure}

\textbf{Quantitative scores.} We construct ground-truth edited image distributions using the Stable Diffusion pipeline. For the first distribution, we take images, add noise with $t=25$, and denoise with the style prompt using each baseline's diffusion checkpoints for 50 steps, resulting in edited images. The second distribution consists of the same images stylized using domain-adapted generators. These two distributions are used to compute FID, and individual image pairs between them are used to compute CLIP similarity scores. We generate a third distribution using unedited images to evaluate identity preservation (ID) and $\Delta\mathcal{D}$. Scores for ID and $\Delta\mathcal{D}$ are then calculated between using image pairs from the second and third distributions.~\cref{fig:supp_quant} visualizes sample images in those three distributions.

\textbf{Prompts.} We use empty strings for negative prompts for our method. For positive prompts, we use the following list for all methods:
\begin{itemize}
    \item \texttt{Portrait a person in Pixar style, cute, big eyes, Disney, sharp, 8K, skin detail, best quality, realistic lighting, good-looking, uniform light, extremely detailed}
    \item \texttt{Portrait of a Greek statue, closeup, elegant and timeless, intricate and detailed carving, smooth marble texture, ancient Greek aesthetics}
    \item \texttt{A portrait of Joker from the movie The Dark Knight}
    \item \texttt{Charcoal pencil sketch of human face, lower third, high contrast, black and white}
    \item \texttt{Portrait of a werewolf}
    \item \texttt{Portrait of a zombie}
\end{itemize}

\begin{figure}[t!]
\centering
\scriptsize
\setlength{\tabcolsep}{1pt}
\scalebox{1.0}{
\begin{tabular}{ccccccc}
& Input & \shortstack{$\mathbf{\Sigma}$-weighted} & \shortstack{$k=1$} & \shortstack{$k=2$} & \shortstack{$k=3$} & \shortstack{$k=4$} \\

\raisebox{0.2in}{\rotatebox[origin=t]{90}{Caricature}}&
\includegraphics[width=0.15\linewidth]{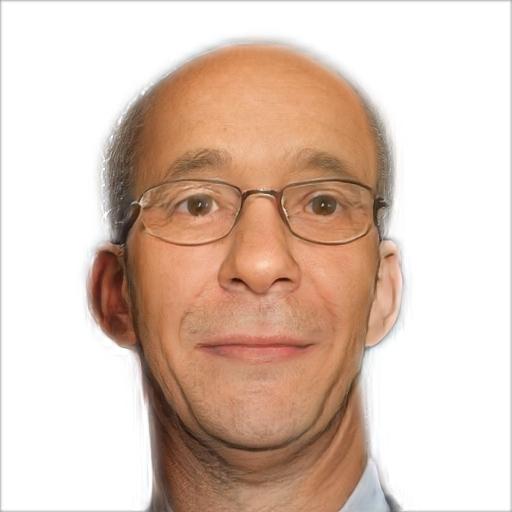}&
\includegraphics[width=0.15\linewidth]{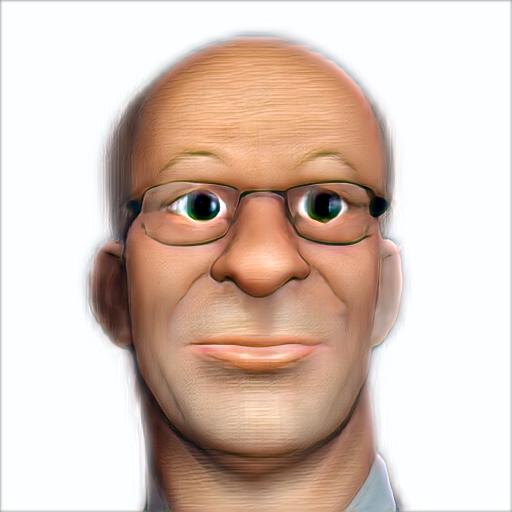}&
\includegraphics[width=0.15\linewidth]{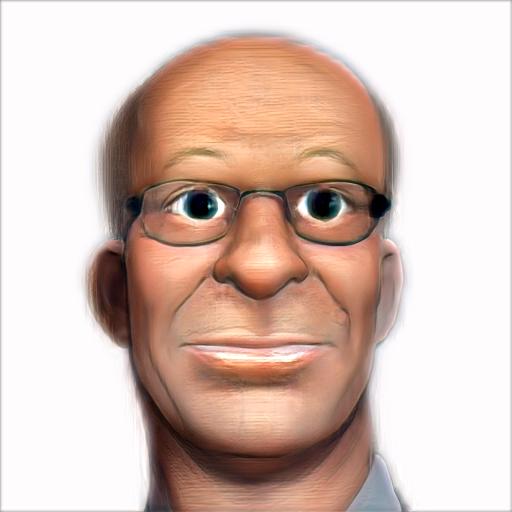}&
\includegraphics[width=0.15\linewidth]{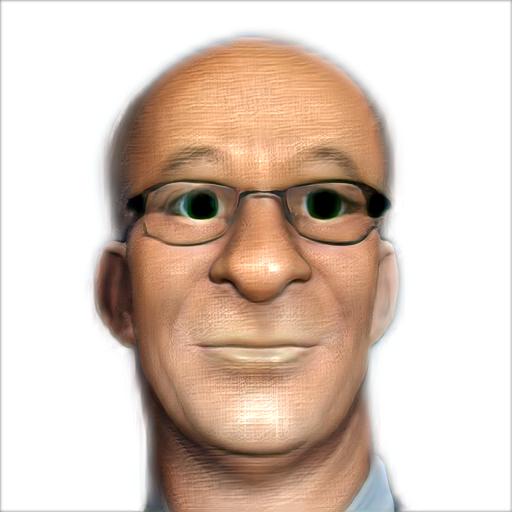}&
\includegraphics[width=0.15\linewidth]{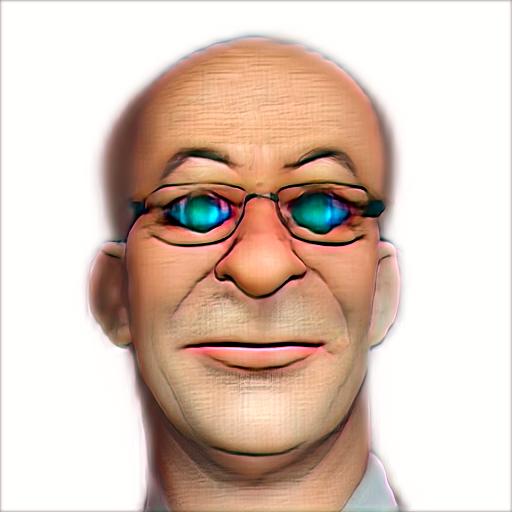}&
\includegraphics[width=0.15\linewidth]{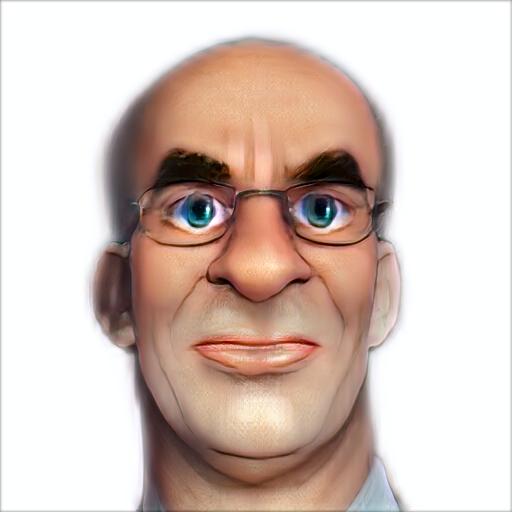}
\\
\raisebox{0.2in}{\rotatebox[origin=t]{90}{Charcoal}}&
\includegraphics[width=0.15\linewidth]{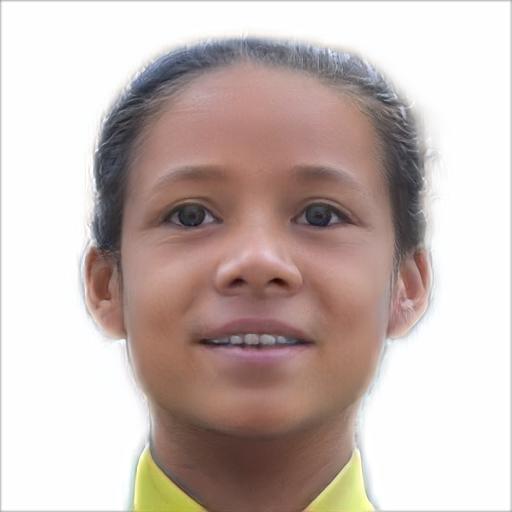}&
\includegraphics[width=0.15\linewidth]{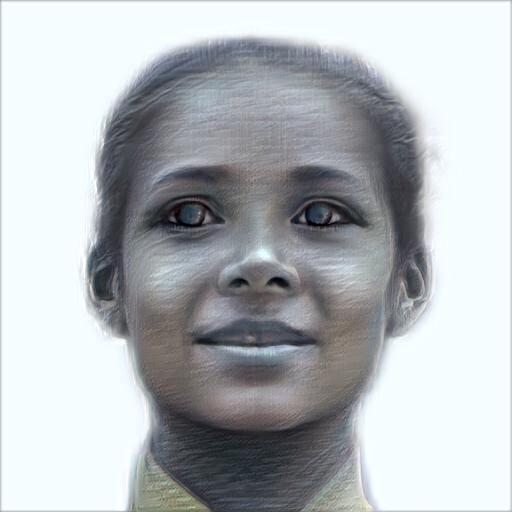}&
\includegraphics[width=0.15\linewidth]{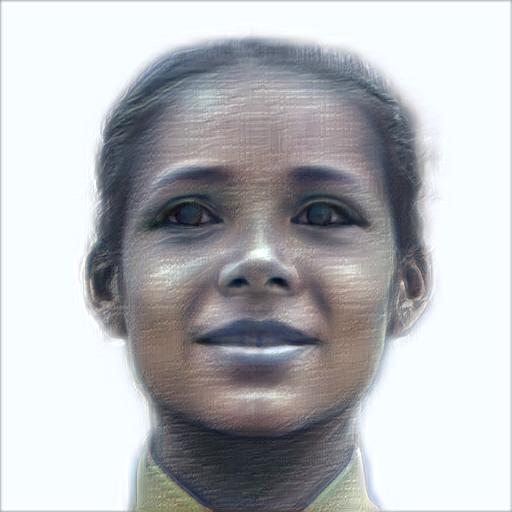}&
\includegraphics[width=0.15\linewidth]{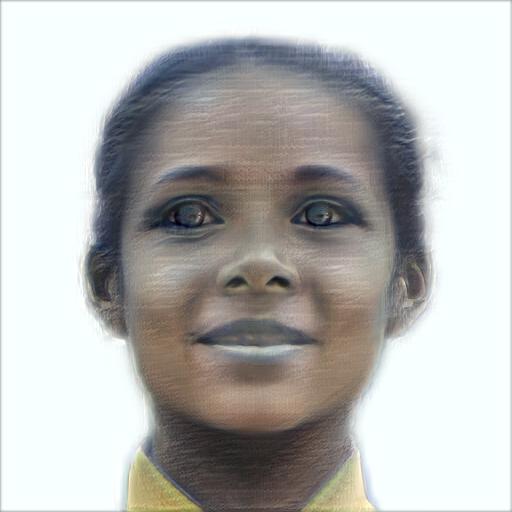}&
\includegraphics[width=0.15\linewidth]{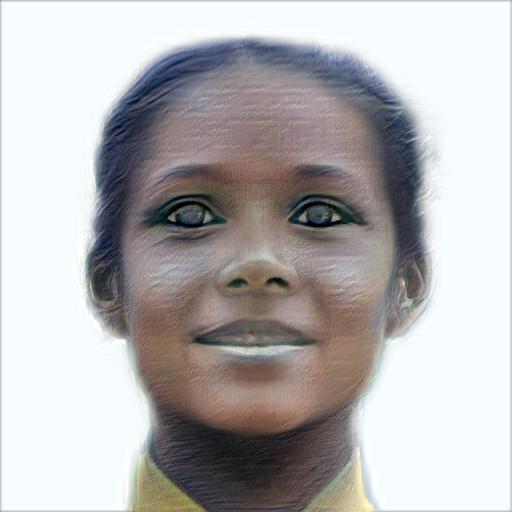}&
\includegraphics[width=0.15\linewidth]{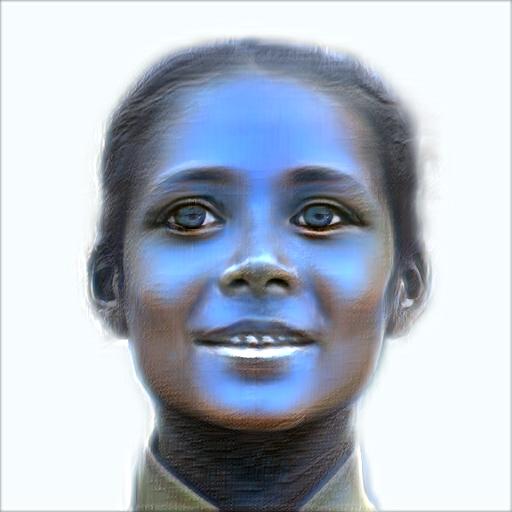}
\\
\raisebox{0.2in}{\rotatebox[origin=t]{90}{Statue}}&
\includegraphics[width=0.15\linewidth]{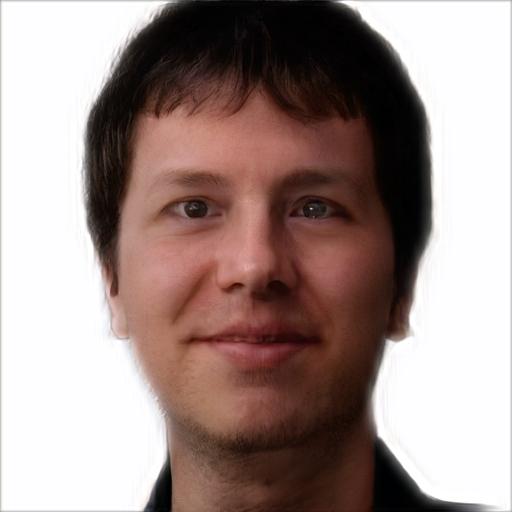}&
\includegraphics[width=0.15\linewidth]{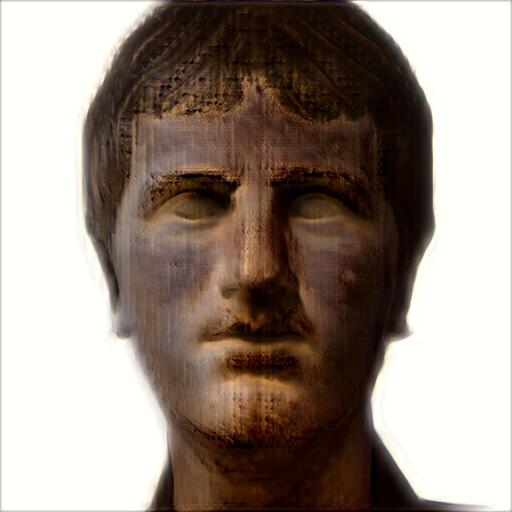}&
\includegraphics[width=0.15\linewidth]{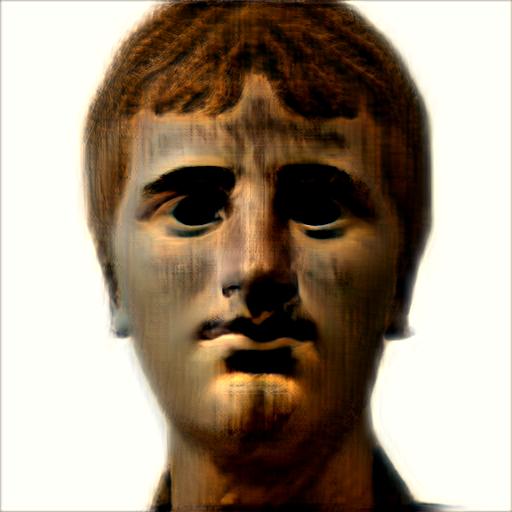}&
\includegraphics[width=0.15\linewidth]{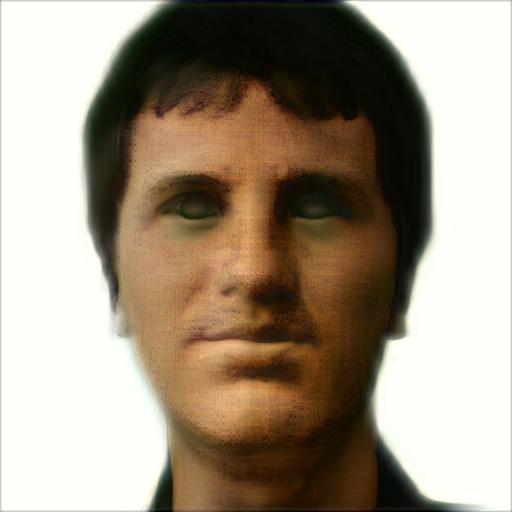}&
\includegraphics[width=0.15\linewidth]{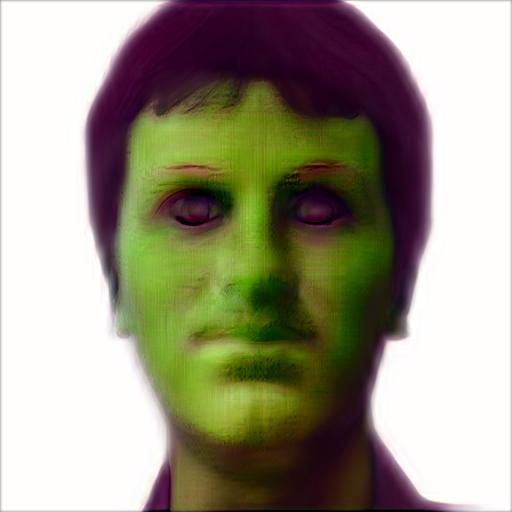}&
\includegraphics[width=0.15\linewidth]{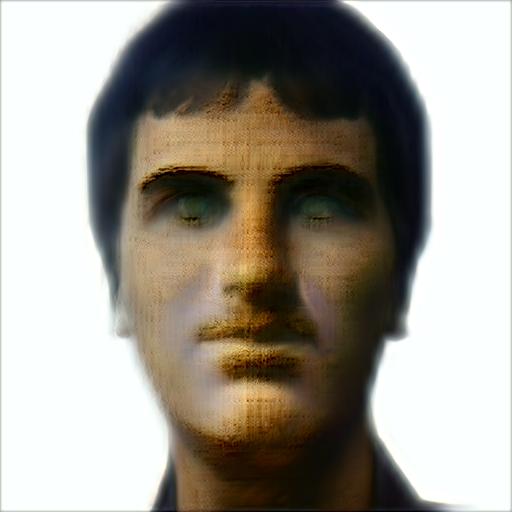}
\\
\raisebox{0.2in}{\rotatebox[origin=t]{90}{Oil Painting}}&
\includegraphics[width=0.15\linewidth]{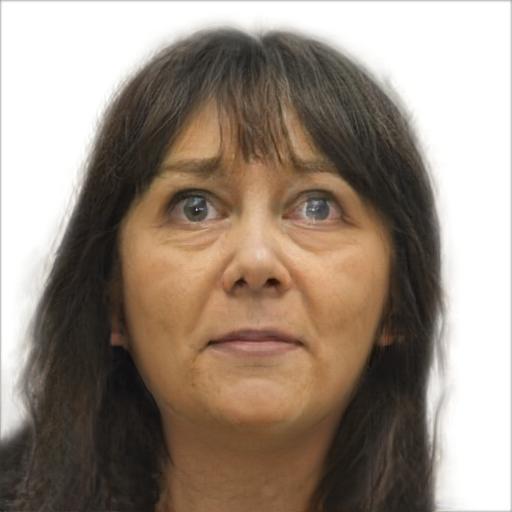}&
\includegraphics[width=0.15\linewidth]{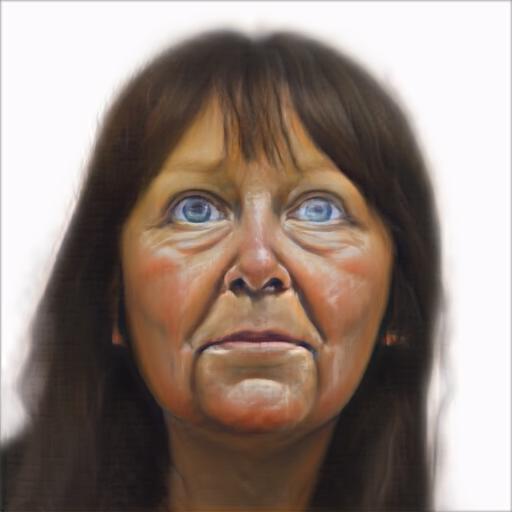}&
\includegraphics[width=0.15\linewidth]{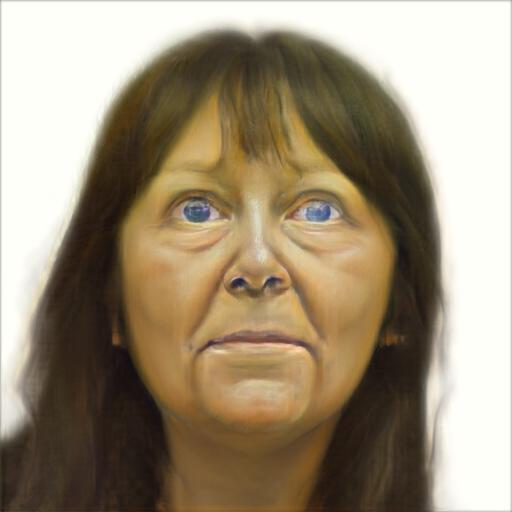}&
\includegraphics[width=0.15\linewidth]{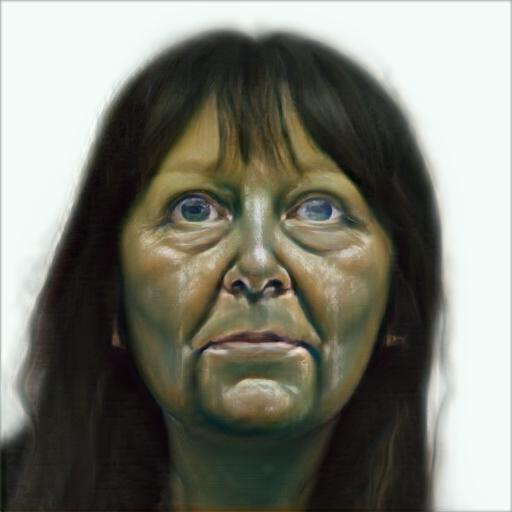}&
\includegraphics[width=0.15\linewidth]{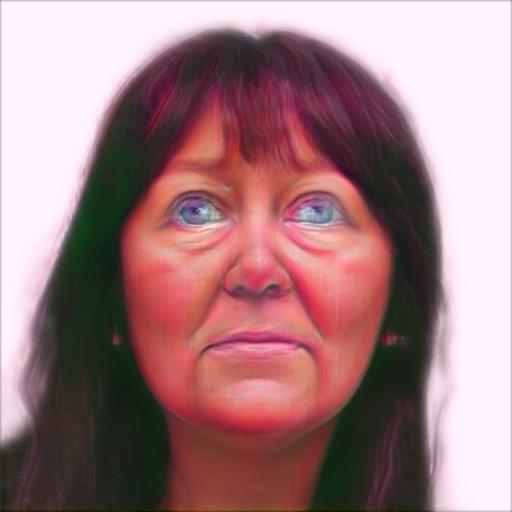}&
\includegraphics[width=0.15\linewidth]{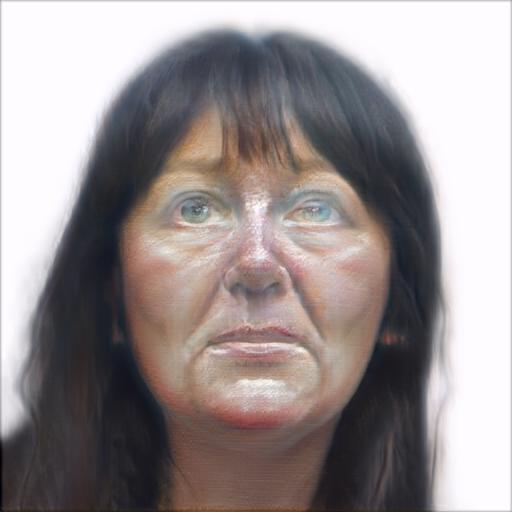}
\\
\raisebox{0.2in}{\rotatebox[origin=t]{90}{Pixar}}&
\includegraphics[width=0.15\linewidth]{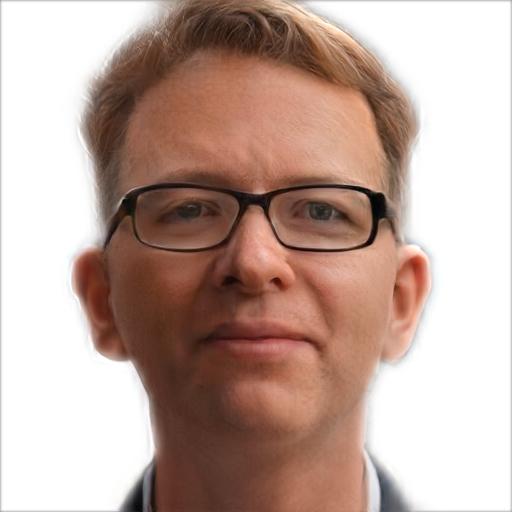}&
\includegraphics[width=0.15\linewidth]{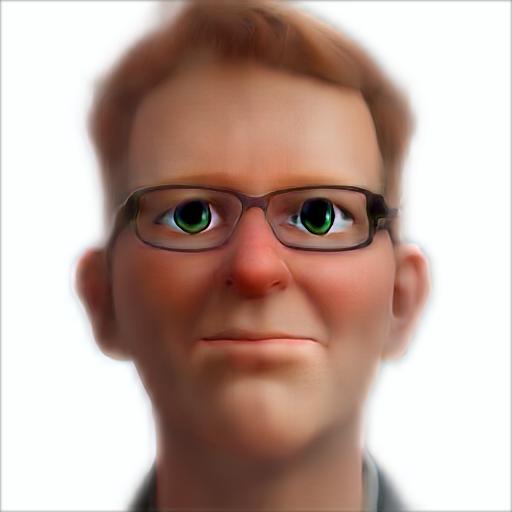}&
\includegraphics[width=0.15\linewidth]{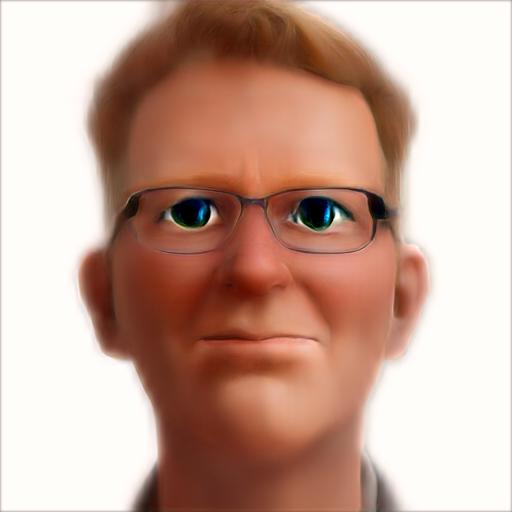}&
\includegraphics[width=0.15\linewidth]{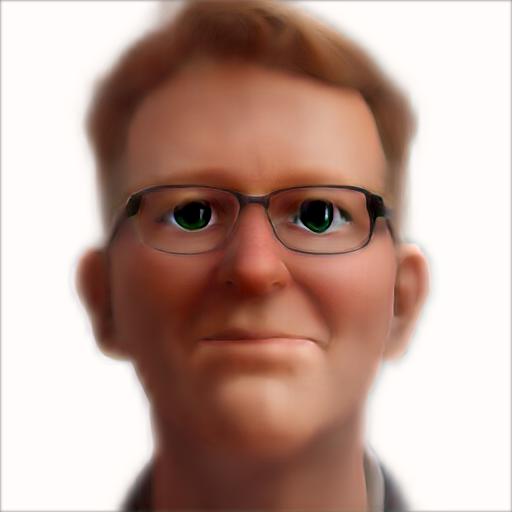}&
\includegraphics[width=0.15\linewidth]{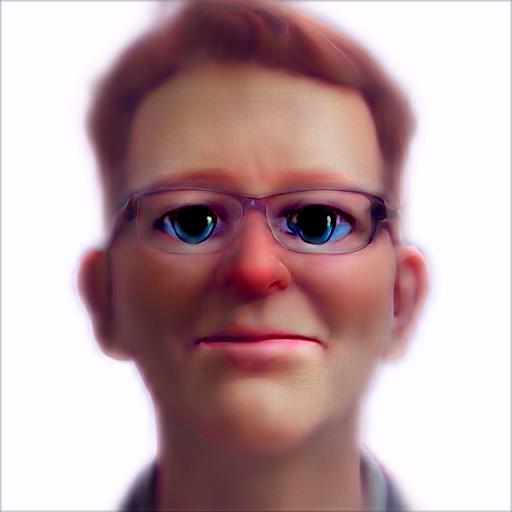}&
\includegraphics[width=0.15\linewidth]{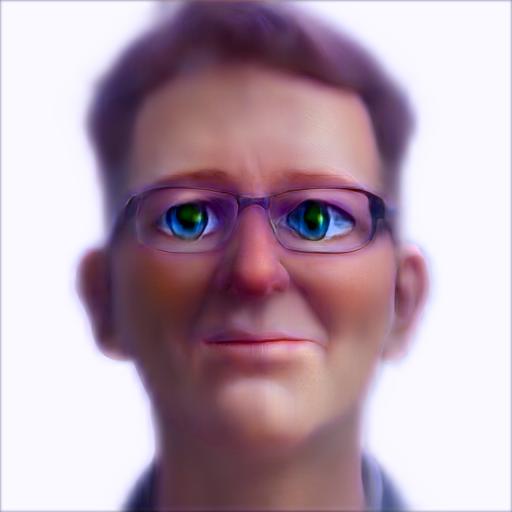}
\\
\raisebox{0.2in}{\rotatebox[origin=t]{90}{Werewolf}}&
\includegraphics[width=0.15\linewidth]{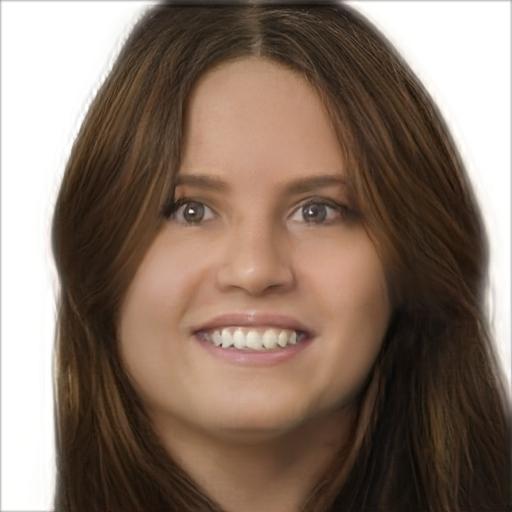}&
\includegraphics[width=0.15\linewidth]{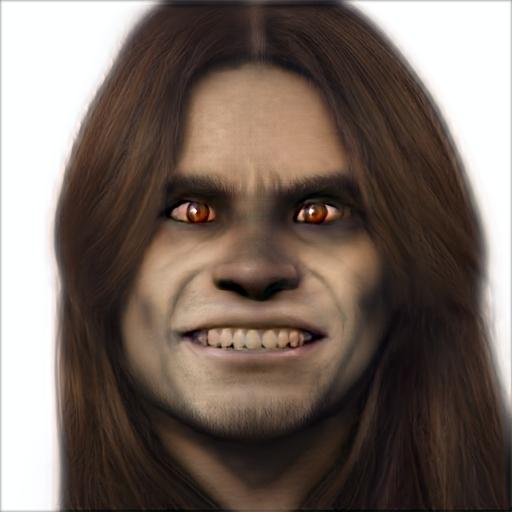}&
\includegraphics[width=0.15\linewidth]{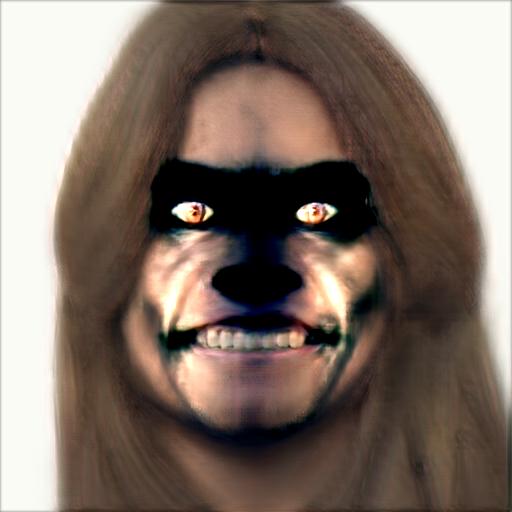}&
\includegraphics[width=0.15\linewidth]{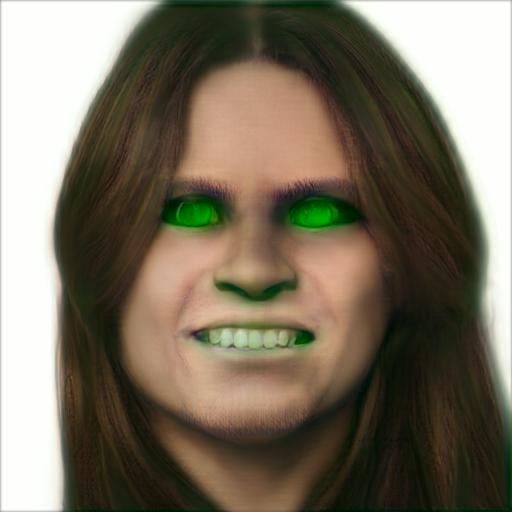}&
\includegraphics[width=0.15\linewidth]{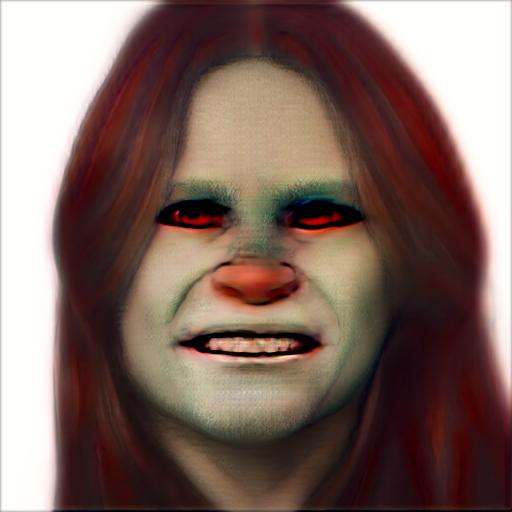}&
\includegraphics[width=0.15\linewidth]{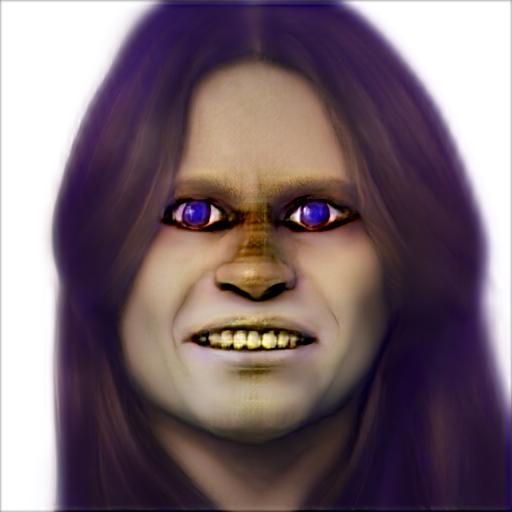}
\\
\raisebox{0.2in}{\rotatebox[origin=t]{90}{Zombie}}&
\includegraphics[width=0.15\linewidth]{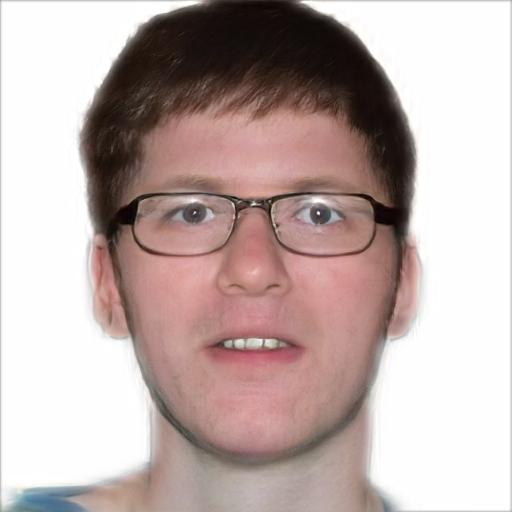}&
\includegraphics[width=0.15\linewidth]{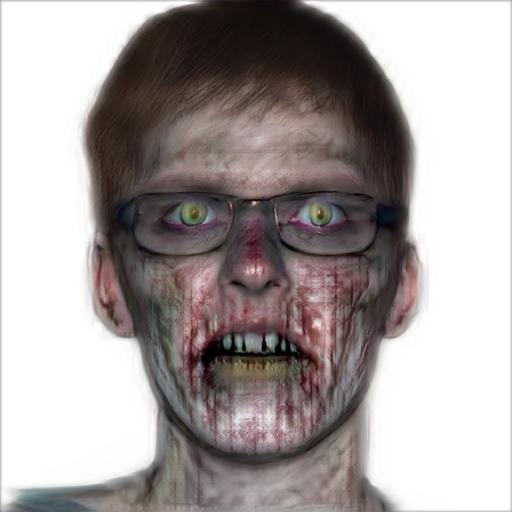}&
\includegraphics[width=0.15\linewidth]{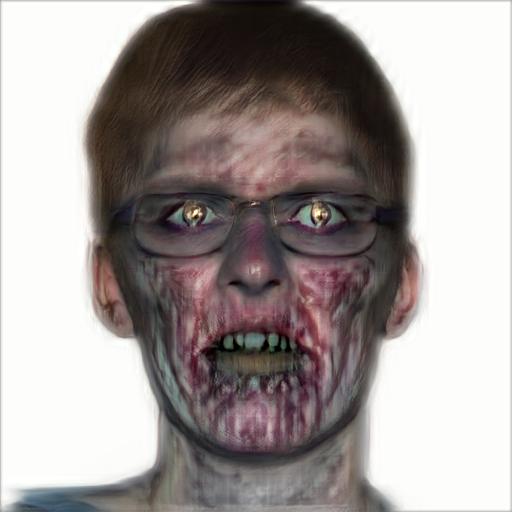}&
\includegraphics[width=0.15\linewidth]{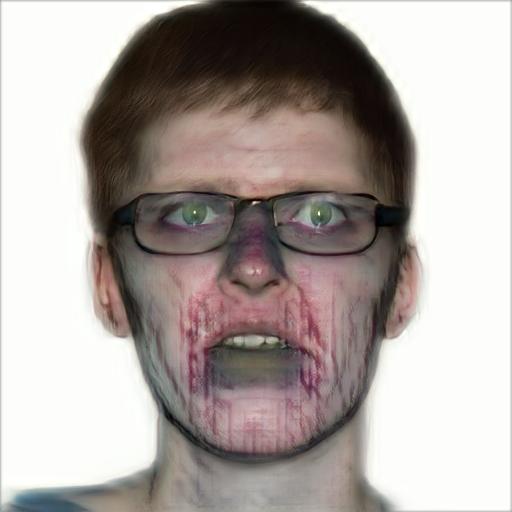}&
\includegraphics[width=0.15\linewidth]{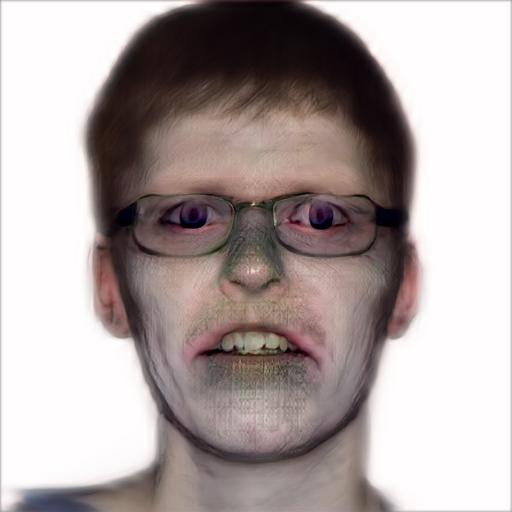}&
\includegraphics[width=0.15\linewidth]{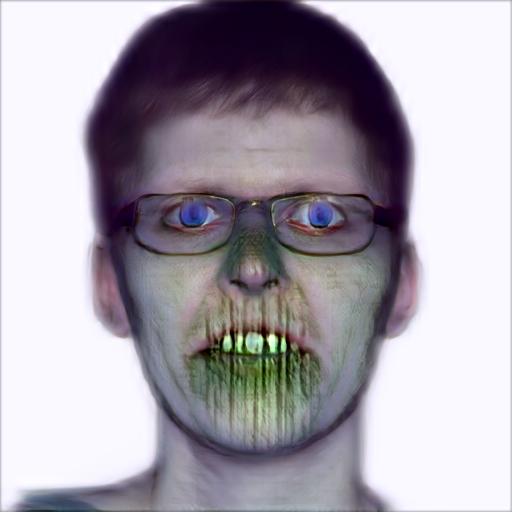}
\\
\end{tabular}
}
\caption{Demonstration of our SVD‐based stylization across multiple face styles. The Input column shows the original images. $\Sigma$‐weighted uses all top singular values with decreasing weights, preserving both coarse and fine features. Columns k=1 through k=4 depict rank‐k approximations; as k increases, more high‐frequency details are retained, resulting in sharper, more faithful stylizations.}
\label{fig:ablation_svd_ranks}
\end{figure}

\textbf{Rank weighing on score tensors.} \cref{fig:ablation_svd_ranks} illustrates how an SVD‐based approach can decompose a stylized portrait into coarse and fine components, and then reconstruct it at different “rank” levels (k). This progressive refinement underlines how SVD can serve as a powerful control mechanism for score distillation, letting the user dial in how many spatial “frequencies” of the style are included.

\section{User study}

\begin{figure}[t]
    \centering
\includegraphics[width=0.75\linewidth]{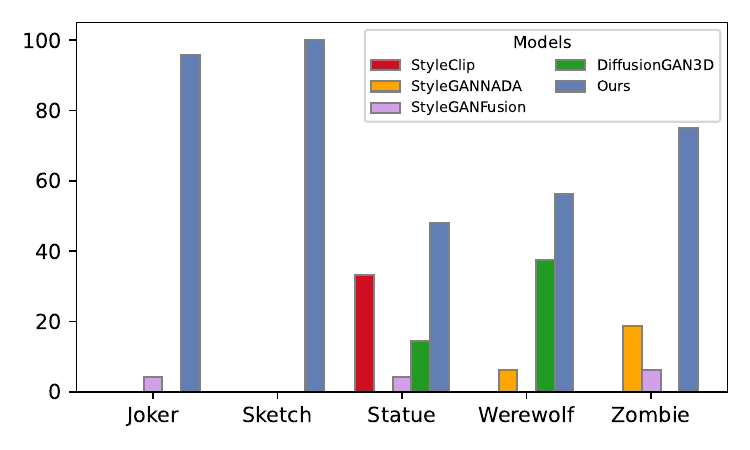}
    \caption{Percentage of user preferences. Users overwhelmingly favor ours compared to other domain adaptation methods.}
\vspace{-0.5cm}
    \label{fig:mask_plot}
\end{figure}

We conduct a user study with 25 participants to evaluate the quality of 3D stylization and identity preservation across different methods. Participants are shown images generated by five different models: StyleCLIP, StyleGAN-NADA, StyleGANFusion, DiffusionGAN3D, and our own approach. For each image, they are asked to select which output best balances stylization and identity preservation. The methods are presented in random order for each image to minimize bias. 
Results of this user study are shown in~\cref{fig:mask_plot}. The data indicate our method is consistently preferred across all the prompts tested, with participants overwhelmingly selecting it as the best for both stylization and identity preservation compared to others.

\section{Additional results}

\begin{table}[ht!]
\renewcommand{\arraystretch}{1.0}
\scriptsize
\centering
\setlength{\tabcolsep}{3.5pt}
\begin{tabular}{r|r|ccccc}
&  & {Pixar} & {Joker} & {Werewolf} & {Sketch} & {Statue} \\
\hline
\multirow{2}{*}{\rotatebox[origin=c]{90}{\scriptsize 2D}} & InstructPix2Pix   & \cellcolor{color5}\,0.1461 & \cellcolor{color4}\,0.1164 & \cellcolor{color5}\,0.1427 & \cellcolor{color2}\,0.0790 & \cellcolor{color1}\,0.0900 \\
& InstantID         & \cellcolor{color3}\,0.0897 & \cellcolor{color5}\,0.1185 & \cellcolor{color3}\,0.1055 & \cellcolor{color7}\,0.1218 & \cellcolor{color4}\,0.1290 \\
\hline
\multirow{5}{*}{\rotatebox[origin=c]{90}{\scriptsize 3D}} & StyleCLIP         & \cellcolor{color4}\,0.1045 & \cellcolor{color2}\,0.0958 & \cellcolor{color6}\,0.1962 & \cellcolor{color3}\,0.0878 & \cellcolor{color6}\,0.1658 \\
& StyleGAN-NADA     & \cellcolor{color2}\,0.0459 & \cellcolor{color6}\,0.1380 & \cellcolor{color7}\,0.2617 & \cellcolor{color4}\,0.0890 & \cellcolor{color5}\,0.1480 \\
& StyleGANFusion    & \cellcolor{color7}\,0.1668 & \cellcolor{color7}\,0.1904 & \cellcolor{color4}\,0.1387 & \cellcolor{color5}\,0.1168 & \cellcolor{color3}\,0.1212 \\
& DiffusionGAN3D    & \cellcolor{color6}\,0.1566 & \cellcolor{color3}\,0.0977 & \cellcolor{color2}\,0.0922 & \cellcolor{color6}\,0.1216 & \cellcolor{color7}\,0.2442 \\
& \textbf{Ours}     & \cellcolor{color1}\,0.0326 & \cellcolor{color1}\,0.0713 & \cellcolor{color1}\,0.0856 & \cellcolor{color1}\,0.0742 & \cellcolor{color2}\,0.1122 \\
\end{tabular}
\vspace{-0.25cm}
\caption{KID scores on the test set.}
\label{tab:kid}
\end{table}

\cref{tab:kid} reveals the KID scores on the same test set used in the main paper. Our method outperforms all baselines in KID across domains, with the exception of the Statue domain. Notably, in the Sketch stylization setting, while InstructPix2Pix reports a slightly better FID, our method achieves superior KID scores.

\begin{figure*}
\centering
\includegraphics[width=0.99\linewidth]{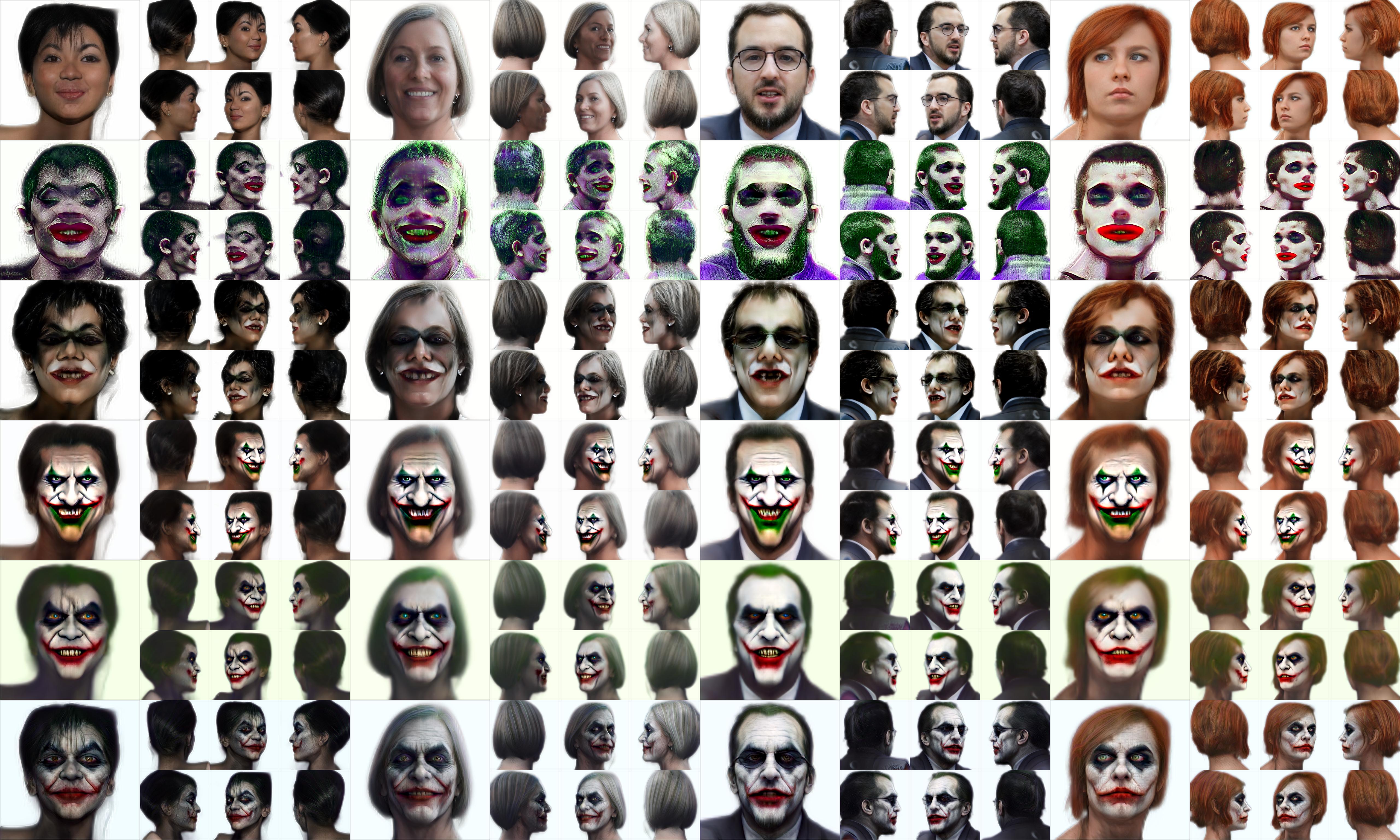}
\caption{Joker edits. From top to bottom: input, StyleCLIP, StyleGAN-NADA, StyleGANFusion, DiffusionGAN3D, ours.}
\label{fig:supp_joker}
\end{figure*}

\begin{figure*}
\centering
\includegraphics[width=0.99\linewidth]{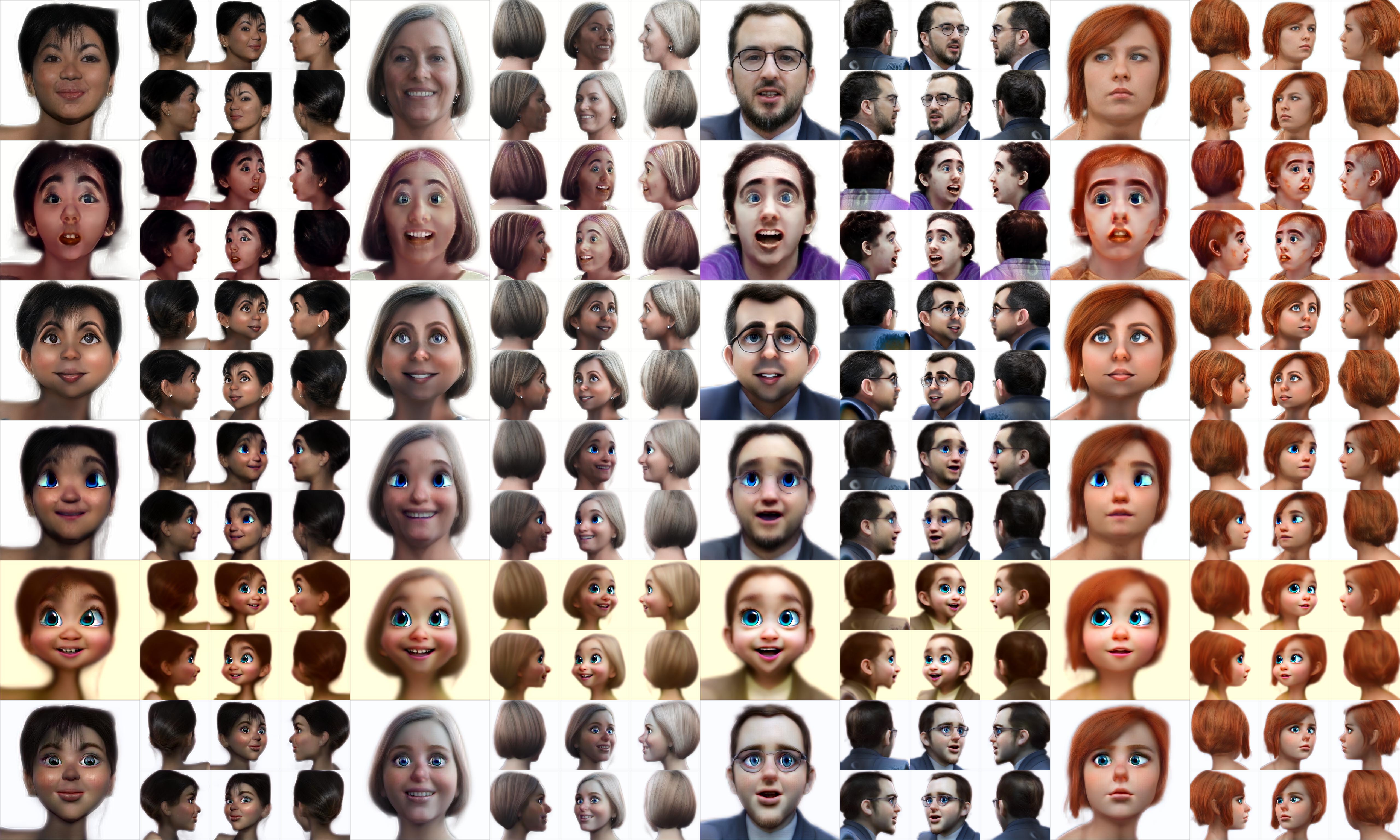}
\caption{Pixar edits. From top to bottom: input, StyleCLIP, StyleGAN-NADA, StyleGANFusion, DiffusionGAN3D, ours.}
\label{fig:supp_pixar}
\end{figure*}

\begin{figure*}
\centering
\includegraphics[width=0.99\linewidth]{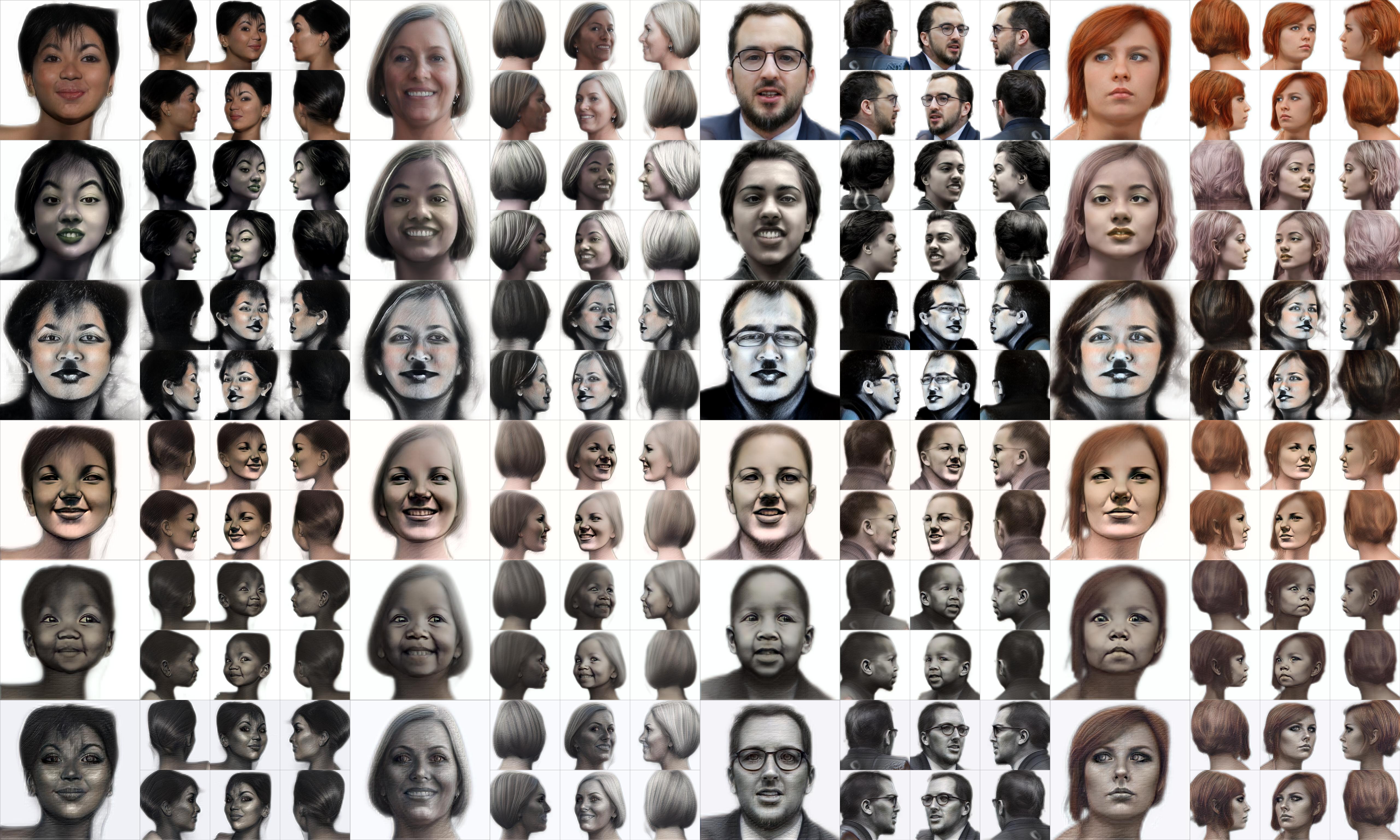}
\caption{Sketch edits. From top to bottom: input, StyleCLIP, StyleGAN-NADA, StyleGANFusion, DiffusionGAN3D, ours.}
\label{fig:supp_sketch}
\end{figure*}

\begin{figure*}
\centering
\includegraphics[width=0.99\linewidth]{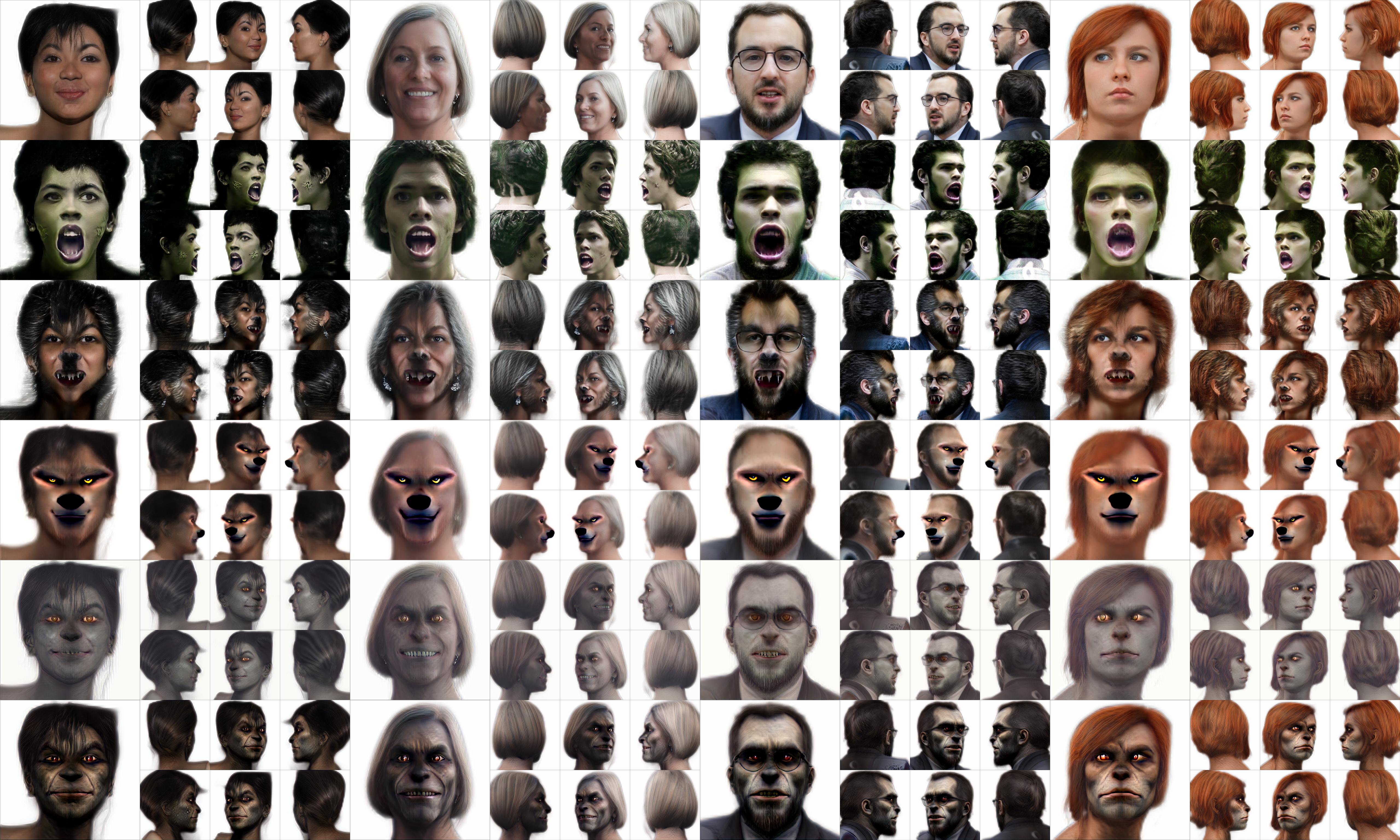}
\caption{Werewolf edits. From top to bottom: input, StyleCLIP, StyleGAN-NADA, StyleGANFusion, DiffusionGAN3D, ours.}
\label{fig:supp_werewolf}
\end{figure*}

\begin{figure*}
\centering
\includegraphics[width=0.99\linewidth]{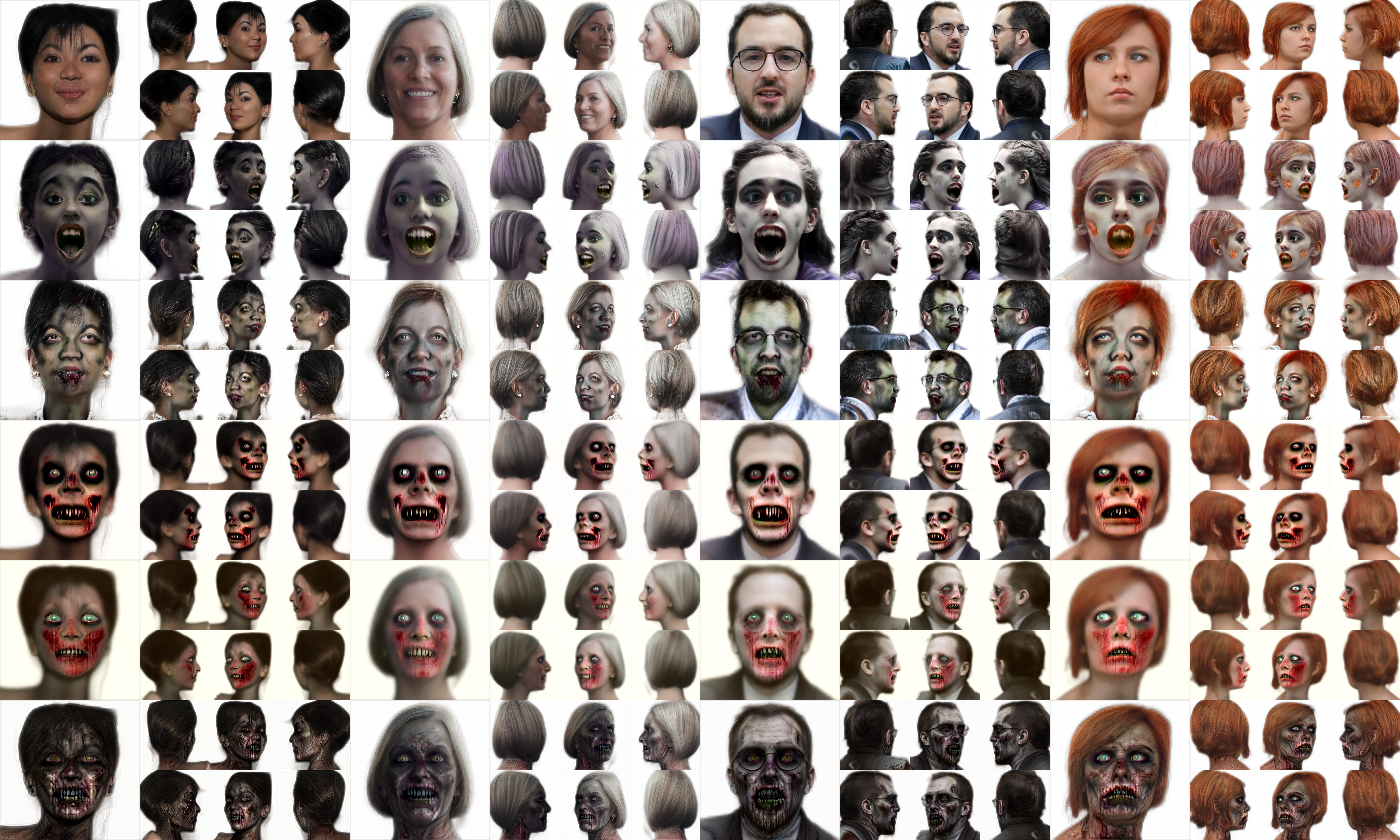}
\caption{Zombie edits. From top to bottom: input, StyleCLIP, StyleGAN-NADA, StyleGANFusion, DiffusionGAN3D, ours.}
\label{fig:supp_zombie}
\end{figure*}

\begin{figure*}
\centering
\includegraphics[width=0.99\linewidth]{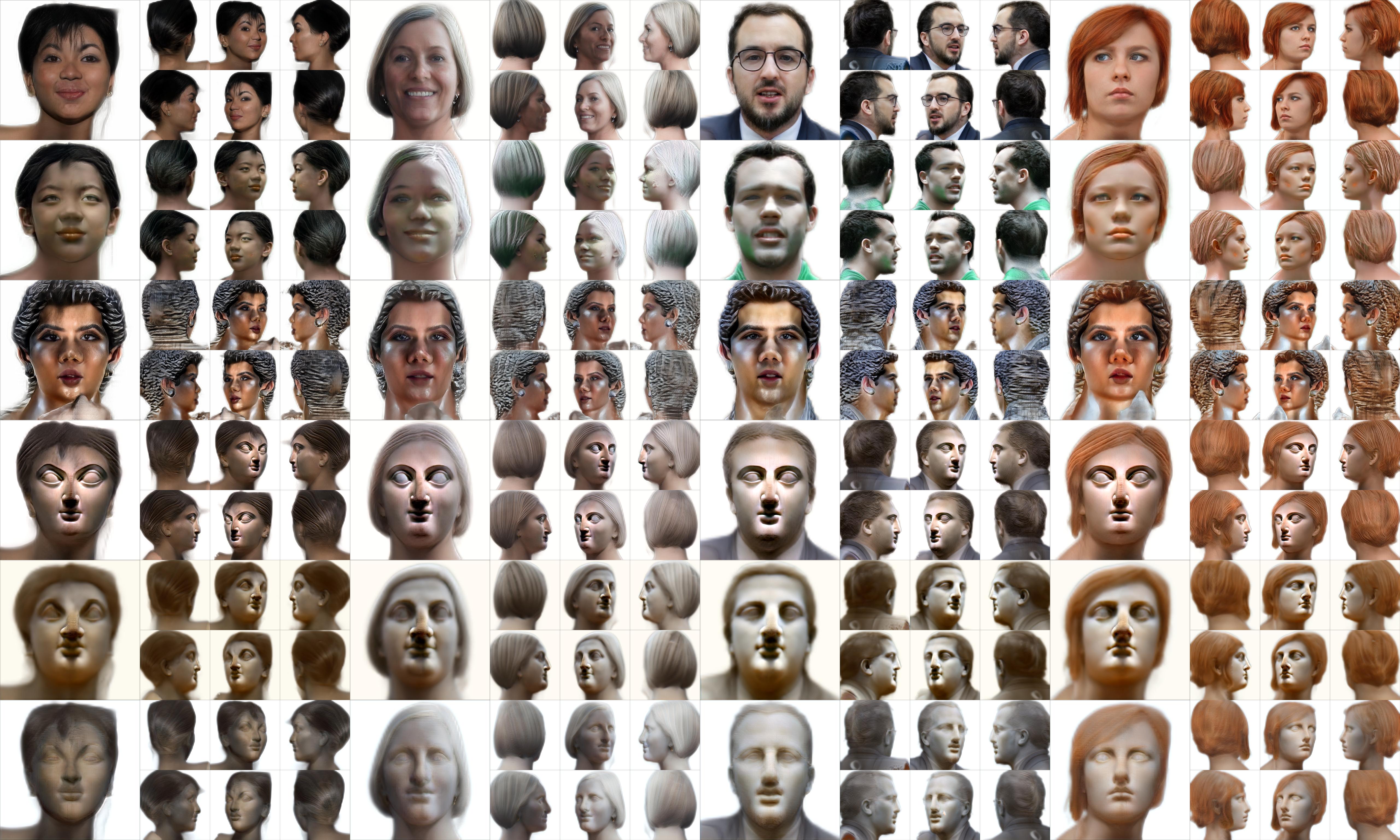}
\caption{Statue edits. From top to bottom: input, StyleCLIP, StyleGAN-NADA, StyleGANFusion, DiffusionGAN3D, ours.}
\label{fig:supp_statue}
\end{figure*}

\cref{fig:supp_joker,fig:supp_sketch,fig:supp_pixar,fig:supp_statue,fig:supp_werewolf,fig:supp_zombie} visualizes the outputs of methods for different prompts in 360-degrees.

\end{document}